\documentclass[12pt,letterpaper,oneside]{report}

\usepackage[letterpaper,margin=1in]{geometry}

\usepackage{setspace}
\usepackage{graphicx}
\usepackage[dvipsnames]{xcolor}
\usepackage[utf8]{inputenc}
\usepackage[linesnumbered,lined,ruled]{algorithm2e}
\usepackage{algpseudocode}
\usepackage{amsmath,amsfonts,amssymb,amsthm}
\usepackage{enumitem}
\usepackage{float}
\usepackage[T1]{fontenc}

\definecolor{Gred}{RGB}{250, 50, 50}
\definecolor{ToCgreen}{RGB}{0, 128, 0}
\usepackage[colorlinks]{hyperref}
\hypersetup{
  colorlinks=true,
  citecolor=black,
  linkcolor=black,
  filecolor=black,
  urlcolor=black
}

\usepackage{listings}
\usepackage{url}
\usepackage{xspace}
\usepackage[capitalize,noabbrev]{cleveref}
\crefname{tab}{Table}{Tables}
\crefname{fig}{Figure}{Figures}
\crefname{alg}{Algorithm}{}
\crefname{equ}{Equation}{}

\usepackage{lmodern}

\usepackage{microtype}

\usepackage{appendix}
\usepackage{multirow}
\usepackage{subcaption}
\usepackage{footnote}
\usepackage{tablefootnote}
\usepackage[sectionbib]{chapterbib}

\usepackage{booktabs}

\usepackage[numbers,sectionbib]{natbib}

\newcommand{\etdFrontMatter}[1]{
	\chapter*{#1}
	\addcontentsline{toc}{chapter}{#1}
}

\renewcommand{\abstract}{\etdFrontMatter{Abstract}}
\newcommand{\acknowledgments}{\etdFrontMatter{Acknowledgments}}

\begin{document}
\pagenumbering{roman}
\begin{titlepage}
    \newgeometry{top=0.5in, bottom=0.5in, left=1in, right=1in}

    \begin{center}
        \vspace*{\fill}
        \thispagestyle{empty}
        
        {\Large Deep Learning for Protein Complex Prediction and Design} \\
        \vspace{\baselineskip}
        
        \large
        by \\
        Ziwei Xie \\
        
        \vspace{\baselineskip}
        
        A thesis submitted
        \\ in partial fulfillment of the requirements for \\
        the degree of\\
        \vspace{\baselineskip}
        Doctor of Philosophy in Computer Science \\
        
        \vspace{\baselineskip}
        
        at the \\
        TOYOTA TECHNOLOGICAL INSTITUTE AT CHICAGO \\
        Chicago, Illinois \\
        \vspace{\baselineskip}
        March 2026 \\
        
        \vspace{\baselineskip}

        Thesis Committee: \\
        Jinbo Xu (Thesis Advisor)\\
        Madhur Tulsiani \\
        Aly Azeem Khan \\
        Gabriel Rocklin \\
        
        \vspace*{\fill}
    \end{center}
    
    \restoregeometry

    \clearpage
    \thispagestyle{empty}
    \begin{center}
        \null\vspace*{\fill}
        Copyright \copyright\ 2026 by Ziwei Xie \\
        All Rights Reserved
        \vskip 15pt\relax
        \vspace*{\fill}
    \end{center}

\end{titlepage}

\setcounter{tocdepth}{2}

\abstract
Accurately modeling and designing protein complex structures is a central problem in computational structural biology, with broad implications for understanding cellular function and developing therapeutics. This thesis investigates two fundamental
  aspects of this problem using deep learning: domain-specific architectures that capture the hierarchical nature of protein structures, and search algorithms that efficiently navigate the vast sequence spaces of protein complexes to identify
  interacting homologs for improving complex structure prediction and to design protein sequences.

  I first develop GLINTER, a graph neural network-based method for predicting interfacial contacts between proteins. GLINTER combines structural representations from monomeric structures with co-evolutionary signals extracted via transformer, outperforming existing methods on both heterodimeric and homodimeric targets and effectively guiding protein–protein docking. I then address a key bottleneck in complex structure prediction: identifying interacting homologs across species. I propose ESMPair, which uses protein language models to pair homologs from individual
  chains. ESMPair significantly improves structure prediction accuracy for heterodimers. Finally, I introduce RedNet, a multiscale graph transformer for fixed-backbone protein binder design. RedNet integrates backbone and side-chain information with a contrastive decoding algorithm that optimizes binding affinity and specificity, generating binders with improved thermodynamic properties that can discriminate between highly structurally similar targets.

  Together, these contributions demonstrate that domain-specific deep learning architectures, combined with principled search strategies, can extract complementary information from protein structures, evolutionary data, and experimental
  measurements to advance both protein complex structure prediction and its inverse problem, fixed-backbone protein binder design.

\acknowledgments
I would like to thank my advisor, Prof. Jinbo Xu, for his support, guidance,
  and insightful feedback throughout my PhD. I am grateful to my committee
members
  — Prof. Madhur Tulsiani, Prof. Aly Khan, and Prof. Gabriel Rocklin — for
their
   feedback on my proposal and thesis and for helping me navigate the various
  processes. I also thank Prof. Avrim Blum for serving as my local advisor in
  2024, and Prof. Greg Shakhnarovich for helping me navigate the final steps
  toward my thesis.

  I thank my lab mates for many inspiring discussions, and the  staff,
  colleagues, and faculty for making TTIC a wonderful place to learn and
conduct research.

 Finally, I want to thank my parents, my grandparents, and the rest of my
family for their unconditional love, trust, and support throughout my life.
This thesis would not have been possible without them.

\clearpage
\tableofcontents
\listoffigures
\listoftables
\listofalgorithms
\clearpage

\pagenumbering{arabic}
\chapter{Introduction}

Proteins, composed of linear chains of amino acids that fold into three-dimensional structures, perform a wide range of functions in many biological processes essential to life. The biological functions of proteins are primarily determined by its dynamical structures and specific interactions with other molecules. Protein-protein interactions form the basis of complex cellular machinery, including the proteasome for protein degradation, and protein complexes that regulate gene expression. Interactions between proteins and small molecules, enable cellular metabolism and signal transduction. Protein-nucleic acid interactions are essential for genome organization, transcriptional regulation, and epigenetic modifications that control gene expression programs. Accurately modeling and designing protein complexes is therefore a central problem in biology and has immense promise for biotechnological and therapeutic applications, from designing enzymes with novel catalytic properties to creating biologics that modulating disease-associated targets.

Experimental determination of protein complex structures through X-ray crystallography, while remaining the gold standard, is constrained by technical challenges in protein purification and crystallization, often requiring months of optimization and substantial resources. Similarly, directed evolution approaches for designing protein binders, though powerful, are fundamentally limited by the requirement for suitable starting templates, the development of robust high-throughput screening assays, and extensive iterations of mutagenesis and selection. These experimental techniques demand specialized equipment, domain expertise, and significant resources, making them challenging for rapid development. Consequently, computational methods that can accurately predict protein complex structures and design protein binders de novo offer a promising avenue to circumvent the limitations of experimental approaches.

Proteins are challenging molecular systems to model and design due to their vast conformational and sequence spaces, complex interatomic potentials, and long-range and high-order interactions between residues. Physics-based approaches using
  empirical force fields have succeeded in predicting structures of small proteins and designing sequences for idealized scaffolds, but struggle with large, multi-domain protein complexes.

  Machine learning—particularly deep learning—offers a complementary approach that addresses many of these limitations. These methods learn relationships between protein sequences, structures, and functions from large datasets, including the
  Protein Data Bank~\cite{berman2000protein}, UniProt~\cite{uniprot2023}, multiplexed assays of variant effects (MAVEs)~\cite{esposito2019mave}, and ClinVar~\cite{landrum2018clinvar}. AlphaFold2 exemplifies this approach: by leveraging
  evolutionary information from multiple sequence alignments, it predicts protein structures from sequence with unprecedented accuracy~\cite{jumper2021highly}.

  In this chapter, I provide an overview of the core tasks in predicting and designing protein complex structures, survey current machine learning approaches to these problems, and summarize the contributions of this thesis.
  
\section{Protein Structure Prediction}
\subsection{Experimental Protein Structure Determination}
X-ray crystallography has long served as the gold standard for determining protein structures at atomic resolution. Since the determination of myoglobin and hemoglobin structures in the late 1950s, the field has expanded
  dramatically; the Protein Data Bank (PDB) now archives over 200,000 experimentally determined structures~\cite{berman2000protein}. The technique works by directing X-rays through a protein crystal and analyzing the resulting diffraction pattern to reconstruct the three-dimensional structure, frequently around 2~\AA \cite{pdb101crystallographic}. However, obtaining well-ordered,
  diffraction-quality crystals remains a major bottleneck, particularly for flexible proteins, membrane proteins, and large complexes~\cite{smyth2000x}.

  Recent advances in Cryo-Electron Microscopy (Cryo-EM) have ushered in a ``resolution revolution''~\cite{kuhlbrandt2014resolution}, enabling the determination of macromolecular structures at near-atomic resolution without the need for
  crystallization. Compared to X-ray crystallography, Cryo-EM does not require crystallization, as samples are rapidly vitrified in their near-native hydrated state and imaged using an electron beam~\cite{cheng2018single}. Single-particle analysis, in
  particular, has become a powerful approach for resolving structures of complexes that resist crystallization. However, Cryo-EM still faces challenges with small proteins (typically below 50 kDa), and
   achieving resolutions below 2~\AA ~remains difficult for many targets.
   

 Nuclear Magnetic Resonance (NMR) spectroscopy complements these approaches by probing protein dynamics directly in solution, capturing conformational flexibility and transient interactions that are often lost during crystallization~\cite{wuthrich2003nmr}. NMR is particularly valuable for studying intrinsically disordered proteins and protein–ligand interactions, though it becomes increasingly challenging for proteins larger than 40 kDa.

\begin{figure}[ht]
    \centering
    \includegraphics[width=0.8\textwidth]{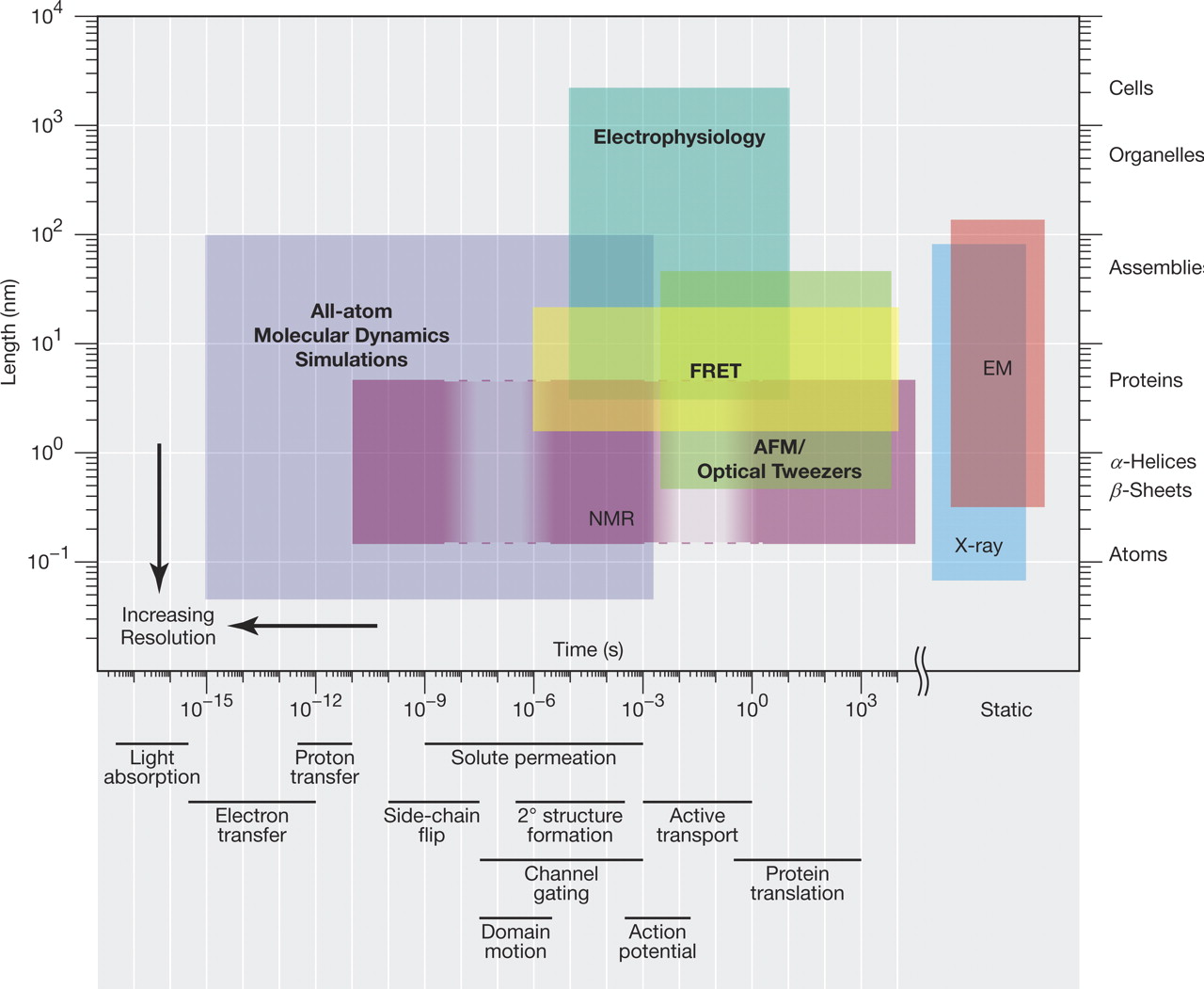}
    \caption{An illustration of spatial resolution of different methods.
    AFM, atomic force microscopy; EM, electron microscopy; FRET, Fr\"oster resonance energy transfer; NMR, nuclear magnetic resonance. Reproduced from \cite{dror2012biomolecular}, licensed under the Attribution–Noncommercial–Share Alike 3.0 Unported License (http://creativecommons.org/licenses/by-nc-sa/3.0/).
    }
    \label{fig:dror2011-dimension}
\end{figure}

  Several lower-resolution methods provide additional structural insights. The comparison of different structure determination methods is shown in ~\Cref{fig:dror2011-dimension}. Small-Angle X-ray Scattering (SAXS) characterizes the overall shape and oligomeric state of macromolecules in solution~\cite{putnam2007x}. Förster
  Resonance Energy Transfer (FRET) tracks real-time distance changes between labeled sites, revealing conformational dynamics~\cite{schuler2008protein}. Cross-linking mass spectrometry (XL-MS) and hydrogen–deuterium
  exchange mass spectrometry (HDX-MS) provide complementary information on spatial proximity and solvent accessibility, respectively~\cite{leitner2016crosslinking}.

\subsection{Computational Protein Structure Prediction}

Physics-based protein structure prediction rests on Anfinsen's thermodynamic hypothesis: that a protein adopts the conformation corresponding to the global minimum of its free-energy surface~\cite{anfinsen1973principles}. Translating this
  principle into practice requires two components: a reliable energy function to describe the protein's potential energy surface, and an efficient search algorithm for to find its global minimum.

  Molecular dynamics (MD) simulations approach this problem by modeling proteins and their surrounding solvent as classical particles governed by empirically derived force fields, propagating the system through time via numerical integration of
  equations of motion at femtosecond resolution~\cite{karplus2002molecular}. While MD can in principle provide atomic-resolution models of conformational equilibria and folding transitions, its effectiveness is limited by two persistent
  challenges: the accuracy of current force fields in reproducing true energy surfaces~\cite{piana2011robust}, and the difficulty of sampling conformational space sufficiently to observe folding events. Specialized hardware such as the Anton
  supercomputer has helped address the sampling bottleneck, enabling equilibrium simulations long enough to observe the folding of small, fast-folding proteins~\cite{lindorff2011fast}, but extending this to larger and
  slower-folding proteins remains prohibitive.

  Ab initio modeling methods such as Rosetta take a different approach, directly searching for low-energy conformations by assembling short fragments of known protein structures using Monte Carlo sampling guided by physically motivated energy
  functions that emphasize short-range interactions—van der Waals, hydrogen bonding, and desolvation—while dampening long-range electrostatics~\cite{bradley2005toward}. Combined with all-atom refinement, Rosetta has achieved
  near-atomic accuracy for small proteins across all secondary structure classes~\cite{das2007structure}. However, both MD and ab initio methods scale poorly to large proteins and multi-component complexes, where the combinatorial growth of
  conformational space overwhelms current sampling strategies and energy function accuracy~\cite{kuhlman2019advances}.

The recent advances in protein structure prediction have come from the effective utilization of evolutionary information. Early approaches relied on homology modeling, which builds three-dimensional models by copying structural
  fragments from evolutionarily related proteins with experimentally determined structures~\cite{kelley2015phyre2}. As more structures were deposited in the PDB and sequence databases grew, homology modeling became increasingly powerful, but
  remained fundamentally limited to proteins with detectable sequence similarity to known structures.

Residues that are close in three-dimensional space often co-evolve to maintain structural stability, leaving detectable patterns in multiple sequence alignments (MSAs). As early as 1999, Lapedes et al. proposed using Markov Random Fields
  (MRFs) to model pairwise couplings between co-evolving residues~\cite{lapedes1999correlated}, but limited sequence data and computational resources meant this work went largely unnoticed. A decade later, Weigt et al. developed a
  message-passing algorithm to infer direct co-evolutionary couplings across protein–protein interfaces~\cite{weigt2009identification}, and Marks et al. showed that co-evolutionary signals alone could determine accurate protein
  folds~\cite{marks2011protein}. These methods depended critically on large, diverse MSAs—a requirement increasingly met by the rapid growth of sequence databases.


  Deep learning further transformed contact prediction and protein structure prediction. MetaPSICOV combined multiple co-evolution methods within a neural network~\cite{jones2015metapsicov}, and RaptorX-Contact showed that deep residual networks could substantially improve
  accuracy~\cite{wang2017accurate}. AlphaFold1 predicted inter-residue distance distributions from co-evolutionary features, achieving top performance at CASP13~\cite{senior2020improved}. AlphaFold2 shifted to end-to-end structure prediction,
  using an attention-based architecture to directly output three-dimensional coordinates with accuracy comparable to experimental methods~\cite{jumper2021highly}. AlphaFold-Multimer extended this to protein complexes~\cite{evans2022protein}, and
   most recently, AlphaFold3 and RoseTTAFold All-Atom have expanded to incorporate nucleic acids, small molecules, ions, and covalent modifications within unified frameworks~\cite{abramson2024accurate, krishna2024generalized}.

Despite these advances, current methods still depend heavily on homologous sequences and perform poorly on proteins with shallow MSAs, such as orphan proteins with few known relatives~\cite{chakravarty2022alphafold}. Most methods predict a
  single static structure and cannot capture the conformational ensembles that underlie protein function~\cite{saldano2022impact}. Additionally, these models do not explicitly learn physical energetics, limiting their ability to predict how
  mutations affect stability or binding affinity~\cite{buel2022can}.

\section{Computational Protein Design}

\subsection{Experimental Protein Engineering Methods}
Directed evolution mimics natural selection in the laboratory to engineer proteins with improved or new properties. The process works by repeatedly generating sequence diversity—through random mutagenesis or DNA recombination—then screening or
   selecting for improved variants. A key challenge is linking each protein to the DNA that encodes it. Display technologies solve this by attaching each variant to its genetic template. Phage display presents variants on bacteriophage surfaces,
   enabling screening of libraries exceeding $10^{10}$ variants~\cite{smith1985filamentous, winter2019harnessing}, while yeast surface display enables quantitative selection using fluorescence-activated cell sorting (FACS)~\cite{boder1997yeast}.
   Continuous evolution platforms such as phage-assisted continuous evolution (PACE) take this further by combining mutagenesis and selection into a single uninterrupted process, achieving hundreds of generations of evolution in
  days~\cite{esvelt2011system}.

  Despite its impact, directed evolution can screen only $\sim10^{8}$--$10^{13}$ variants, a small fraction of the theoretical sequence space ($20^{N}$ for a protein of $N$ residues), and generally requires a starting template with some
  level of the desired activity. Rational design circumvents these limitations by using structural knowledge to introduce targeted mutations, though it depends on a detailed understanding of structure–function relationships. Semi-rational
  approaches bridge this gap by focusing directed evolution on specific regions—such as active-site residues—using combinatorial saturation mutagenesis to reduce library size while enriching for functional variants~\cite{lutz2010beyond}.
  
\subsection{Computational Protein Design}
Structure-based protein design aims to identify amino acid sequences that fold into a desired three-dimensional structure and perform a specific function. The field was pioneered by physics-based approaches that use energy functions to
  evaluate sequence compatibility with a target backbone. Dahiyat and Mayo demonstrated the first fully automated computational design of a protein, screening $\sim$1.9 $\times$ $10^{27}$ sequences using energy functions ~\cite{dahiyat1997novo}. Kuhlman et al. subsequently used the Rosetta energy function to design Top7, a protein with a fold not found in nature, marking a milestone for the field~\cite{kuhlman2003design}. OSPREY introduced provable
  algorithms with guarantees of finding the global minimum energy conformation, enabling rigorous design with ensemble-based methods~\cite{hallen2018osprey}. These physics-based methods optimize sequences and side-chain conformations (rotamers)
  to minimize energy, and have been extended to multistate design, where sequences are simultaneously optimized across multiple conformational states to achieve specificity—for example, stabilizing a desired binding interaction while
   destabilizing off-target ones~\cite{davey2012multistate}. Another key application of structure-based design is protein binder engineering. Cao et al. demonstrated that miniprotein binders with nanomolar to picomolar affinity could be designed de novo from the target structure alone using
  Rosetta~\cite{cao2022design}.
  
\begin{figure}[t]
    \centering
    \includegraphics[width=\textwidth]{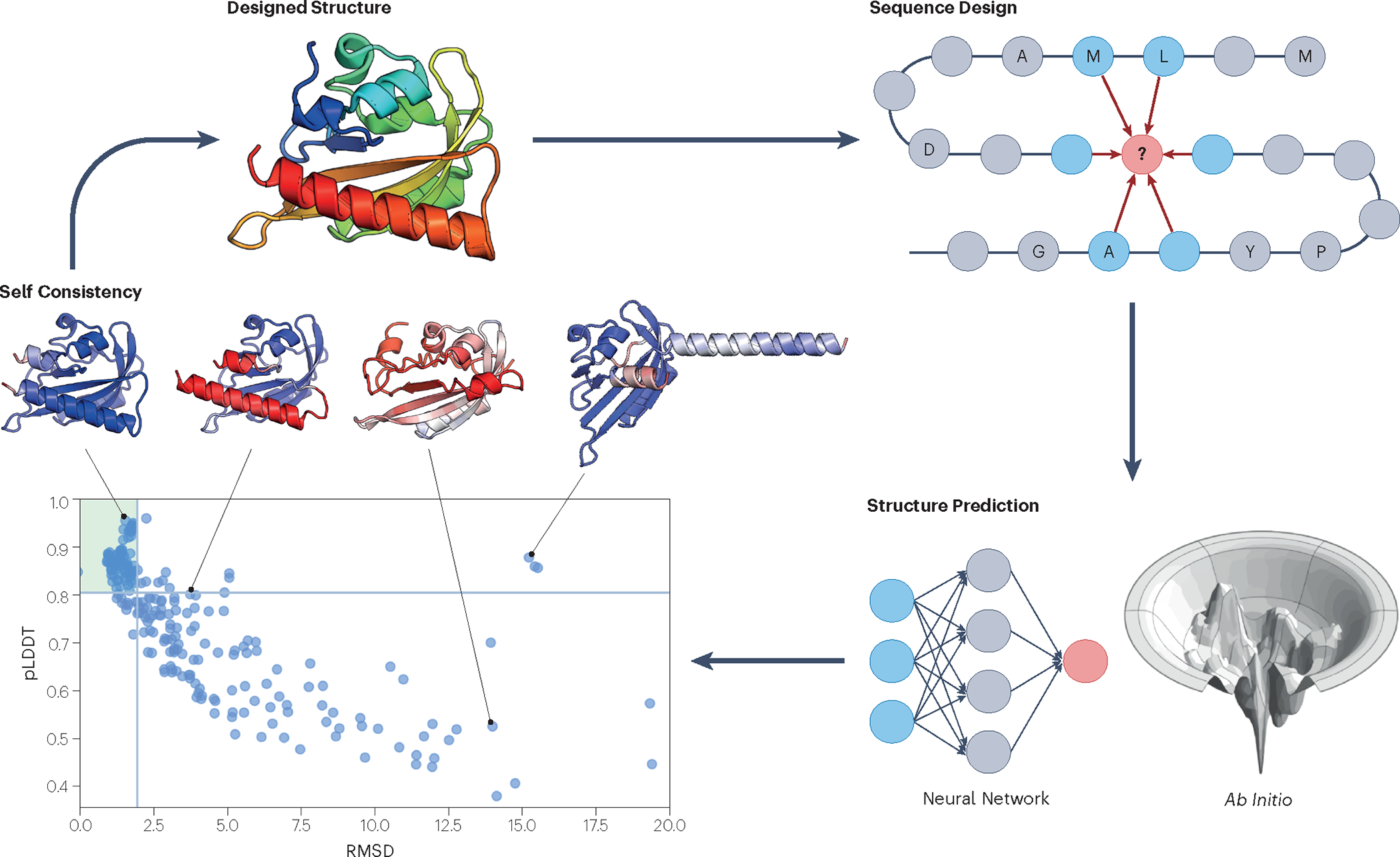}
    \caption{Overview of the deep learning protein design pipeline, illustrating backbone generation, sequence design, and structure prediction-based filtering. Figure adapted from~\cite{chu2024sparks}.}
    \label{fig:chu2024fig5}
  \end{figure}

  Deep learning is rapidly transforming each component of the conventional structure-based protein design pipeline: backbone generation, sequence design, and filtering~\cite{chu2024sparks}; a workflow shown in \Cref{fig:chu2024fig5}. 

Early generative approaches used GANs~\cite{anand2018generative} and VAEs such as Ig-VAE~\cite{eguchi2022igvae}, but struggled with structural plausibility and diversity. Hallucination-based methods offered an alternative
  by optimizing sequences through gradient descent on structure prediction networks to generate novel topologies~\cite{anishchenko2021novo}, and inpainting approaches fill in sequence and structure around functional sites in a single forward
  pass~\cite{wang2022scaffolding}. BindCraft applied AlphaFold2-based hallucination to binder design, achieving experimental success rates of 10–100\% across diverse targets~\cite{pacesa2025one}. Diffusion models such as RFdiffusion~\cite{watson2023novo} and Chroma~\cite{ingraham2023illuminating} generate novel, designable protein backbones from noise. A growing frontier extends this to all-atom generation.
  Protpardelle~\cite{chu2024allatom} and Pallatom~\cite{qu2024pallatom} generate full side-chain conformations alongside backbone and sequence, while RFdiffusion2 designs enzyme active sites directly from functional group geometries at
  atom-level resolution~\cite{ahern2025rfdiffusion2}. AlphaProteo similarly generates all-atom protein
  binders conditioned on target structures~\cite{zambaldi2024alphaproteo}.

Given a fixed backbone, structure-conditioned methods such as ProteinMPNN use message passing on protein graphs to predict sequences that fold into the target structure, achieving higher experimental success rates than
  Rosetta~\cite{ingraham2019generative, dauparas2022robust}. Complementary to structure-conditioned approaches, protein language models such as ProGen, ESM, and EvoDiff generate protein sequences without requiring a backbone
  template~\cite{rives2021biological, madani2023large, alamdari2023protein, hayes2025simulating}.
  
  Designed candidates must be filtered for foldability, stability, and other properties. Structure prediction methods~\cite{jumper2021highly, abramson2024accurate} serve as in silico validation, verifying that designed sequences fold into
  intended structures before experimental testing~\cite{zambaldi2024denovo, corso2025boltzgen, pacesa2025one}. As discussed in the previous section, current structure prediction models produce single static structures and cannot
  capture conformational ensembles, and binding affinity and mutational effect prediction remain unreliable, limiting their utility as design filters.

Together, advances in backbone generation, sequence design, and computational filtering have enabled the design of novel binders, enzymes, and therapeutic proteins~\cite{kortemme2024novo, chu2024sparks}. Yet significant challenges remain.
  Current workflows require generating and screening thousands of candidates to find a few that function experimentally. Properties critical for therapeutic development—such as solubility, low immunogenicity, and manufacturability—are rarely
  optimized during design. Achieving catalytic efficiency comparable to natural enzymes remains particularly difficult. Furthermore, co-design and all-atom generative methods, while promising, currently lag behind the two-stage pipeline of
  backbone diffusion followed by fixed-backbone sequence design in both designability and structural accuracy.

 \section{Deep Learning for Protein Modeling}

  Deep learning for protein modeling is a fast-moving field. We give a brief survey of recent progress across structure prediction, design, dynamics, self-supervised representation learning, and other downstream applications.

  \paragraph{Structure Prediction.}
  AlphaFold2 introduced an end-to-end architecture that predicts atomic coordinates directly from sequence and MSA inputs~\cite{jumper2021highly}. Its core components, the Evoformer for joint sequence-MSA representation learning and the structure module with Invariant
  Point Attention for SE(3)-equivariant coordinate prediction, define the template that subsequent methods have built on. AlphaFold-Multimer extends this framework to protein complexes by training on multimeric structures and pairing chains across
  MSAs~\cite{evans2022protein}. Independent re-implementations broadened the architecture: RoseTTAFold uses a three-track network that jointly reasons over sequence, residue pairs, and atomic coordinates~\cite{baek2021accurate}; OpenFold reproduces AlphaFold2 with open
  training data and code~\cite{ahdritz2024openfold}; ESMFold replaces MSAs with embeddings from a large protein language model, trading evolutionary signal for inference speed and applicability to orphan sequences~\cite{lin2023evolutionary}. Recent work extends these
  architectures to all-atom biomolecular assemblies. RoseTTAFold All-Atom and AlphaFold3 unify proteins, nucleic acids, small molecules, ions, and covalent modifications in a single framework, with AlphaFold3 additionally replacing the structure module with a diffusion
  head over atomic coordinates~\cite{krishna2024generalized, abramson2024accurate}.

  \paragraph{Protein Design.} Hallucination methods optimize sequences through a frozen predictor by gradient ascent on confidence or geometric objectives, generating structures with desired
  properties~\cite{anishchenko2021novo}. BindCraft applies AlphaFold2 hallucination to binder design, achieving experimental success rates of ten to one hundred percent across diverse targets~\cite{pacesa2025one}. RFdiffusion adapts the RoseTTAFold structure module into a
   denoising diffusion model that generates backbones from noise, supporting unconditional generation, motif scaffolding, and binder design~\cite{watson2023novo}. RFdiffusion2 and AlphaProteo extend these ideas to all-atom and target-conditioned binder
  generation~\cite{ahern2025rfdiffusion2, zambaldi2024denovo}. Pretrained predictors also serve as in silico filters: designs from sequence-only or backbone-conditioned generators are scored by predictor confidence metrics (pLDDT, pAE, ipTM), and self-consistency between
  the designed structure and the predictor's reconstruction from the designed sequence is the primary success criterion before experimental testing~\cite{watson2023novo, pacesa2025one, corso2025boltzgen}. This filter-and-screen workflow now underlies most experimentally
  validated design pipelines.

  \paragraph{Modeling Protein Dynamics.}
  Structure predictors output single static conformations, while biological function depends on conformational ensembles. A growing line of work adapts these architectures to capture dynamics. AFCluster clusters and subsamples MSAs to elicit alternative conformations from
   AlphaFold2 at inference time without retraining, exposing functional states such as kinase active and inactive forms~\cite{wayment2024predicting}. AlphaFlow and ESMFlow fine-tune AlphaFold2 and ESMFold with flow matching to sample diverse conformations from sequence,
  with training targets drawn from the PDB or short molecular dynamics trajectories~\cite{jing2024alphafold}. BioEmu fine-tunes a structure prediction backbone on a large in-house corpus of all-atom molecular dynamics trajectories to approximate Boltzmann-distributed
  ensembles in a single forward pass, at much lower cost than direct simulation~\cite{lewis2025scalable}. Together, these methods point toward end-to-end networks that approximate distributions over conformations rather than point estimates.

  \paragraph{Self-Supervised Learning of Protein Representations.}
  Protein language models learn representations of amino acid sequences from large protein sequence databases. The ESM family trains off-the-shelf transformers with masked language modeling on UniRef, producing embeddings that transfer to structure, function, and variant effect
  prediction~\cite{rives2021biological, lin2023evolutionary}. MSA Transformer extends the architectures to aligned sequences via axial attention, capturing co-evolutionary signal directly in attention maps that prove useful for contact and interface
  prediction~\cite{rao2021msa}. Hybrid models inject structural information into the sequence vocabulary: ProstT5 conditions a sequence-to-sequence model on 3Di structural tokens and SaProt tokenizes residues jointly with Foldseek structural alphabets~\cite{su2023saprot,
   heinzinger2023prostt5}. Hybrid sequence-structure models consistently outperform pure-sequence models on variant effect prediction and
  structure-based design.

  \paragraph{Other Applications.}
  Structure prediction architectures and protein language models have been adapted to a range of other problems. AlphaMissense fine-tunes AlphaFold2 with population frequency signal to score missense pathogenicity at proteome scale~\cite{cheng2023accurate}, and zero-shot
  scoring with PLM likelihoods correlates with deep mutational scanning data and clinical pathogenicity~\cite{meier2021language, frazer2021disease}. Variant effect prediction nonetheless remains challenging: accuracy degrades for low-frequency variants, multi-residue
  mutations, and proteins with shallow MSAs. Predicted structures also accelerate experimental determination, serving as molecular replacement templates in X-ray crystallography
   for targets that resist conventional phasing~\cite{terwilliger2024alphafold}.

\section{Contributions}

In this thesis, I investigate core aspects of modeling and designing protein complexes using deep learning.

  In Chapter 2, we develop GLINTER, a supervised deep learning method for predicting interfacial contacts between proteins. GLINTER combines structural representations from monomers with attention maps from the MSA Transformer (ESM-MSA) derived
  from interologs. We show that GLINTER outperforms existing methods such as ComplexContact and DeepHomo on both heterodimers and homodimers, and can effectively guide protein–protein docking.

  Chapter 3 addresses a related challenge: identifying interacting homologs (interologs) using protein language models (PLMs). We propose ESMPair, which uses column-wise attention scores from the pretrained ESM-MSA-1b model to pair sequences
  from individual chains. ESMPair significantly improves complex structure prediction, particularly for heterodimers and eukaryotic targets where phylogeny-based pairing methods struggle.

  Chapter 4 introduces RedNet, a framework for fixed-backbone binder design that builds on the graph neural network approach established in GLINTER. RedNet uses a multiscale graph transformer that incorporates both backbone geometry and
  side-chain information, along with a contrastive decoding algorithm that optimizes for binding affinity and specificity. We demonstrate that RedNet generates binders with improved properties and can discriminate between highly similar targets.

  Chapter 5 concludes with future directions for improving biomolecular modeling and expanding its scientific applications.

\bibliographystyle{siam}
\bibliography{references}

\chapter{Graph Learning of Protein Interfacial Contacts}

\noindent This chapter presents work with Jinbo Xu. It was published in \textit{Bioinformatics} in 2022~\citep{xie2022deep}. Code is available at \url{https://github.com/zw2x/glinter}.

\section{Introduction}

Proteins perform functions by interacting with other molecules or forming protein complexes. As a result, the full characterization of protein–protein interactions with structural details is crucial to atom-level understanding of protein functions. The in silico structural characterization of protein complexes, or quaternary protein structure prediction, is a longstanding challenge in computational structural biology. Given individual protein chains (and possibly their structures), interfacial contact prediction aims to predict which pairs of residues on the protein surface are geometrically close to each other after the protein chains bind together. Interfacial contacts may facilitate generating and filtering docking decoys \citep{baldassi2014fast,geng2020iscore,hopf2014sequence,ovchinnikov2014robust}, and reveal important biophysical properties and evolutionary information of protein interfaces \citep{uguzzoni2017large}. They are also useful for the redesign of protein–protein interfaces \citep{laine2015local} and prediction of binding affinity \citep{vangone2015contacts}.

Co-evolution analysis by global statistical methods \citep{burger2008accurate,weigt2009identification} has been used for inter-protein contact prediction. A recent study \citep{cong2019protein} showed that co-evolution-based in silico protein–protein interaction screening methods produced more true protein–protein interactions than high-throughput experimental techniques. Nevertheless, accurate co-evolution analysis needs a large number of sequence homologs and thus, may not work well on a large portion of heterodimers for which it is very challenging to find sufficient number of interacting paralogs (interlogs) \citep{bitbol2016inferring,gueudre2016simultaneous,zeng2018complexcontact}. On the other hand, protein language models, which are trained on individual protein sequences or multiple sequence alignment (MSAs), are shown to perform similarly as or better than global statistical methods on intra-chain contact prediction when few sequence homologs are available \citep{rao2021msa,rives2021biological}. It was shown before that a deep learning model trained by individual protein chains works fine on protein complex contact prediction \citep{zeng2018complexcontact,zhou2018deep}. Therefore, we hypothesize that a deep language model trained on individual protein chains may also generalize well to protein–protein interactions, reducing the required number of interlogs. Protein language models are also much faster since they require only one-time forward computation during inference and thus, more suitable for proteome-scale screening of protein–protein interactions.

RaptorX ComplexContact \citep{zeng2018complexcontact,zhou2018deep} possibly is the first deep learning method for interfacial contact prediction. It is mainly developed for heterodimers, although can be used for homodimers. Nevertheless, its deep models are purely trained on individual protein chains instead of protein complexes. Further, ComplexContact does not make use of any (experimental or predicted) structures of constituent monomers of a dimer. Recently, some deep learning methods are developed specifically for contact prediction of a homodimer, e.g. DNCON\_inter \citep{quadir2021dncon2} and DeepHomo \citep{yan2021accurate}, both using ResNet originally implemented in RaptorX \citep{wang2017accurate}. In addition to evolution information, DeepHomo uses docking maps, native intra-chain contacts, and experimental structural features derived from monomers to achieve state-of-the-art performance. However, it is slow in calculating docking maps and thus, cannot scale well to proteome-scale prediction. Some deep learning methods also use learned representations of tertiary structures, including voxels \citep{derevyanko2019protein,townshend2019end} and radial/point cloud representations on protein surfaces \citep{dai2021protein,gainza2020deciphering,sverrisson2020fast}. Meanwhile, some representations include anisotropy information in the structures \citep{fout2017protein,pittala2020learning} while others do not.

Given the tremendous progress in protein structure prediction \citep{jing2021fast,jumper2021highly,wang2017accurate,xu2019distance,xu2021improved} and the fast growing number of protein sequences, it is important to leverage predicted structures of constituent monomers and large sequence corpus to produce accurate, proteome-scale interfacial contact predictions. An interfacial contact prediction method shall effectively extract coevolution signals from a small number of interlogs, and make use of predicted structures of constituent monomers. Here, we propose a new supervised deep learning method GLINTER for interfacial contact prediction that integrates representations learned from (experimental and predicted) monomer structures and attentions generated by the MSA Transformer (ESM-MSA) \citep{rao2021msa} from interlogs of the dimer under prediction. GLINTER applies to both heterodimers and homodimers, outperforming ComplexContact, DeepHomo and BIPSPI on the 13th and 14th CASP-CAPRI datasets. The contacts predicted by GLINTER may also improve the ranking of the HDOCK-generated docking decoys \citep{yan2017hdock}. Further, our method runs very quickly, which makes it suitable for proteome-scale study.

\section{Data and methods}

\subsection{Network architecture}
\begin{figure}[!ht]
    \centering
    \includegraphics[width=0.8\textwidth]{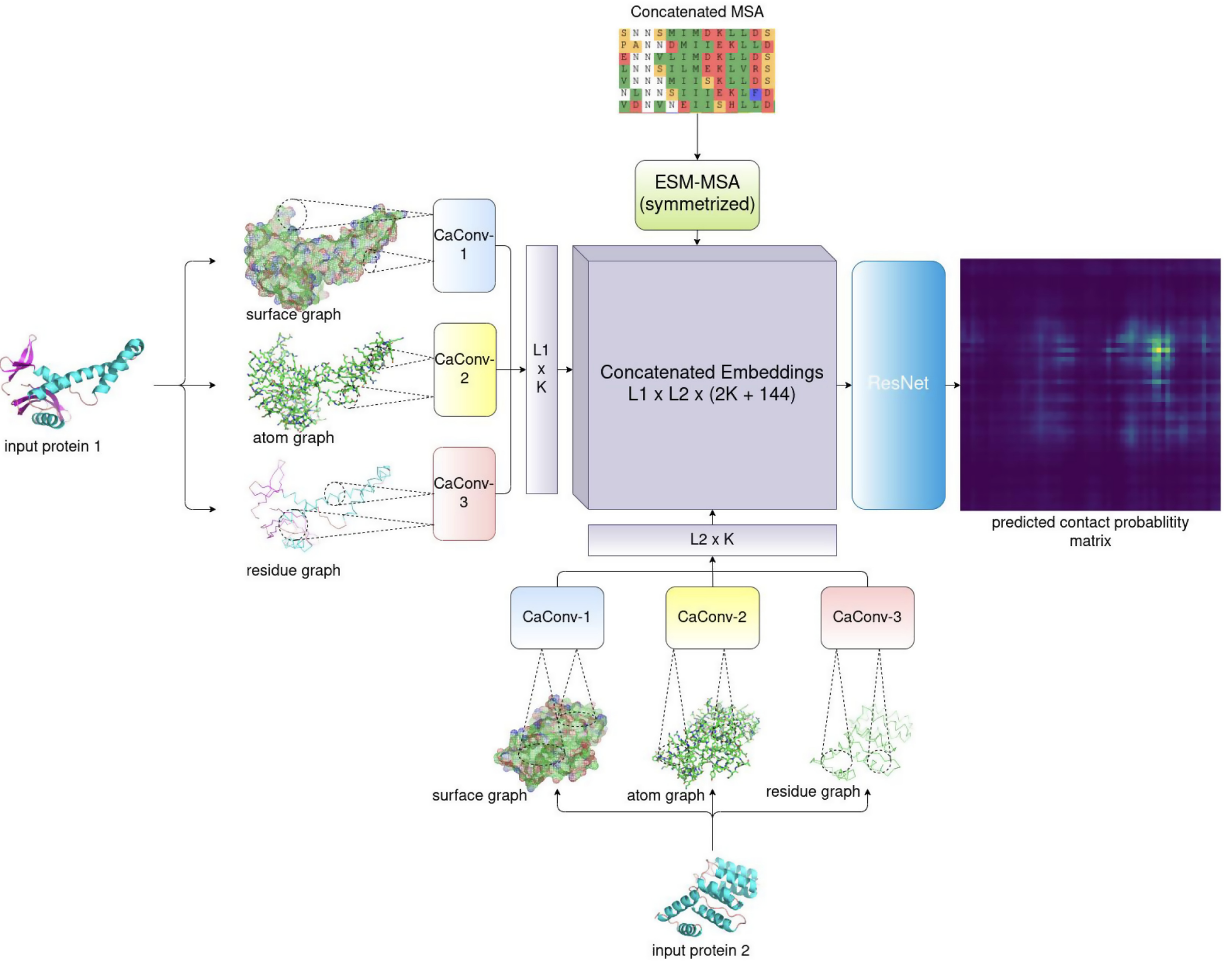}
    \caption{Overview of the GLINTER architecture. $L_1$ and $L_2$ are the lengths of the two protein chains, K is the number of channels in a CaConv layer and 144 is the total number
of heads in the row attention weights generated by Facebook’s MSA Transformer (Rao et al., 2021)}
    \label{fig:glinter-arch}
\end{figure}
As shown in \cref{fig:glinter-arch}, our network, denoted as GLINTER, consists
of two major modules: a Siamese graph convolutional network
(GCN) and a 16-block ResNet \citep{he2016deep}. The GCN extracts local features from three types of graphs
derived from monomer structures. The ResNet takes as input the
outputs of the GCN module and the attention weights generated by
the MSA Transformer \citep{rao2021msa} and yields interfacial contact prediction. One ResNet block has two convolutional layers,
each with 96 filters and a $3\times3$ kernel. ELU and BatchNorm are used
in each block. ResNet is connected to a fully connected layer and a
softmax layer for contact probability prediction. The pseudocode of the main architecture is shown in \Cref{alg:glinter_main}.

\begin{algorithm}[t!]
    \setlength{\algomargin}{0em}
    \DontPrintSemicolon
    \SetAlgoSkip{medskip}

    \caption{GLINTER Main Architecture}\label{alg:glinter_main}

    \SetKwProg{Fn}{def}{:}{}
    \SetKwComment{Comment}{\# }{}
    \SetKw{KwReturn}{return}

    \SetKwFunction{GLINTER}{GLINTER}
    \SetKwFunction{GCN}{GCN}
    \SetKwFunction{MSATransformer}{MSATransformer}
    \SetKwFunction{ResNet}{ResNet}
    \SetKwFunction{FC}{FC}
    \SetKwFunction{Softmax}{Softmax}
    \SetKwFunction{BuildGraphs}{BuildGraphs}
    \SetKwFunction{Concat}{Concat}
    \SetKwFunction{OuterConcat}{OuterConcat}

    \Comment{$\mathcal{P}_r$, $\mathcal{P}_l$: monomer structures; $\mathbf{M}$: paired MSA}

    \Fn{\GLINTER{$\mathcal{P}_r$, $\mathcal{P}_l$, $\mathbf{M}$}}{

      \Comment{Construct graphs from each monomer}
      $\mathcal{G}_r \gets \BuildGraphs(\mathcal{P}_r)$ \Comment*[r]{C$\alpha$, atom, surface graphs}
      $\mathcal{G}_l \gets \BuildGraphs(\mathcal{P}_l)$\;

      \BlankLine
      \Comment{Extract per-monomer features via Siamese GCN}
      $\mathbf{h}_r \gets \GCN(\mathcal{G}_r)$\;
      $\mathbf{h}_l \gets \GCN(\mathcal{G}_l)$ \Comment*[r]{shared weights}

      \BlankLine
      \Comment{Form pairwise representation}
      $\mathbf{P} \gets \OuterConcat(\mathbf{h}_r, \mathbf{h}_l)$ \Comment*[r]{$\mathbf{P}_{ij} = [\mathbf{h}_r^{(i)} \| \mathbf{h}_l^{(j)}]$}

      \BlankLine
      \Comment{Extract and symmetrize ESM row attention}
      $\mathbf{A} \gets \MSATransformer(\mathbf{M})$\;
      $\mathbf{A} \gets \mathbf{A}_{[:N_r, N_r:]} + \mathbf{A}_{[N_r:, :N_r]}^\top$\;

      \BlankLine
      \Comment{Concatenate and predict contacts}
      $\mathbf{Z} \gets \Concat(\mathbf{P}, \mathbf{A})$\;
      $\mathbf{Z} \gets \ResNet(\mathbf{Z})$ \Comment*[r]{16 blocks, 96 filters, $3 \times 3$ kernel}
      $\mathbf{C} \gets \Softmax(\FC(\mathbf{Z}))$\;

      \KwReturn $\mathbf{C}$\;
    }

\end{algorithm}
 
At each graph convolution layer (denoted as CaConv), we calculate the message for a graph edge and node as follows. For an edge
$e$, we feed its feature and the features of its two ends to a subnetwork to generate a message. For a node $q$, we first aggregate all messages of its adjacent nodes using max pooling, and then pass the result to a subnetwork to generate a message of $q$, i.e.
\begin{equation}
    g(q) = g(\max_{v \in N_q} f([x_q,x_v,e(q,v)]))
\end{equation}

where $x_q$ is the feature of node $q$, $v$
is a node in the neighborhood $N_q$ of $q$, $x_v$ is the feature of $v$, $e(q,v)$ is
the feature of edge $(q, v)$ and the non-linear functions $g$ and $f$ are
two fully connected layers of 128 hidden units with BatchNorm and
ReLU. The pseudocode of the GCN is shown in \Cref{alg:glinter_caconv}.

\begin{algorithm}[t!]
    \setlength{\algomargin}{0em}
    \DontPrintSemicolon
    \SetAlgoSkip{medskip}

    \caption{CaConv: Graph Convolution with Local Reference Frames}\label{alg:glinter_caconv}

    \SetKwProg{Fn}{def}{:}{}
    \SetKwComment{Comment}{\# }{}
    \SetKw{KwReturn}{return}

    \SetKwFunction{CaConv}{CaConv}
    \SetKwFunction{MLP}{MLP}
    \SetKwFunction{MaxPool}{MaxPool}

    \Comment{$\mathbf{X}$: node features; $\mathbf{E}$: edge features; $\mathbf{p}$: positions; $\mathbf{R}$: local frames}

    \Fn{\CaConv{$\mathbf{X}$, $\mathbf{E}$, $\mathbf{p}$, $\mathbf{R}$}}{

      \For{$q \in \mathcal{G}$}{

        \Comment{Compute edge messages in local reference frame}
        \For{$v \in \mathcal{N}_q$}{
          $\tilde{\mathbf{p}}_v \gets \mathbf{R}_q^\top (\mathbf{p}_v - \mathbf{p}_q)$ \Comment*[r]{standardize coordinates}
          $\mathbf{m}_{q,v} \gets \MLP([\mathbf{x}_q \| \mathbf{x}_v \| e(q,v) \| \tilde{\mathbf{p}}_v])$\;
        }

        \BlankLine
        \Comment{Aggregate and update node feature}
        $\bar{\mathbf{m}}_q \gets \MaxPool_{v \in \mathcal{N}_q}(\mathbf{m}_{q,v})$\;
        $\mathbf{x}'_q \gets \MLP(\bar{\mathbf{m}}_q)$\;
      }

      \KwReturn $\mathbf{X}'$\;
    }

  \end{algorithm}

Both coordinates and normals are used to represent the geometric properties of a monomer structure \citep{sverrisson2020fast}. We
standardize the geometric features so that they are invariant to the
coordinate system used by the monomer structure. While calculating
an message for any node $q$ (i.e. computation of $f$), all the adjacent
nodes of $q$ are first translated using q as the origin, and then rotated
using its predefined local reference frame \citep{pages2019protein,sanyal2020proteingcn}. The standardized features are then concatenated with
other features to form the actual inputs of function $f$. 
We use a separate graph convolution network (GCN) module to
process each graph. When multiple graphs are used for a monomer,
the outputs of all its GCN modules are concatenated to form a single
output vector of this monomer. The outputs of two monomers are
then outer-concatenated to form a pairwise representation of this
dimer. When the ESM row attention weight is used, the attention
matrix generated by Facebook’s MSA Transformer is concatenated
to the pairwise representation, which is then fed to the ResNet for
interfacial contact probability prediction.

\subsection{Features}
\paragraph{Graph representation of protein structures} 
We build three different graphs from one protein structure: residue
graph, atom graph and surface graph. In a residue graph, a node is a
residue represented by its CA atom, and there is an edge between
two residue nodes if and only if the Euclidean distance between their
CA atoms is within a certain cutoff, e.g. 8 \AA . In an atom graph, a
node is a heavy atom or a residue represented by its CA atom, and
there is one edge between one residue node and one atom node if
and only if their Euclidean distance is within a certain cutoff.

We use Reduce \citep{word1999asparagine}, MSMS \citep{sanner1996reduced}
and trimesh \citep{dawson2019trimesh} to construct the triangulated surface of a protein structure. To build a surface graph, we first use Reduce to add hydrogen atoms and then construct the triangulated surface of a protein structure using MSMS. MSMS may generate a large number of vertices on the triangulated surface. We use the ``remove\_closest'' algorithm in the trimesh library to sample a subset of vertices such that any two of them are at least 0.8\AA\ away from each other. The cutoff is set to 0.8\AA, because the number of remaining vertices does not change much when it is less than 0.8\AA. The surface can be essentially interpreted as a mesh enclosing
the protein. Two neighboring triangles in the surface share either
one edge or at least one vertex. In a surface graph, one node represents one residue or one vertex on the triangulated surface. There is
one edge between one residue node and one triangle vertex if and
only if their Euclidean distance is within a certain cutoff. It takes
only a few seconds to build a surface graph and thus, our method
scales well on large-scale prediction \citep{cong2019protein}.

\paragraph{Features}
\begin{table}[!t]
  \centering
  \caption{Features used in spatial graphs, where $L$ is the number of residues, $N$ is the number of atoms, $M$ is the number of sampled surface vertices, and $E$ is the number of edges in an atom graph.}
  \label{tab:glinter-spatial-graph-features}
  \begin{tabular}{p{3cm}p{7cm}c}
  \toprule
  Spatial Graph & Feature & Dimension \\
  \midrule
  Residue graph & Positional Specific Scoring Matrix (PSSM) & $L \times 20$ \\
                & Residue solvent accessible surface areas (SASA) & $L \times 1$ \\
                & Amino acid types (AA) & $L \times 21$ \\
                & Normalized sequential positions & $L \times 1$ \\
                & C$_\alpha$ coordinates & $L \times 3$ \\
                & C$_\alpha$-centered local reference frame (LRF) & $L \times 3 \times 3$ \\
  \midrule
  Atom graph    & Atom solvent accessible surface areas & $N \times 1$ \\
                & Atom chemical types & $N \times 10$ \\
                & Residue's amino acid type & $N \times 21$ \\
                & Edge type & $E \times 1$ \\
                & Atom coordinates & $N \times 3$ \\
  \midrule
  Surface graph & Vertex coordinates & $M \times 3$ \\
                & Vertex normal vectors & $M \times 3$ \\
  \bottomrule
  \end{tabular}
  \end{table}

\cref{tab:glinter-spatial-graph-features} summarizes all the features. The geometric
features of a residue node include its coordinates and a local reference frame derived from the N-CA-C plane. It uses the CA-C bond as the x-axis, the
vector perpendicular to the plane formed by the N-CA and CA-C
bonds as the z-axis, and their cross-product as the y-axis. Such a representation is rotation invariant and thus, may generalize well without data augmentation in contrast to the network that is not
rotation invariant. The other features of a residue node include position-specific scoring matrix (PSSM), residue solvent accessible surface areas (summation of the solvent accessible surface areas of all
atoms in the residue), the one-hot encoding of amino acid type, and
the sequence index of the residue divided by the protein sequence
length (which is used to provide order information for neural network architectures that are order invariant) \citep{jing2021fast}.

In an atom graph, an edge has a binary feature called ‘edge type’.
It is equal to 1 if the nodes of this edge belong to the same residue.
An atom is encoded by a 10-dimensional 1-hot vector, indicating
four backbone atom types (CA, N, C, O) and six side chain atom
types (CB, C, N, O, S, H).
In a surface graph, we use the coordinates and normals generated
by MSMS as the features of a triangle vertex \citep{gainza2020deciphering},
which indicate the contour and orientation of some local patches on
the surfaces. Normals are initially computed by MSMS, then validated by trimesh’s default protocol.

\paragraph{Coevolution signals generated by Facebook’s MSA transformer}
We use the row attention weights generated by the MSA Transformer
as interfacial co-evolution signals. We build a joint MSA for a heterodimer using the protocol proposed by ComplexContact \citep{zeng2018complexcontact}. For a homodimer, we simply concatenate each sequence in the
MSA with itself. We then select a diverse set of sequences from the joint
MSA as the input of the MSA Transformer. That is, we filter the MSA
with HHfilter \citep{steinegger2019hh} and assign Henikoff weights to
sequences. We further symmetrized the generated inter-chain attentions, following the MSA
Transformer’s protocol \citep{rao2021msa}.

\subsection{Datasets}
Following DeepHomo \citep{yan2021accurate}, we say there is one
true contact between two residues (of two monomers) if in the experimental complex structure, the minimal distance between their
respective heavy atoms is less than 8 \AA . We define the interfacial contact density of a given dimer by $N/(L_1L_2$), where $N$ is the number of
inter-protein contacts and $L_1$ and $L_2$ are the respective lengths of the
constituent monomers.

\paragraph{CASP-CAPRI data}
We use all 32 dimers (23 homodimers and 9 heterodimers) with at
most 1000 residues in the 13th and 14th CASP-CAPRI datasets 40
as our test set. We do not include the dimers with more than 1000
residues since Facebook’s MSA Transformer cannot handle such a
large protein. To avoid redundancy between our training and test
sets and to fairly compare GLINTER with recently published methods, we do not use the 11th and 12th CASP-CAPRI data. We run
HHblits on the 'uniclust30\_2016\_09'
database to build MSAs for
individual chains and then concatenate two MSAs to form a joint
MSA for a heterodimer using the method described in
ComplexContact \citep{zeng2018complexcontact}. We use monomer (bound) experimental structures as inputs since their unbound structures are
unavailable. We also tested the 3D structure models of individual
chains predicted by AlphaFold \citep{jumper2020high,senior2020improved} in CASP13 and 14, except for T0974s2 which did not have a
predicted 3D model. The median interfacial contact density of this
dataset is 1.79\%. Calculated by FreeSASA \citep{mitternacht2016freesasa}, the
median buried solvent accessible surface area (SASA) of this dataset
is 2507\AA
.
\paragraph{3D complex data}

Our training set has 5306 homodimers and 1036 heterodimers
derived from 3DComplex \citep{levy20063d}. We do not include the
dimers with more than 1000 residues due to MSA Transformer’s
limit \citep{rao2021msa}. We say two dimers are at most $x\%$ similar,
if the maximum sequence identity between their constituent monomers is no more than $x\%$ and build a joint MSA as described in the
previous subsection.
The median interfacial contact density of the training set is
0.76\%. The median buried SASA of the training set is 2393 \AA.

\paragraph{PDB2018 data}
We build two more test sets from the complexes released to PDB
after January 1, 2018. One test set (denoted as ‘HomoPDB2018’)
has 165 homodimers and the other one (denoted as
‘HeteroPDB2018’) has 72 heterodimers. We define homodimers and
heterodimers in the same way as the 3DComplex data. We exclude
dimers similar to the training set, judged by MMseqs2 E-value < 1.
We cluster dimers using the 40\% sequence identity threshold and
also remove dimers with interfacial contact density $< 0.7\%$, which
is slightly lower than the median interfacial contact density of the
training set. The medians of the buried SASAs of ‘HomoPDB2018’
and ‘HeteroPDB2018’ are 2557 and 2346 \AA$^2$, respectively. The
medians of the interfacial contact densities of ‘HomoPDB2018’ and
‘HeteroPDB2018’ are 2.41\% and 3.52\%, respectively. It should be
noted that although we remove dimers similar to our training set,
there may be some redundancy between our test dimers and the
training sets used by the other competing methods. Therefore, the
estimated performance of the competing methods on the PDB2018
data may be overly optimistic.

\subsection{Training and evaluation}
We use weighted cross-entropy as the loss function since the interfacial contact density is very small (the median of the training set is 0.76\%). We initially trained our network on a small training subset
using weights 5, 10, 50 and 100 and found that the weight 5 yields
the best average top-10 precision in the first few epochs. So in the
formal training, we set the weight of a contact to be five times that
of a non-contact. We trained our deep models using Adam as the
optimizer \citep{kingma2014adam}, with the hyperparameters
$\beta_1=0.9,\beta_2=0.9999,\epsilon=10^{-8}$. The learning rate is initialized to
$0.0001$ and reduced by half every four epochs. All models are
trained for 20 epochs on two Titan X GPUs, with minibatch size 1
on each GPU. It takes 20–40 min to train one epoch. For a given
hyperparameter setting, we select the model with the best top-10
precision on the validation data as the final model.

Since our deep network is rotation invariant, we do not augment
the training set by rotating a monomer multiple times. Nevertheless,
we randomly rotate a monomer once before training to prevent our
deep network from learning unexpected artifacts in the dataset. For
a heterodimer, we use both of the orders of its two proteins in training. For evaluation, we predict two contact maps for one heterodimer by exchanging the order of its two proteins, and then
compute the geometric average of the two predicted contact map
probability matrices as the final prediction. We evaluate contact prediction in terms of top $k$ precision where
$k \in [10,25,50,L/10]$ and $L/5$ and $L$ is the length of the shorter protein
in a dimer. When the number of native contacts is less than $k$, we
still use $k$ as the denominator while computing the top $k$ precision.
Inter-chain contact maps are more sparse than intra-chain contact
maps, so we evaluate a smaller number of predicted inter-chain
contacts.

\subsection{Methods to compare}\label{sec:glinter-methods-compare}
We compare GLINTER with DeepHomo, ComplexContact and
BIPSPI. DeepHomo is a ResNet-based method developed for only
homodimers. ComplexContact is a sequence-only and ResNet-based
method developed mainly for heterodimers. Both DeepHomo and
ComplexContact take as input the coevolution information computed by CCMpred \citep{seemayer2014ccmpred} while GLINTER does
not. BIPSPI works for both homodimers and heterodimers and can
take both structures and MSAs as input.

\section{Results}
We test our method with the bound experimental structures while
comparing it with BIPSPI, DeepHomo and ComplexContact, as
mentioned in \cref{sec:glinter-methods-compare}. We also study the impact of the quality of
predicted structures on our method.

\subsection{Evaluation of interfacial contact prediction}

  \begin{table}[t]
  \centering
  \caption{Average contact prediction precision (\%) on the CASP-CAPRI and PDB data}
  \label{tab:contact-prediction}
  \small  
  \setlength{\tabcolsep}{4pt}  
  \begin{tabular}{l ccc ccc ccc ccc}
  \toprule
  & \multicolumn{3}{c}{HomoCASP} & \multicolumn{3}{c}{HeteroCASP} & \multicolumn{3}{c}{HomoPDB} & \multicolumn{3}{c}{HeteroPDB} \\
  \cmidrule(lr){2-4} \cmidrule(lr){5-7} \cmidrule(lr){8-10} \cmidrule(lr){11-13}
  \# top & 10 & L/10 & L/5 & 10 & L/10 & L/5 & 10 & L/10 & L/5 & 10 & L/10 & L/5 \\
  \midrule
  BIPSPI & 16 & 16 & 14 & 11 & 11 & 14 & 20 & 21 & 19 & 18 & 18 & 19 \\
  DH & 30 & 27 & 23 & -- & -- & -- & 24 & 25 & 24 & -- & -- & -- \\
  CC & -- & -- & -- & 14 & 14 & 11 & -- & -- & -- & 14 & 13 & 14 \\
  GLINTER & \textbf{54} & \textbf{51} & \textbf{47} & \textbf{44} & \textbf{48} & \textbf{37} & \textbf{48} & \textbf{48} & \textbf{47} & \textbf{47} & \textbf{47} & \textbf{46} \\
  GLINTER$^*$ & 43 & 40 & 37 & 24 & 30 & 23 & -- & -- & -- & -- & -- & -- \\
  \bottomrule
  \end{tabular}

  \vspace{0.5em}
  \begin{flushleft}
  \footnotesize
  All values use native structures except GLINTER$^*$ which uses AlphaFold-predicted structures. `DH' = DeepHomo, `CC' = ComplexContact. HomoCASP: 23 homodimers; HeteroCASP: 9 heterodimers from CASP-CAPRI. HomoPDB and HeteroPDB from PDB2018
  test set. Bold indicates best performance.
  \end{flushleft}
  \end{table}

\paragraph{Performance on the CASP-CAPRI data} 
As shown in \cref{tab:contact-prediction}, tested on the 23 test homodimers, GLINTER
has 54\% top 10 precision and 51\% top $L/10$ precision where $L$ is
the sum of the two monomer protein sequence lengths, while
DeepHomo has 30\% top 10 precision and 27\% top $L/10$ precision.
Tested on the nine heterodimers, GLINTER has 44\% top 10 precision and 48\% top $L/10$ precision, while ComplexContact has 14\%
top 10 precision and 14\% top $L/10$ precision. Even using the monomer structures predicted by AlphaFold-1 and AlphaFold-2 as input,
GLINTER has 43\% top 10 precision on the homodimers and 24\%
top 10 precision on the heterodimers.

\paragraph{Performance on the PDB2018 data} 
 \begin{table}[!t]
  \centering
  \caption{Average contact prediction precision (\%) on the HomoPDB2018 test set, which includes 165 homodimers released to PDB after January 1, 2018.}
  \label{tab:glinter-homopdb-results}
  \begin{tabular}{lccccc}
  \toprule
  \# top predictions & 10 & 25 & 50 & L/10 & L/5 \\
  \midrule
  BIPSPI (native) & 19.82 & 19.15 & 18.70 & 20.89 & 19.09 \\
  DeepHomo (native) & 24.48 & 24.48 & 23.10 & 25.50 & 24.41 \\
  GLINTER (native) & \textbf{47.94} & \textbf{45.94} & \textbf{43.49} & \textbf{47.78} & \textbf{46.69} \\
  \bottomrule
  \end{tabular}

  \vspace{0.5em}
  \begin{flushleft}
  \small
  \textbf{Note:} All methods use experimental (native) monomer structures.
  \end{flushleft}
  \end{table}
\begin{table}[!t]
  \centering
  \caption{Average contact prediction precision (\%) on the HeteroPDB2018 test set, which includes 72 heterodimers released to PDB after January 1, 2018.}
  \label{tab:glinter-heteropdb-results}
  \begin{tabular}{lccccc}
  \toprule
  \# top predictions & 10 & 25 & 50 & L/10 & L/5 \\
  \midrule
  BIPSPI (native) & 18.33 & 18.17 & 18.06 & 18.30 & 19.07 \\
  ComplexContact (native) & 14.31 & 18.22 & 21.27 & 13.13 & 13.95 \\
  GLINTER (native) & \textbf{46.81} & \textbf{41.67} & \textbf{37.44} & \textbf{47.16} & \textbf{46.04} \\
  \bottomrule
  \end{tabular}
  \end{table}

As shown in \cref{tab:contact-prediction}, tested on the 165 HomoPDB2018 homodimers, GLINTER has 48\% top 10 precision, while BIPSPI and
DeepHomo have 20 and 24\% top 10 precision, respectively. Tested
on the 72 HeteroPDB2018 targets, GLINTER has 47\% top 10 precision, while BIPSPI and ComplexContact have 18 and 14\% top 10
precision, respectively. See detailed results in \cref{tab:glinter-homopdb-results,tab:glinter-heteropdb-results}. \\

In summary, GLINTER consistently outperforms DeepHomo
and ComplexContact by a large margin no matter which test sets
are evaluated and whether experimental or predicted monomer
structures are used. \\

\subsection{Ablation study}
\begin{table}[!t]
  \centering
  \caption{Average interfacial contact precision (\%) of different deep learning models on the CASP-CAPRI data when experimental monomer structures are used.}
  \label{tab:casp-model-comparison}
  \small
  \begin{tabular}{llccccc}
  \toprule
  Model & D-cut & 10 & 25 & 50 & L/10 & L/5 \\
  \midrule
  ESM-Attention & -- & 30.93 & 26.50 & 23.69 & 29.38 & 28.02 \\
  CNN+ESM-Attention & -- & 35.31 & 28.25 & 20.12 & 24.31 & 33.98 \\
  \midrule
  Residue & 6 & 28.13 & 24.50 & 20.94 & 26.82 & 23.57 \\
          & 8 & 30.63 & 26.25 & 22.94 & 30.06 & 27.01 \\
          & 10 & 29.69 & 26.13 & 21.38 & 28.26 & 25.00 \\
  \midrule
  Residue+ESM & 8 & 42.81 & 37.13 & 34.25 & 41.80 & 37.12 \\
  \midrule
  Residue+Atom & 8,4 & 21.56 & 19.13 & 18.75 & 22.58 & 18.82 \\
               & 8,6 & 32.50 & 28.50 & 23.06 & 31.85 & 27.39 \\
               & 8,8 & 26.56 & 25.00 & 22.69 & 27.46 & 25.65 \\
               & 8,10 & 29.69 & 25.50 & 22.19 & 26.55 & 26.75 \\
  \midrule
  Residue+Atom+ESM & 8,6 & 39.06 & 34.00 & 31.88 & 36.26 & 33.84 \\
                   & 8,8 & 40.60 & 36.00 & 32.06 & 39.00 & 35.47 \\
  \midrule
  Residue+Surface & 8,4 & 33.13 & 26.75 & 23.81 & 31.18 & 27.04 \\
                  & 8,6 & 33.75 & 28.25 & 25.63 & 33.69 & 27.58 \\
                  & 8,8 & 33.43 & 29.25 & 25.81 & 33.34 & 29.37 \\
                  & 8,10 & 32.50 & 29.25 & 26.19 & 32.36 & 27.90 \\
  \midrule
  Residue+Surface+ESM & 8,6 & 44.06 & 37.88 & 33.88 & 41.38 & 37.92 \\
                      & 8,8 & 39.69 & 34.50 & 31.31 & 36.95 & 36.11 \\
  \midrule
  Residue+Atom+Surface & 8,6,6 & 39.69 & 31.75 & 27.81 & 35.97 & 31.53 \\
  \midrule
  Residue+Atom+Surface+ESM & 8,6,6 & \textbf{51.56} & \textbf{44.63} & \textbf{38.00} & \textbf{50.36} & \textbf{44.43} \\
  \bottomrule
  \end{tabular}

  \vspace{0.5em}
  \begin{flushleft}
  \small
  \textbf{Note:} Column ``D-cut'' shows the distance cutoffs used to define graph edges (in \AA). For example, ``8,6,6'' indicates that the residue graph, atom graph and surface graph use 8~\AA, 6~\AA, and 6~\AA{} to define edges,
  respectively.
  \end{flushleft}
  \end{table}
\begin{table}[!t]
  \centering
  \caption{Average interfacial contact precision (\%) of different deep learning models on the CASP-CAPRI data when monomer structures are predicted by AlphaFold.}
  \label{tab:model-comparison-alphafold}
  \small
  \begin{tabular}{llccccc}
  \toprule
  Model & D-cut & 10 & 25 & 50 & L/10 & L/5 \\
  \midrule
  ESM-Attention & -- & 29.06 & 25.63 & 22.94 & 27.26 & 26.97 \\
  CNN+ESM-Attention & -- & 21.56 & 19.50 & 17.75 & 19.59 & 18.67 \\
  \midrule
  Residue & 6 & 21.56 & 18.75 & 16.94 & 22.03 & 18.68 \\
          & 8 & 26.56 & 23.62 & 19.50 & 25.93 & 22.07 \\
          & 10 & 22.50 & 20.13 & 18.00 & 23.17 & 20.75 \\
  \midrule
  Residue+ESM & 8 & 33.44 & 30.00 & 29.06 & 33.21 & 29.69 \\
  \midrule
  Residue+Atom & 8,4 & 23.44 & 21.50 & 18.06 & 24.45 & 19.66 \\
               & 8,6 & 26.25 & 21.25 & 17.69 & 23.91 & 20.07 \\
               & 8,8 & 26.25 & 23.63 & 18.56 & 28.72 & 22.70 \\
               & 8,10 & 25.31 & 23.13 & 19.44 & 25.97 & 23.17 \\
  \midrule
  Residue+Atom+ESM & 8,6 & 35.31 & 30.25 & 27.94 & 31.86 & 28.98 \\
                   & 8,8 & 30.62 & 28.25 & 26.63 & 30.21 & 27.13 \\
  \midrule
  Residue+Surface & 8,4 & 24.06 & 19.75 & 17.19 & 23.22 & 20.54 \\
                  & 8,6 & 22.19 & 19.75 & 18.81 & 21.27 & 17.45 \\
                  & 8,8 & 27.81 & 23.88 & 20.18 & 25.19 & 22.45 \\
                  & 8,10 & 27.50 & 24.88 & 21.19 & 26.52 & 24.48 \\
  \midrule
  Residue+Surface+ESM & 8,6 & 36.56 & 32.75 & 29.75 & 34.51 & 31.81 \\
                      & 8,8 & 34.69 & 30.50 & 28.25 & 32.87 & 30.24 \\
  \midrule
  Residue+Atom+Surface & 8,6,6 & 29.38 & 23.88 & 20.25 & 27.07 & 24.06 \\
  \midrule
  Residue+Atom+Surface+ESM & 8,6,6 & \textbf{37.81} & \textbf{34.38} & \textbf{31.31} & \textbf{37.23} & \textbf{32.94} \\
  \bottomrule
  \end{tabular}

  \vspace{0.5em}
  \begin{flushleft}
  \small
  \textbf{Note:} Column ``D-cut'' shows the distance cutoffs used to define graph edges (in \AA). For example, ``8,6,6'' indicates that the residue graph, atom graph and surface graph use 8~\AA, 6~\AA, and 6~\AA{} to define edges,
  respectively. The performance of the ESM-Attention model in this table differs from the native structure results because the predicted and experimental monomer structures do not have exactly the same set of residues.
  \end{flushleft}
  \end{table}
\begin{figure}[!ht]
    \centering
    \includegraphics[width=0.8\textwidth,height=0.6\textheight,keepaspectratio]{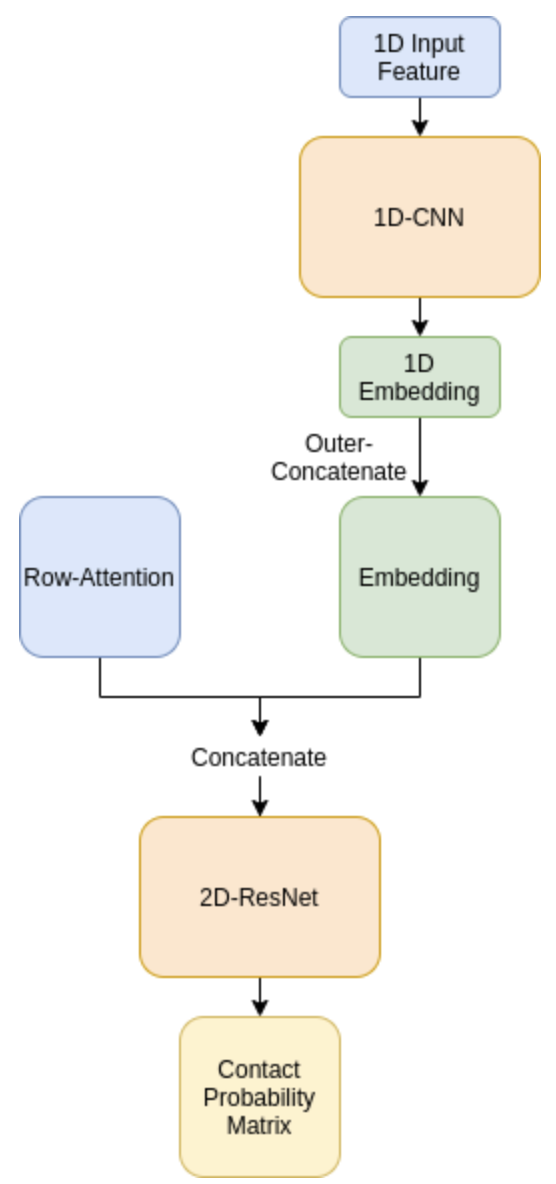}
    \caption{CNN+ESM-Attention model}
    \label{fig:glinter-sfig2}
\end{figure}

We train the GLINTER models under eight different settings (different sets of input features). \cref{tab:casp-model-comparison} and \cref{tab:model-comparison-alphafold} show
their test results with monomer experimental structures and
AlphaFold-predicted monomer structures, respectively. We have
studied the following eight settings:

``Residue'', ``Residue+ESM'',
``Residue+Atom'', ``Residue+Atom+ESM'', ``Residue+Surface'',
``Residue+Surface+ESM'', ``Residue+Atom+Surface'' and ``Residue+Atom+Surface+ESM'' models. Here, ``Residue'', ``Atom''
and ``Surface'' represent the residue, atom and surface graphs, respectively. ``ESM'' means that the ESM row attention weights are
used. Using the ESM row attention weights does not change the network architecture, but increases the input dimension of the first
ResNet block, as shown in \cref{fig:glinter-arch}.

To evaluate the contribution of the ESM row attention weights,
we test a sequence-only model called ``ESM-Attention'' that uses only
the ESM row attention weights as input. Its major module is a 2D ResNet with the
same architecture as the one used in the Residue+ESM model.
To evaluate the contribution of the graph convolution module,
we develop a sequence-structure-hybrid model denoted as
``CNN+ESM-Attention'', which uses an 1D convolutional network
(CNN) and the same set of input features. Similar to the
Residue+ESM model, the CNN+ESM-Attention model consists of
two major modules: a Siamese 1D CNN and a ResNet. The 1D
CNN has four convolution layers (each with 128 filters and kernel
size 5) and the ResNet is the same as that used in the Residue+ESM
model (\cref{fig:glinter-sfig2}). Both the ESM-Attention and the
CNN+ESM-Attention models are trained on the same dataset using
the same protocols as the GLINTER models. \\

\paragraph{Contribution of the graph convolution module}

  \begin{table}[!t]
  \centering
  \caption{Average interfacial contact precision (\%) of the ESM-Attention, CNN+ESM-Attention and Residue+ESM models on the CASP-CAPRI data}
  \label{tab:esm-comparison}
  \begin{tabular}{lccccc}
  \toprule
  No. of top predictions & 10 & 25 & 50 & L/10 & L/5 \\
  \midrule
  ESM-Attention & 31 & 27 & 24 & 29 & 28 \\
  CNN+ESM-Attention & 35 & 28 & 20 & 24 & \textbf{34} \\
  Residue+ESM & \textbf{43} & \textbf{37} & \textbf{34} & \textbf{42} & 32 \\
  \bottomrule
  \end{tabular}

  \vspace{0.5em}
  \begin{flushleft}
  \small
  \textbf{Note:} The ESM-Attention model only uses MSAs as inputs, while the CNN+ESM-Attention and Residue+ESM models use MSAs and experimental monomer structures as inputs.
  \end{flushleft}
  \end{table}

As shown in \cref{tab:esm-comparison,tab:casp-model-comparison}, the CNN+ESM-Attention model has similar performance as the ESM-Attention
model. The best CNN+ESM-Attention model has 35\% top-10 precision and 24\% top-L/10 precision, while the ESM-Attention model
has 31\% top-10 precision and 29\% top-$L/10$ precision. In contrast,
the Residue+ESM model has 43\% top-10 precision and 42\% top-$L/10$ precision, which suggests that the residue graph (derived from
monomer structures) used by GLINTER is indeed very helpful for
interfacial contact prediction.
\paragraph{Dependency on distance cutoff}
\begin{table}[t]
  \centering
  \caption{Average top-10 interfacial contact precision (\%) of the `Residue+Atom' and `Residue+Surface' models on the CASP-CAPRI data when experimental monomer structures are used}
  \label{tab:glinter-tab3}
  \begin{tabular}{lcccc}
  \toprule
   & 8,4 & 8,6 & 8,8 & 8,10 \\
  \midrule
  Residue+Atom & 22 & 33 & 27 & 30 \\
  Residue+Surface & \textbf{33} & \textbf{34} & \textbf{33} & \textbf{33} \\
  \bottomrule
  \end{tabular}

  \vspace{0.5em}
  \begin{flushleft}
  \small
  \textbf{Note:} The first row shows the distance cutoffs used to define graph edges. For example, `8,6' for `Residue+Atom' indicates that the residue graph and atom graph use 8 and 6~\AA{} to define edges, respectively.
  \end{flushleft}
  \end{table}
The distance cutoff used to define graph edges is an important
hyperparameter. According to our observation, a model with a larger distance cutoff tends to have a lower training loss, although
its prediction performance may not be as good. A model with a
smaller distance cutoff may have a higher training loss and much
worse prediction performance. As shown in \cref{tab:glinter-tab3,tab:casp-model-comparison}, the top $k$ precision of GLINTER models
increases along with the distance cutoff until reaching the optimal
value. For example, the top-10 precision of the Residue+Atom
model increases from 22 to 33\% as the distance cutoff increases
from 4 to 6 \AA , and then decreases to 27\% when the distance cutoff is
8 \AA . This saturation effect on the distance cutoffs is also observed in
\citep{townshend2019end}.
Different types of graphs may rely on distance cutoffs differently.
For example, the top 10 precision of the Residue+Surface model is
around 33\% when the distance cutoff defining the surface graph
ranges from 4 to 10 \AA , while the precision of the ``Residue+Atom''
model changes a lot with respect to the distance cutoff. Here, we determine the optimal distance cutoff using the experimental monomer
structures, which may not have the optimal performance when predicted monomer structures are used.

\paragraph{Dependency on the quality of predicted monomer structures}
\begin{figure}[!ht]
    \centering
    \includegraphics[width=0.8\textwidth]{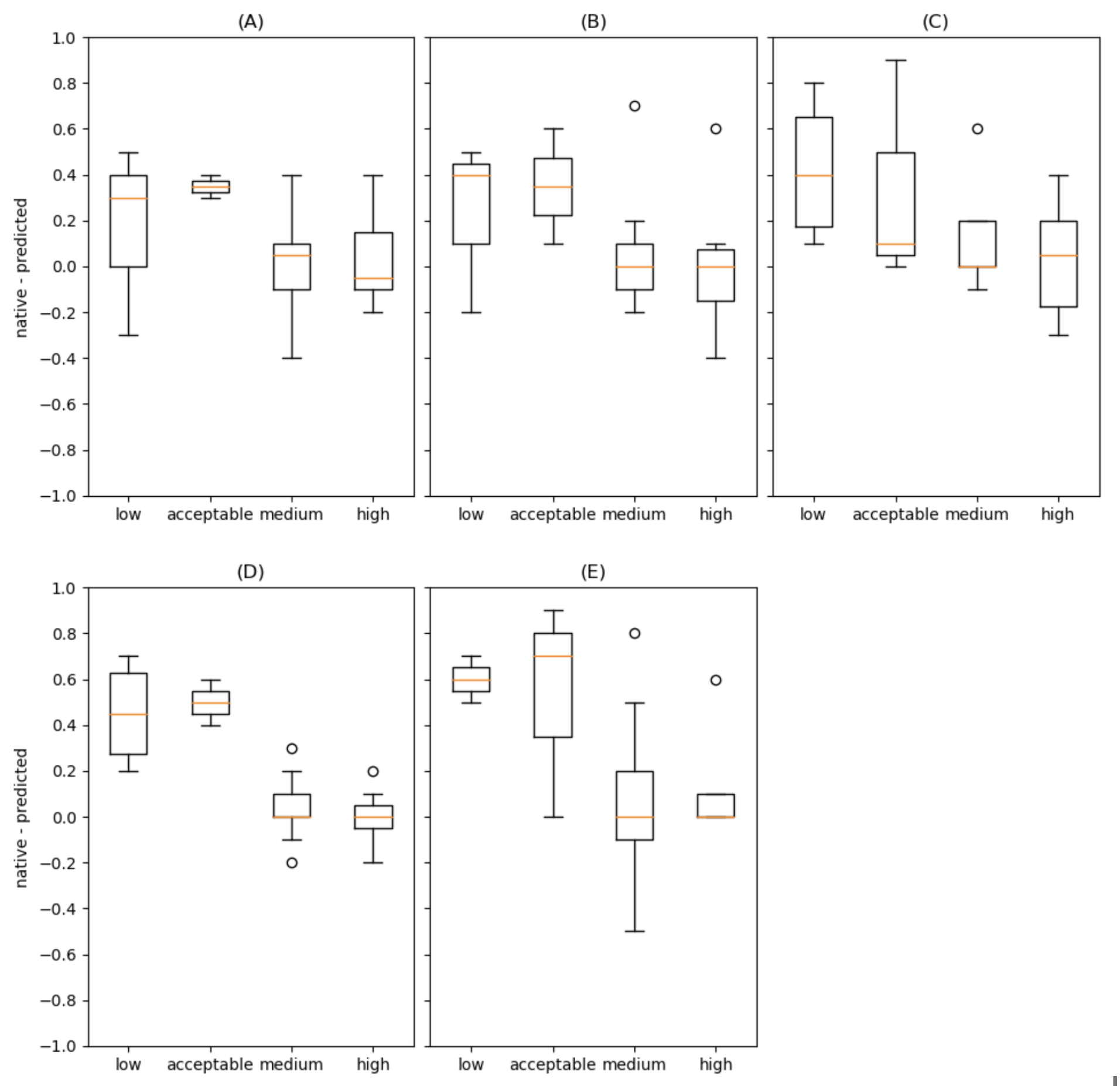}
    \caption{The x-axis is the TMscore of the predicted monomer structures. The y-axis is the difference of the top 10 precision resulting from the experimental and predicted monomer structures. (A) “Residue, D-cut=8”. (B) “Residue + Atom, D-cut=8,6” (C) “Residue + Surface, D-cut=8,6” (D) “Residue + Atom + Surface, D-cut=8,6,6” (E) “Residue + Atom + Surface + ESM, D-cut=8,6,6”. In all the box plots, the upper edge of the box is the third quartile (Q3), and the lower edge of the box is the first quartile (Q1), the orange line is the median, the upper cap is the highest datum below Q3 + 1.5(Q3 - Q1), and the lower cap is the lowest datum above Q1 - 1.5(Q3-Q1). }
    \label{fig:glinter-tmscore-ablation}
\end{figure}
GLINTER models are trained with monomer experimental structures. Here, we study their prediction performance when the
AlphaFold-predicted monomer structures are used. We use the lower
TMscore \citep{zhang2005tm} of the two constituent monomer models to measure the structure quality of a dimer under test.
We exclude the test dimers without any correct top $k$ predicted contacts when their native structures are used as input. Since there are
only dozens of test targets, we divide them into four groups according to their TMscores: low quality ($0.2 \le \text{TMscore} < 0.5$), acceptable quality ($0.5  \le \text{TMscore} < 0.7$), medium quality ($0.7
\le \text{TMscore} < 0.9$) and high quality ($0.9  \le \text{TMscore} < 1.0$).
\cref{fig:glinter-tmscore-ablation} shows that even trained on bound experimental structures, our methods work well on predicted structures with medium or high quality (i.e. $\text{TMscore} > 0.7$). When the
predicted monomer structures have lower quality ($\text{TMscore} < 0.7$),
GLINTER models perform better with experimental structures than
predicted structures. By comparing \cref{fig:glinter-tmscore-ablation}~D and
E, we find that the ESM row attention weight may not be able to reduce the precision gap incurred by predicted structures. This suggests that the ESM row attention weight derived purely from MSAs
may not necessarily improve the robustness of our structure-based
models.

\begin{figure}[!ht]
    \centering
    \includegraphics[width=0.8\textwidth]{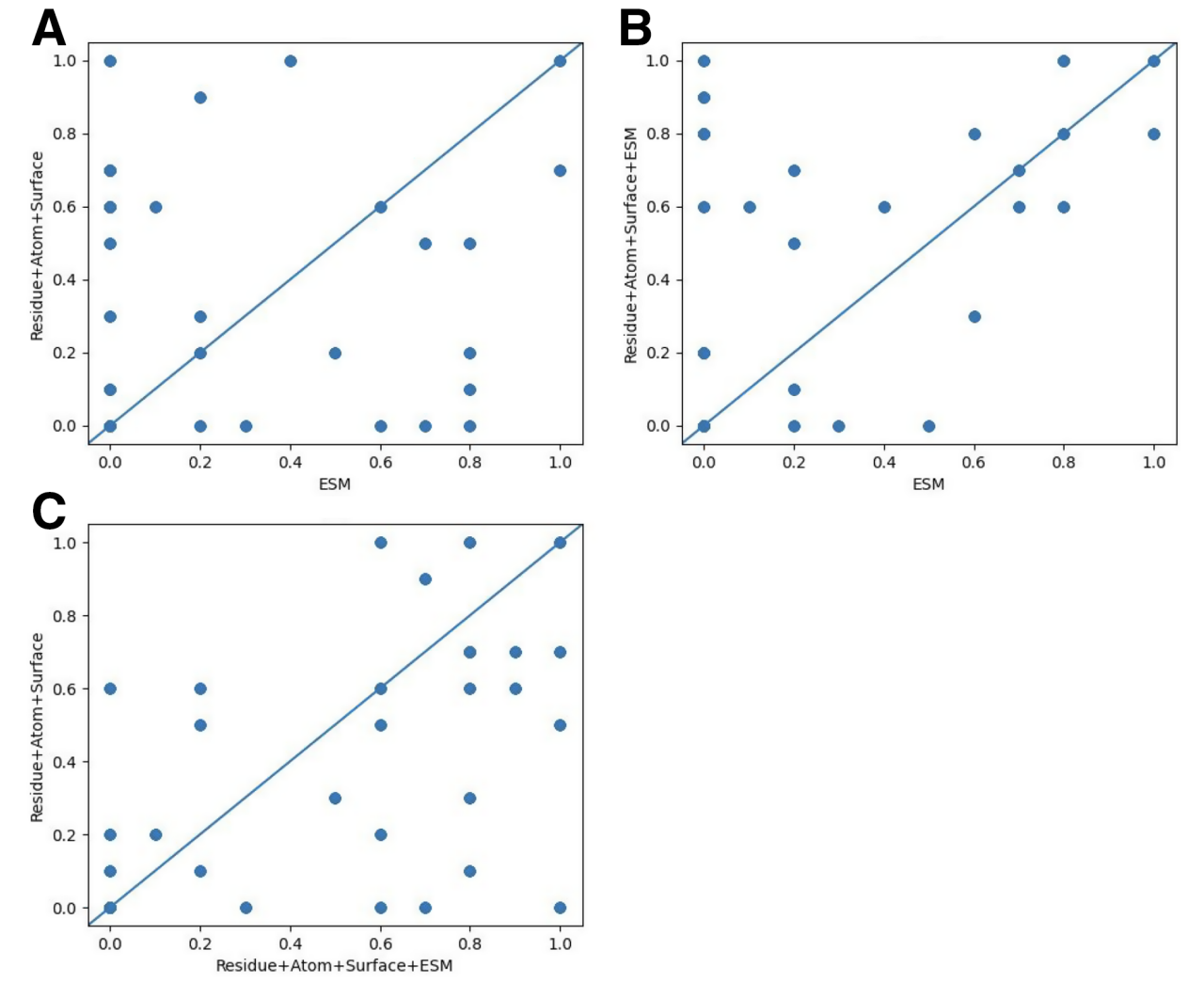}
    \caption{ Comparison of top-10 precision of three models: ESM,
  Residue+Atom+Surface and Residue+Atom+Surface+ESM. (A) compares
  Residue+Atom+Surface and ESM, (B) compares Residue+Atom+Surface+ESM
  and ESM, and (C) compares Residue+Atom+Surface and
  Residue+Atom+Surface+ESM}
    \label{fig:glinter-fig2}
\end{figure}
\paragraph{Contribution of the ESM row attention weight}

As shown in \cref{tab:contact-prediction,tab:casp-model-comparison}, on the 32 dimer targets, the ESM-Attention model has top 10 and $L/10$ precision 31 and
29\%, respectively, greatly outperforming BIPSPI, which has top 10 and
$L/10$ precision 15 and 14\%, respectively. That is, even though the MSA
Transformer is pre-trained with the MSAs of single-chain protein
sequences, it works for inter-chain contact prediction. Over the nine
heterodimer targets, the top 10 precision of ComplexContact and ESMAttention is 14 and 28\%, respectively. As shown in \cref{tab:casp-model-comparison,tab:model-comparison-alphafold}, no matter whether native or predicted monomer
structures are used the ESM row attention weight consistently improves
the performance of GLINTER models, which confirms that coevolution
signals are very useful for inter-chain contact predictions.
\cref{fig:glinter-fig2}~A compares the performance of the ESM-Attention model
(which is a sequence-only model) and the Residue+Atom+Surface model
(which is a structure-only model) when the native structures are used.
They have similar overall performance, but perform very differently on
individual test targets, which suggests that the ESM row attention weight
and structure information are highly complementary to each other. On
the majority of test targets, the Residue+Atom+Surface+ESM model
outperforms the ESM-Attention model (\cref{fig:glinter-fig2}~B) and the
Residue+Atom+Surface model (\cref{fig:glinter-fig2}~C). \cref{fig:glinter-fig2}~A and B differs only
in the y-axis by an ESM feature, so their comparison shows the impact of
the ESM features. \cref{fig:glinter-fig2}~A and C differs only in the x-axis by
Residue+Atom+Surf, so their comparison shows the impact of the
Residue+Atom+Surf features. 


\paragraph{Case study of T0997}
We study an interesting target, T0997, where the ESM-Attention model and the Residue+Atom+Surface model perform much better than the Residue+Atom+Surface+ESM model. In Fig. S5, the cluster of contacts correctly predicted by the ESM-Attention model is different from the cluster correctly predicted by the Residue+Atom+Surface model, so we can hypothesize that the ESM-Attention model (MSA-based) and the Residue+Atom+Surface model (structure-based) focus on very different patterns in T0997. Therefore, it is possible that there are some structure patterns around the cluster correctly predicted by the ESM-Attention model that make the Residue+Atom+Surface+ESM model predictions misaligned with the ground truth contacts; see \cref{tab:model-comparison-alphafold}.

\paragraph{Dependency on the depth of MSAs}
\begin{figure}[!ht]
    \centering
    \includegraphics[width=0.8\textwidth]{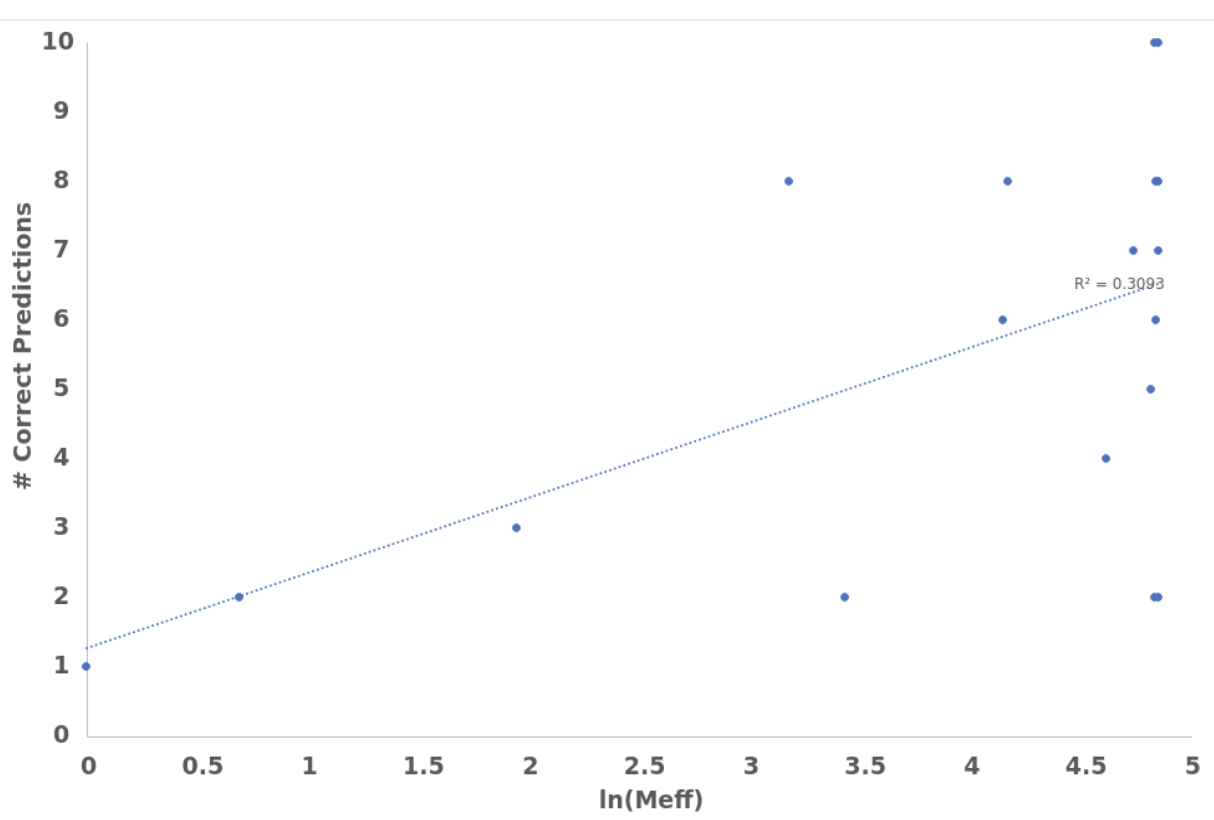}
    \caption{Correlation between $\ln(\text{Meff})$ (x-axis) and the number of correct top-10 predictions (y-axis) of the ESM-Attention model. The targets without correct top-10 predictions are excluded. ($R^2=0.3093$)}
    \label{fig:glinter-msa-ablation}
\end{figure}
It is known that intra-chain contact prediction precision correlates with the depth of MSAs denoted as Meff. Given an MSA, we use CD-HIT to cluster all the sequences in this MSA using 65\% sequence identity as cutoff. The effective MSA depth is  defined as the number of clusters. Here, we study the impact of MSA depth on
interfacial contact prediction when the ESM row attention weight
is used. To remove the impact of inaccurate predicted structures,
here, we test GLINTER models with native monomer structures.
\cref{fig:glinter-msa-ablation} shows that there is certain correlation
($R^2 = 0.3093$) between the number of correct top-10 predictions
by the ESM-Attention model and the $\ln(\text{Meff})$ of the input MSA.

\subsection{Application to selection of docking decoys}
\begin{figure}[!ht]
    \centering
    \includegraphics[width=0.8\textwidth]{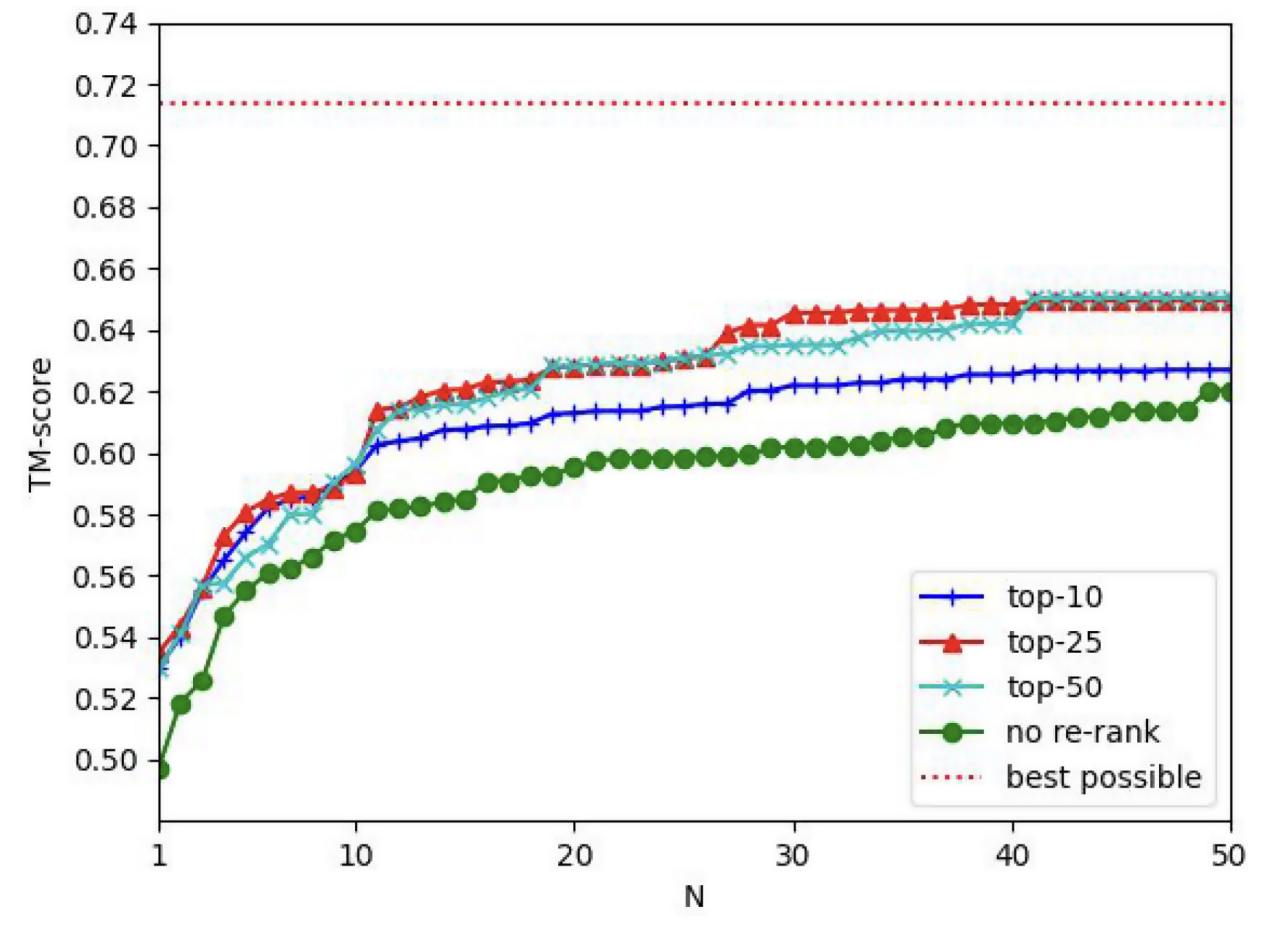}
    \caption{The average quality (measured by TMscore) of the selected decoys by top
predicted contacts. The x-axis is the number of top decoys selected. In the legend,
``top-10'', ``top-25'' and ``top-50'' represent that top 10, 25 and 50 predicted contacts
are used to select docking decoys, respectively. ``best decoy'' indicates the quality of
the best decoys generated by HDOCK}
    \label{fig:glinter-dock}
\end{figure}

A simple application of predicted interfacial contacts is to select the
docking decoys. We use the top $k$ ($k \in [10, 25, 50]$) contacts predicted by the Residue+Atom+Surface+ESM model to rank the docking
decoys generated by HDOCK. The quality of a docking decoy is calculated by comparing it with its experimental complex structure
using MMalign (Mukherjee and Zhang, 2009). For each target, we
select top N decoys ranked by the predicted interfacial contacts and
define their highest TMscore as the ``TMscore of the top N decoys''.
In \cref{fig:glinter-dock}, the y-axis shows the average TMscore of the top N
decoys of all the test dimers. Generally speaking, predicted contacts
may improve the quality of top decoys by 5-8\%. Except when
$N=10$, generally speaking using more top predicted contacts may select better decoys than using only top 10 predicted contacts.

\section{Conclusion}
We have presented an interfacial contact prediction method,
GLINTER, that predicts inter-protein contacts by integrating attention information generated by protein language models and graph
modeling of monomer (experimental and predicted) structures. The
attention may capture evolutionary and coevolutionary information
encoded in MSA. We demonstrate that GLINTER outperforms
existing methods and even if trained with experimental structures, it
generalizes well to predicted structures. The interfacial contacts predicted by our method may help improve selection of docking decoys.
Our ablation study shows that the attention information and structural features are complementary and important for interfacial contact prediction. The features used by GLINTER can be calculated
very efficiently and GLINTER is applicable to both heterodimers
and homodimers. Therefore, potentially GLINTER is applicable to
the proteome-scale study of protein–protein interactions and
complexes.

\bibliographystyle{siam}
\bibliography{references}
\chapter{Improved Protein Heterodimer Structure Prediction with Protein Language Models}

\noindent This chapter presents work with Bo Chen, Jiezhong Qiu, Zhaofeng Ye, Jinbo Xu, and Jie Tang. It was published in \textit{Briefings in Bioinformatics} in 2023~\citep{chen2023improved}. Code is available at
  \url{https://github.com/zw2x/msa_pair}.

\section{Introduction}

While GLINTER demonstrates that graph neural networks can effectively predict interfacial contacts from monomeric structures and co-evolutionary signals, its performance depends on the quality of the input co-evolutionary information. For
  heterodimeric complexes, constructing informative co-evolutionary features requires identifying correct interacting homologs across species — a problem that remains a key bottleneck. In this chapter, we address this challenge directly by
  developing a protein language model-based approach to pair interacting homologs for improved complex structure prediction.

  Deep learning has made substantial progress in protein structure prediction by effectively leveraging evolutionary information~\cite{jumper2021highly,xu2021improved}. These methods utilize the co-evolutionary signals hidden in Multiple Sequence Alignments (MSAs) to infer inter-residue interactions and three-dimensional structures. AlphaFold2~\cite{jumper2021highly} is the representative method, demonstrating unparalleled accuracy in monomer structure prediction. Building on this, AlphaFold-Multimer~\cite{deepmind2021alphafold} extended the capability to protein complexes, significantly outperforming previous systems~\cite{comeau2004cluspro,evans2022protein,yin2022benchmarking}. However, compared to the breakthrough in monomer folding, the accuracy of AlphaFold-Multimer on heterodimer prediction remains limited (success rate $\sim70\%$, mean DockQ $\sim0.6$), leaving substantial room for improvement~\cite{yin2022benchmarking}.
   
   The most important input feature to AlphaFold-Multimer is the multiple sequence
  alignment (MSA)~\cite{evans2022protein,yin2022benchmarking}. Compared with AlphaFold2~\cite{jumper2021highly}, which takes the MSA of a single protein as input, AlphaFold-Multimer needs to build an MSA of interologs for protein complex structure prediction. However, how to construct such an MSA is still
  an open problem for heteromers. It requires the identification of interacting homologs in the MSAs of constituent single chains, which may be challenging since one species may have multiple sequences similar to the target sequence (paralogs). Several algorithms have
  been proposed to identify putative interologs from genome data, such as profiling co-evolved genes~\cite{pellegrini1999profiling} and comparing phylogenetic trees~\cite{pazos2001comparing}. Genome co-localization and species information are two commonly used heuristics to form interologs for
  co-evolution-based complex contact and structure prediction~\cite{deepmind2021alphafold,zeng2018complexcontact}. Genome co-localization is based on the observation that, in bacteria, many interacting genes are coded in operons~\cite{salgado2000operons,dam2007operons} and are co-transcribed to perform their functions.
  However, this rule does not perform well for complexes in eukaryotes with a large number of paralogs, since it becomes more difficult to disambiguate correct interologs~\cite{zeng2018complexcontact,bitbol2016inferring}. The other phylogeny-based method for identifying interologs was first proposed in
   ComplexContact~\cite{zeng2018complexcontact} and later similar ideas were adopted by AlphaFold-Multimer. This method first identifies groups of paralogs (sequences of the same species) from the MSA of each chain, then ranks the paralogs based on their sequence similarity to their
  corresponding primary chain and finally pairs sequences of the same species and with the same rank together. However, these are all hand-crafted approaches which merely take effect on specific domains. In this paper, we instead investigate general and automatic
  algorithms for constructing MSAs of interologs for heterodimers effectively.

  Protein language models (PLMs)~\cite{rives2021biological,rao2021msa,elnaggar2020prottrans} have emerged as a powerful paradigm for protein representation learning, benefiting tasks such as contact prediction and mutation effect prediction~\cite{rao2020esm}. By capturing biological constraints and co-evolutionary information from vast sequence databases, PLMs offer a distinct advantage over traditional methods. This raises a natural question: \emph{Can we leverage the co-evolutionary information captured by PLMs to build effective interologs?}

  In this chapter, we focus on heterodimer structure prediction. We propose ESMPair, a simple yet effective MSA pairing algorithm that leverages column-wise attention scores from ESM-MSA-1b~\cite{rao2021msa} to identify and pair homologs. Extensive experiments on three test sets (pConf70, pConf80, and DockQ49) demonstrate that ESMPair achieves state-of-the-art accuracy, outperforming AlphaFold-Multimer on heterodimer prediction ($+10.7\%$, $+7.3\%$, and $+3.7\%$ in Top-5 Best DockQ score, respectively). We also find that ensemble strategies combining ESMPair with other methods further improve the prediction accuracy. Notably, ESMPair excels on eukaryotic targets and cross-kingdom pairs (eukaryote-bacteria), answering the challenge of identifying interologs in these difficult cases. Furthermore, we show that the diversity of interologs correlates positively with prediction accuracy. Overall, ESMPair effectively incorporates the strength of PLMs to address the challenge of MSA pairing for heterodimer prediction.
\section{Data and Methods}
\begin{figure}[ht!]
    \centering
    \includegraphics[width=\textwidth]{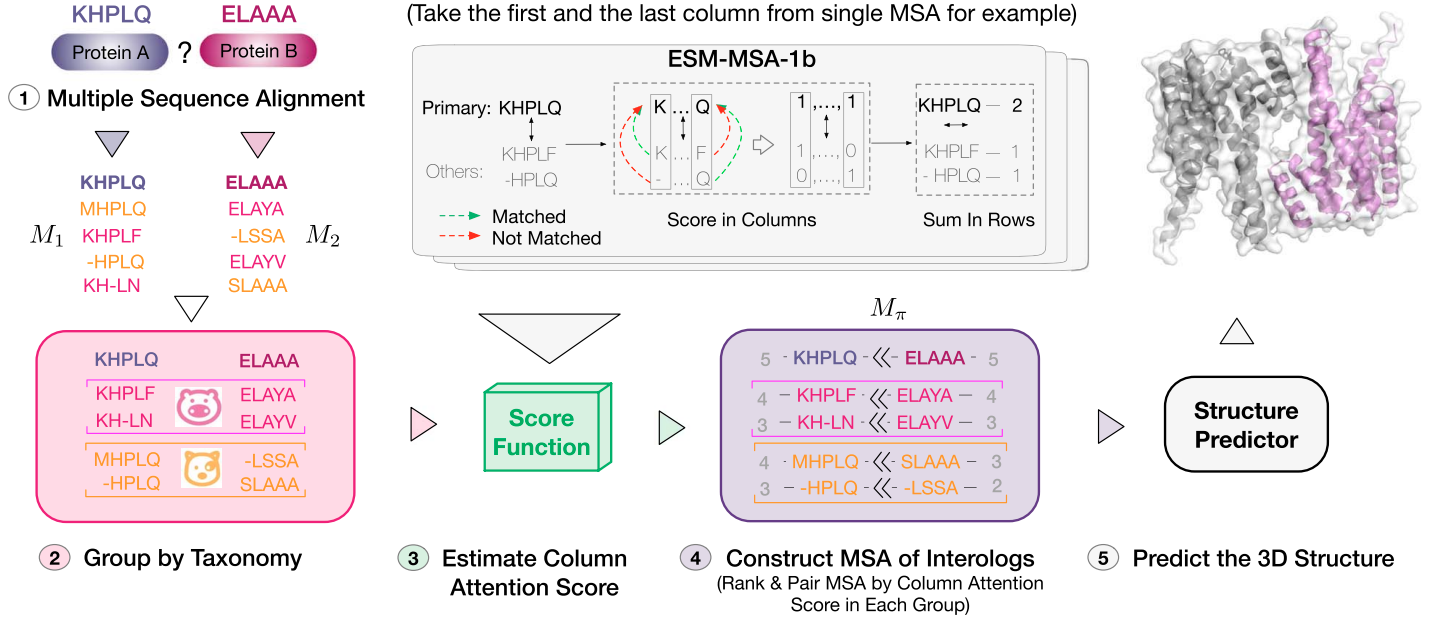}
    \caption{Schematic illustration of ESMPair. Given a pair of query sequences: (1) JackHMMER searches UniProt~\cite{uniprot2023} to generate an MSA for each query, (2) homologs are grouped by species, (3) ESM-MSA-1b estimates column attention
   scores between each homolog and the query, (4) homologs from the same species with the same rank are paired and concatenated into interologs, and (5) AlphaFold-Multimer takes the interolog MSA as input to predict the complex
   structure.}
    \label{fig:esmpair_overview}
  \end{figure}
  We introduce the framework of our proposed PLM-enhanced MSA pairing method, i.e.\ ESMPair. The overall framework of ESMPair
  is illustrated in~\Cref{fig:esmpair_overview}.

  \subsection{The PLM-enhanced MSA pairing pipeline}

  Previous works~\cite{rives2021biological,rao2021msa,elnaggar2020prottrans}
  have confirmed that PLMs can capture the co-evolutionary and inter-residue
  contact signals encoded in protein sequences. Moreover, MSA-based
  PLMs~\cite{rao2021msa} further offer explicit axial attention mechanisms to
  extract evolutionary information from
  MSAs~\cite{wang2020axial,seemayer2014ccmpred}. In light of this, we adopt
  ESM-MSA-1b~\cite{rao2021msa} to explore building MSAs of interologs to improve
   PCP based on AlphaFold-Multimer~\cite{deepmind2021alphafold}.

  \paragraph{Column attention (ESMPair).}
  The column attention weight matrix, calculated by ESM-MSA-1b, measures pairwise
  similarities between aligned residues in each column. Formally, for each chain, we have the MSA $M \in \mathcal{A}^{N \times
  C}$, where $\mathcal{A}$ denotes the set of the amino acid types, $N$ and $C$ represent the number of sequences and residues respectively. The collections of column attention matrices are denoted as
  \begin{equation}
  \{A_{lhc} \in \mathbb{R}^{N \times N} : l \in [L], h \in [H], c \in [C]\},
  \end{equation}
  where $L$ is the number of layers in PLM and $H$ is the number of attention heads of each layer. We first symmetrize each column attention matrix, and then aggregate the symmetrized matrices along the dimensions of $L$, $H$ and $C$ to obtain the pairwise similarity
  matrix among the sequences of MSA, denoted as $S \in \mathbb{R}^{N \times N}$:
  \begin{equation}
  S = \underset{l \in [L], h \in [H], c \in [C]}{\text{AGG}} \left( A_{lhc} + A_{lhc}^\top \right),
  \end{equation}
  where $\text{AGG}$ is an entry-wise aggregation operator such as entry-wise mean operation $\text{MEAN}(\cdot)$, sum operator $\text{SUM}(\cdot)$, etc. Unless otherwise specified, $\text{AGG}$ is specified as $\text{SUM}(\cdot)$ in this paper.

  $S$ is symmetric and its first row, $S_1$, measures the similarity between the
   query sequence and its hit sequences in the MSA. The MSA pairing strategy is
  as follows: for each constituent chain of the query heterodimer, we group hits
   from the MSA by their species, and, within each group, rank sequences
  according to their similarity score in $S_1$. Finally, the sequences of each
  MSA with the same rank from the same species group are concatenated to form a
  paired MSA.



  \subsection{Settings}

  \paragraph{Evaluation metric.}
  We evaluate the accuracy of predicted complex structures using DockQ~\citep{basu2016dockq}, a widely-used metric in the computational structural biology community. Specifically, for each protein complex target, we calculate the highest DockQ score among its top-$N$ predicted
  models selected by their predicted confidences from AlphaFold-Multimer. We refer to this metric as the best DockQ among top-$N$ predictions. We also report other metrics in some experiments like iRMS, TMscore, ICS and IPS, oligomer-LDDT and QS-global.

  \paragraph{Datasets.}
  In order to investigate how improving pairing MSAs can improve the performance of AlphaFold-Multimer, we construct a test set satisfying the following criteria:
  \begin{enumerate}[label={(\roman*)}]
      \item There are at least 100 sequences that can be paired given the species constraints.
      \item The two constituent chains of a heterodimeric target share $<90\%$ sequence identity.
  \end{enumerate}

  We select heterodimers consisting of chains with 20--1024 residues (due to the constraint of ESM-MSA-1b and also to ignore peptide-protein complexes), and the overall number of residues in a dimer is less than 1600 (due to GPU memory constraint) from Protein Data Bank
  (PDB), as accessed on 3 March 2022. We use the default AlphaFold-Multimer MSA search setting to search the UniProt database~\citep{uniprot2021} with JackHMMER~\citep{potter2018hmmer}, which is used for MSA pairing. We also search the Uniclust30 database~\citep{mirdita2017uniclust} with
  HHblits~\citep{steinegger2019hh}, which is used for monomers, i.e.\ block diagonal pairing. We further select those heterodimers with at least 100 sequences that can be paired by AlphaFold-Multimer's default pairing strategy.

  We define two dimers as at most $x\%$ similar if the maximum sequence identity between their constituent monomers is no more than $x\%$. Overall, we select 801 heterodimeric targets from PDB that are at most 40\% similar to any other targets in the dataset and satisfy
  the aforementioned two criteria. Then we use AlphaFold-Multimer (using the default MSA matching algorithm) to predict their complex structures. Based on their predicted confidence scores (pConf) or DockQ scores, 92 targets with their pConf less than 0.7 are denoted as
  the pConf70 test set. We select 0.7 as the low confidence cutoff based on our fitted logistic regression models over 7000 DockQ and pConf pairs, because the conditional probability of the model having medium or better quality given pConf equals 0.7 is slightly greater
  than 0.5 (around 0.6), while the probability is less than 0.5 if pConf equals 0.6. For more comparisons, we also select 0.8 as the cutoff, which results in the pConf80 test set of 168 targets, and 155 targets with their predicted DockQ scores less than 0.49 are denoted
  as the DockQ49 test set.

  \subsection{Baselines}

  Several heuristic MSA pairing strategies have been developed for protein complex contact and 3D structure prediction~\citep{baek2021accurate,evans2022protein}.

  \paragraph{Phylogeny-based method.}
  This strategy was first proposed in ComplexContact~\citep{zeng2018complexcontact} for complex contact prediction and is widely adopted by the community. AlphaFold-Multimer employed a similar strategy. This strategy first groups sequences in an MSA by their species and then ranks
  sequences of the same species by their similarity to the query sequence. When there is more than one sequence in a species group, it joins two sequences of the same rank within the same species group to form an interolog. AlphaFold-Multimer uses this strategy and shows
  state-of-the-art accuracy in complex structure prediction~\citep{deepmind2021alphafold}. Practically, we run the implementation code of AlphaFold-Multimer following the default setting of the official repository (\url{https://github.com/deepmind/alphafold}). Notably, we only evaluate
  the unrelaxed model without the template information for time efficiency~\citep{jumper2021highly}.

  \paragraph{Genetic distances.}
  In bacteria, interacting genes sometimes are co-located in operons and co-transcribed to form protein complexes~\citep{salgado2000operons}. Consequently, we can detect interologs by the genetic distance of two genes. This strategy pairs sequences of the same species based on the
  distances of their positions in the contigs, which are retrieved from ENA. In our implementations, given a sequence from the first chain, we pair it with the sequence from the second chain that is closest to it in terms of genetic distance. If there is more than one
  closest sequence, we select the one that has the lowest e-value to the query sequence of the second chain; the e-value is calculated by the MSA search algorithm used to construct the chain MSA.

  \paragraph{Block diagonalization.}
  This strategy pads each chain sequence with gaps to the full length of the complex~\citep{evans2022protein}. Therefore, each sequence in the constructed joint MSA, except for the query sequence, will include non-gap tokens in exactly one chain and gap tokens in other chains. By
  sorting sequences in the joint MSA, we can make non-gap tokens appear only in the diagonal blocks, thus this strategy is termed block diagonalization. In our implementations, given a sequence from the first (second) chain, we append (prepend) non-gap tokens to it until
  the number of non-gap tokens equals the length of the second (first) chain.

  \subsection{Time/Memory requirement analysis}

  As ESMPair adopts column-wise attention score from ESM-MSA-1b as the metric to construct MSA of interologs, the major running time and memory requirement comes from ESM-MSA-1b. Practically, a single V100 GPU with 32GB memory can run a batch in the shape of 512 sequences
   with the max length of each sequence as 1024 within a few seconds. While other baseline methods like block diagonalization, Genome, or the default strategy of AF-M require no other machine learning models to supplement additional information. Thus, these methods are
  free of memory requirements.

  Generally, the major running time and memory requirement of end-to-end complex structure predictions lies in the process of MSA searching and the final AF-M prediction. Taking the target with PDB ID `4rca' as an example, it contains two subchains with each having about
  300 residues. Statistically, the running time of searching MSA of each subchain in the UniProt database with JackHMMER consumes about half an hour, which results in about 100K MSA sequences for each subchain. After that, ESMPair is applied to construct MSA of interologs
   within a few minutes, resulting in 20K MSA of interologs. As a follow-up, AF-M takes more than 20 minutes to use 5 models for predicting the structure of the target based on the obtained MSA of interologs, with each model predicting once without accessing the template
  and AMBER relaxation.

  In summary, the running time cost of sequence linking methods can be totally ignored in the end-to-end complex prediction pipeline with AF-M.

\section{Results}


\begin{table}[ht]
    \centering
    \caption{DockQ scores and success rate of PLM-enhanced pairing methods and baselines. We report the average of Top-5 Best
DockQ score, Top-1 Best DockQ score and Success Rate (DockQ$\ge 0.23$) (\%) on pConf70, DockQ49and pConf80 test sets. For one test
target, we predicted five different structures using the five AlphaFold-Multimer models.}
    \label{tab:esmpair-dockq}
    \small
    \begin{tabular}{lccc ccc ccc}
      \toprule
      \multirow{2}{*}{Method} & \multicolumn{3}{c}{pConf70} & \multicolumn{3}{c}{DockQ49} & \multicolumn{3}{c}{pConf80} \\
      \cmidrule(lr){2-4} \cmidrule(lr){5-7} \cmidrule(lr){8-10}
       & Top 5 & Top 1 & SR & Top 5 & Top 1 & SR & Top 5 & Top 1 & SR \\
      \midrule
      Block        & 0.199 & 0.179 & 30.4 & 0.212 & 0.194 & 49.0 & 0.351 & 0.319 & 51.2 \\
      Genome       & 0.215 & 0.182 & 33.7 & 0.219 & 0.195 & 49.0 & 0.377 & 0.346 & 54.7 \\
      AF-Multimer  & 0.234 & 0.203 & 42.4 & 0.247 & 0.219 & 58.0 & 0.408 & 0.369 & 62.5 \\
      \midrule
      ESMPair      & \textbf{0.259} & \textbf{0.214} & \textbf{42.4} & \textbf{0.265} & \textbf{0.235} & \textbf{58.7} & \textbf{0.423} & \textbf{0.378} & \textbf{63.1} \\
      \bottomrule
    \end{tabular}
  \end{table}
\begin{table}[ht]
    \centering
    \caption{Comparisons between ESMPair and AF-Multimer on targets from all range pConf scores. We report the average of DockQ
score, TMscore, ICS and IPS as the evaluation metrics (Larger values mean better performance).}
    \label{tab:esmpair-pconf-range}
    \small
    \resizebox{\textwidth}{!}{
    \begin{tabular}{lcccccccccccc}
    \toprule
    \multirow{2}{*}{Methods} & \multicolumn{3}{c}{DockQ} & \multicolumn{3}{c}{TMscore} & \multicolumn{3}{c}{ICS} & \multicolumn{3}{c}{IPS} \\
    \cmidrule(lr){2-4} \cmidrule(lr){5-7} \cmidrule(lr){8-10} \cmidrule(lr){11-13}
     & $< 0.7$ & $\ge 0.7$ & ALL & $< 0.7$ & $\ge 0.7$ & ALL & $< 0.7$ & $\ge 0.7$ & ALL & $< 0.7$ & $\ge 0.7$ & ALL \\
    \midrule
    AF-M & 0.225 & \textbf{0.757} & 0.687 & 0.788 & \textbf{0.901} & 0.886 & 0.234 & \textbf{0.771} & 0.700 & 0.445 & \textbf{0.766} & 0.724 \\
    ESMPair & \textbf{0.299} & 0.753 & \textbf{0.695} & \textbf{0.799} & 0.900 & 0.886 & \textbf{0.326} & 0.766 & \textbf{0.708} & \textbf{0.481} & 0.762 & \textbf{0.726} \\
    \bottomrule
    \end{tabular}
    }
\end{table}

\begin{table}[ht]
    \centering
    \caption{The Top-1 Best DockQ performance of two groups with different sequence length ( $\ge 100$ and $< 100$). The GAP value is the subtraction between the DockQ score of the two different length groups.}
    \label{tab:esmpair-seqlen}
    \small
    \begin{tabular}{lcccc}
    \toprule
    Methods & \# Targets & DockQ ($<100$) & DockQ ($\ge 100$) & GAP \\
    \midrule
    AF-M & 81 & 0.328 & 0.355 & -0.027 \\
    ESMPair & 152 & \textbf{0.359} & \textbf{0.366} & \textbf{-0.007} \\
    \bottomrule
    \end{tabular}
\end{table}

\begin{table}[ht]
    \centering
    \caption{The Top-1 Best DockQ performance with or without full AF-M features.}
    \label{tab:esmpair-fullfeatures}
    \small
    \begin{tabular}{lcccc}
    \toprule
    Methods & DockQ & TMscore & ICS & IPS \\
    \midrule
    \multicolumn{5}{l}{\textbf{Without full AF-M features}} \\
    AF-M & 0.20 & 0.69 & 0.26 & 0.41 \\
    ESMPair & 0.21 & 0.69 & 0.28 & 0.41 \\
    \midrule
    \multicolumn{5}{l}{\textbf{With full AF-M features}} \\
    AF-M & 0.29 & 0.72 & 0.35 & 0.44 \\
    ESMPair & \textbf{0.32} & 0.71 & \textbf{0.39} & 0.45 \\
    \bottomrule
    \end{tabular}
\end{table}

  In this section, we first briefly outline the framework of ESMPair for PCP. Then, we discuss how our proposed method has a better complex prediction accuracy than previous MSA pairing methods. We find that the ensemble strategy showcases better performance than the
  default single strategy. We further quantitatively analyze several key factors and hyperparameters that may impact the performance of our method, and also explore the capability of different measurements to distinguish the acceptable predictions from the unacceptable
  ones. Finally, we compare the performance between ESMPair and AlphaFold-Multimer on CASP15 heteromers.

  \subsection{ESMPair overview}

  The overall framework of ESMPair is illustrated in~\Cref{fig:esmpair_overview} with the details in Methods. In complex structure prediction, predictors such as AlphaFold-Multimer use inter-chain co-evolutionary signals by pairing sequences between MSAs of constituent single
  chains of the query complex. Formally, given a query heterodimer, we obtain individual MSAs of its two constituent chains, denoted as $M_1 \in \mathcal{A}^{N_1 \times C_1}$ and $M_2 \in \mathcal{A}^{N_2 \times C_2}$, where $\mathcal{A}$ is the alphabet used by PLM,
  $N_1$ and $N_2$ are the number of sequences in MSAs $M_1$ and $M_2$, and $C_1$ and $C_2$ are the sequence lengths. The MSA pairing pipeline aims at designing a matching or an injection $\pi : [N_1] \to [N_2]$ between MSAs from each chain to build the MSA of interologs,
  dubbed as $M_\pi \in \mathcal{A}^{N \times (C_1+C_2)}$, where $N$ is the number of sequences in the joint MSA. In practice, the MSA of interologs $M_\pi$ is a collection of the concatenated sequences $\{\text{concat}(M_1[i], M_2[\pi(i)]) : i \in P\}$, where $P$ is the
  indices of the sequences from $M_1$ that can be paired with any sequences from $M_2$ according to the matching pattern $\pi$. Then the MSA of interologs is taken by predictors as input to predict the structure of the query heterodimer.

  \subsection{ESMPair outperforms other MSA pairing methods on heterodimer predictions}

  \paragraph{Overall evaluation.}
  For each test target we predict five 3D structures using AlphaFold-Multimer's five models and then report the average of Top-$k$ ($k=1, 5$) Best DockQ score of the predicted structures and the corresponding success rate (SR) in Table~\ref{tab:esmpair-dockq}. Our method outperforms
  the other methods. To be specific, our method outperforms AF-Multimer's default MSA pairing strategy on all three test sets (0.259 versus 0.234 on pConf70, 0.423 versus 0.406 on pConf80 and 0.265 versus 0.242 on DockQ49, in terms of Top-5 DockQ score). Our experimental
  results confirm that our proposed column-wise-attention-based MSA pairing method, ESMPair, is better than the sequence similarity-based method used in AF-Multimer.

  Among all the MSA pairing methods, block diagonalization performs the worst ($-30\%$ compared with ESMPair in terms of the average of Top-5 best DockQ). The result indicates that the inter-chain co-evolutionary information helps with complex structure prediction. Among
  MSA pairing baselines, AF-Multimer surpasses genetic co-localization by a large margin ($+12.8\%$ Top-5 DockQ). All the proposed PLM-enhanced pairing methods substantially outperform the block diagonalization and the genetic-based methods. Even though AF-Multimer may
  have overly optimistic performance using the default pairing method since the training MSAs are built using it, ESMPair further exceeds it by a large margin ($+4.2\sim10.7\%$ Top-5 DockQ score over
  three test sets).

\begin{figure}[ht!]
    \centering
    \includegraphics[width=\textwidth]{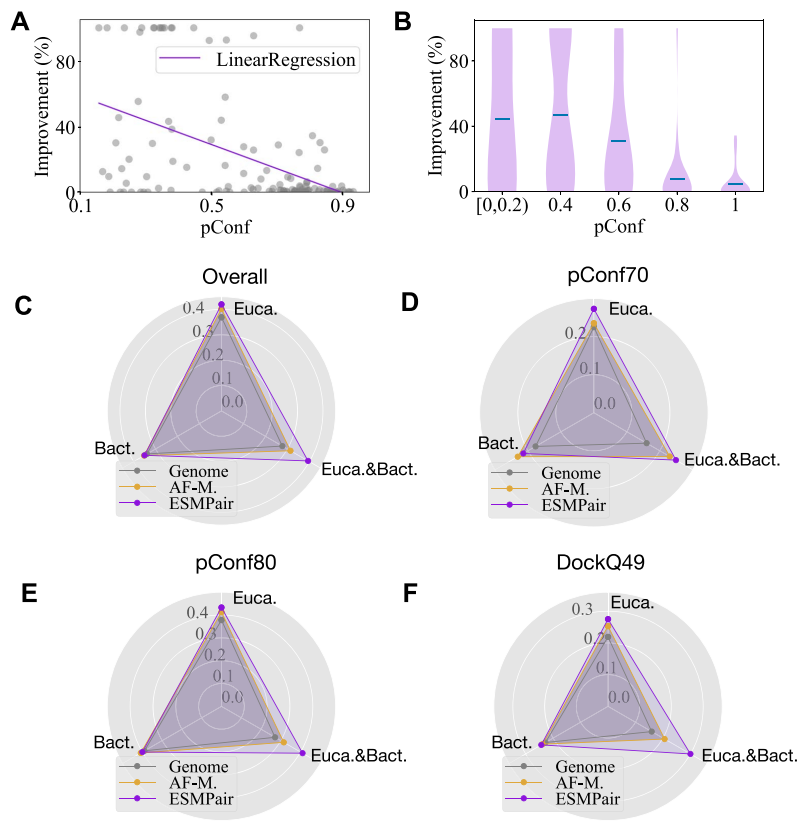}
    \caption{Prediction performance across pConf score regions and taxonomic domains. (a--b) Negative correlation between the relative improvement of ESMPair over AF-Multimer and pConf score. (c--f) DockQ score comparison among ESMPair,
  AF-Multimer, and Genome on Eucaryote, Bacteria, and Eucaryote\&Bacteria domains. Eucaryote\&Bacteria denotes heterodimers whose two chains belong to different domains. Heterodimers in our dataset originate from Eucaryotes, Bacteria, Viruses,
  and Archaea; we group Bacteria, Viruses, and Archaea into the Bacteria domain. Across all test sets, ESMPair significantly outperforms both baselines on Eucaryote targets.}
    \label{fig:esmpair_pconf}
  \end{figure}

  \paragraph{ESMPair performs better on low pConf targets.}
  As shown in Table~\ref{tab:esmpair-dockq}, the performance gap between ESMPair and AF-Multimer becomes narrower on pConf80 than on pConf70, with improvement ratio from 3.7\% to 10.7\%. For an in-depth analysis, we quantitatively analyze the correlations between the predicted
  confidence score (pConf) estimated by AF-Multimer and the performance gap of the average of Top-5 Best DockQ score between ESMPair and AF-Multimer on DockQ49, as illustrated in~\Cref{fig:esmpair_pconf}(a--b).

  The relative improvement is negatively correlated (Pearson Correlation Coefficient is $-0.49$) with the predicted confidence score. When pConf is less than 0.2, the relative improvements even achieve 100\%, while when pConf is more than 0.8, ESMPair performs nearly on
  par with AF-Multimer. This is because AF-Multimer can do well on relatively easier targets; it is very challenging to further improve it.

  To further quantify the performance comparisons between ESMPair and AF-Multimer on natural targets with all range of pConf scores, we follow the same data processing pipeline to generate the dataset without applying any filtering based on pConf scores. Subsequently, we
  randomly select 300 targets, and use ESMPair and the default pairing strategy of AF-Multimer to predict their structures. Of these 300 targets, 256 have pConf scores greater than or equal to 0.7 while 44 have pConf scores lower than 0.7. For convenience, we use model 1
  to generate predictions for each target, and make only one prediction per target. We report the average of DockQ score, TMscore, Interface Contact Score (ICS) and Interface Path Score (IPS) as the evaluation metrics, shown in Table~\ref{tab:esmpair-pconf-range}. The results suggest
  ESMPair outperforms AF-M on targets with low pConf (pConf $< 0.7$) scores, whereas it performs comparably with AF-M on those with high pConf (pConf $\geq 0.7$) scores.

  \paragraph{ESMPair has a higher prediction accuracy on eukaryote targets.}
  We further compare the DockQ distribution of ESMPair, AF-Multimer and Genome on three kingdoms, i.e.\ Eukaryote, Bacteria and Eukaryote \& Bacteria, which is a special domain where the two constituent chains in the heterodimer belong to the two domains, respectively. To
   be specific, all the heterodimers from pConf70, DockQ49 and pConf80 are divided via the domains of Eukaryotes, Bacteria, Viruses, Archaea, Eukaryotes;Bacteria, respectively. Note that we group the data from Bacteria, Viruses and Archaea as the Bacteria domain. \Cref{fig:esmpair_pconf}(c--f) demonstrates that ESMPair performs better than the other two MSA pairing methods on the Eukaryotes data by a large margin (0.420 for ESMPair, 0.402 for AF-Multimer and 0.369
  for Genome on the overall data). As it is notoriously difficult to identify homologous protein sequences for the Eukaryotes data, ESMPair has a desirable property to build effective interologs on the Eukaryotes. While in the Bacteria data, three strategies have similar
  performance (around 0.35 on the whole data). Most strikingly, we find ESMPair has an extraordinary performance on the Euka.\,\&\,Bact.\ data over the other two methods (0.394 for ESMPair, 0.314 for AF-Multimer and 0.277 for Genome on the overall data). We further check
  the performance gap for each target from the Euka.\,\&\,Bact.\ data. ESMPair performs significantly better on three out of six targets: 0.443 (ESMPair) versus 0.013 (AF-Multimer) on 5D6J, 0.289 versus 0.201 on 6B03, and 0.864 versus 0.854 on 7AYE. Besides, ESMPair
  performs on par with AF-Multimer on the other three targets. These results shed light on the robustness of PLMs.

  \paragraph{ESMPair outperforms AF-Multimer on most of the newly released targets.}
  \begin{figure}[ht!]
    \centering
    \includegraphics[width=\textwidth]{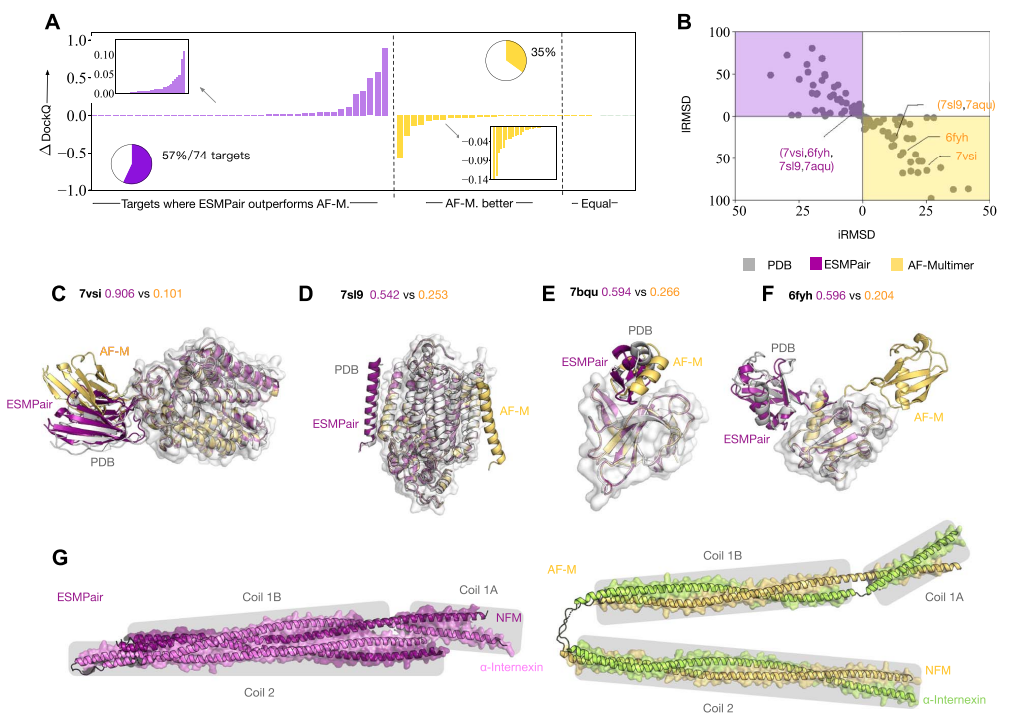}
    \caption{Comparison of ESMPair and AF-Multimer on newly released targets (a--f) and an unresolved case (g). (a--f) Evaluations on 74 targets released after 30 April 2018. (a) Bar chart showing the relative performance gap between ESMPair and
   AF-Multimer across three categories: ESMPair outperforms AF-Multimer, AF-Multimer outperforms ESMPair, and equal performance. (b) Interface and ligand RMSD distributions of structures predicted by ESMPair (purple) and AF-Multimer (yellow).
  (c--f) Four representative cases: AF-Multimer predicts incorrect ligand orientations for 7VSI and 7AQU, and incorrect binding sites for 7SL9 and 6FYH. (g) The intermediate filament NFM--INA heterodimer predicted by ESMPair forms a four-helix
  bundle. Gray boxes indicate the interacting motifs of coil 1A, coil 1B, and coil 2 of the two proteins.}
    \label{fig:esmpair-newcases}
  \end{figure}
  
  We further select 74 targets that AF-Multimer does not train on~\citep{deepmind2021alphafold}, i.e.\ the targets whose release date is later than 30 April 2018, from the test dataset. Then we compare the performance of predicted structures on these targets between ESMPair and
  AF-Multimer in~\Cref{fig:esmpair-newcases}. ESMPair outperforms AF-Multimer on most of the targets (57\%) with a relatively larger performance gap, while AF-Multimer outperforms ESMPair on fewer targets (35\%) with a relatively lower gap. We further
  plot the distributions between interface RMSD and ligand RMSD of predicted structures via ESMPair and AF-Multimer in~\Cref{fig:esmpair-newcases}(b). The holistic distributions predicted by ESMPair are closer to the origin of coordinates than those predicted by AF-Multimer, which
   strongly proves that ESMPair is superior to AF-Multimer on the predictions of newly released targets.

  Furthermore, we show why ESMPair performs better than AF-Multimer by analyzing four PDB targets, 7VSI, 7AQU, 6FYH and 7SL9, in~\Cref{fig:esmpair-newcases}(c--f). Among these, 7VSI and 6FYH have a larger predicted iRMSD and lRMSD variance by AF-Multimer, because AF-Multimer
  predicts the wrong binding sites. While AF-Multimer predicts the right binding sites on 7SL9 and 7AQU that have a smaller predicted iRMSD and lRMSD variance, it unfortunately predicts the wrong ligand orientations. In contrast, our proposed ESMPair correctly predicts
  the binding sites on the receptor and also places the ligand in the approximately correct relative orientation.

  To better illustrate the usage of ESMPair in predicting the protein complexes without known resolved 3D structures, we inspected the intermediate filament heterodimer formed between the neurofilament medium polypeptide (NFM, UniProt ID P08553) and $\alpha$-internexin
  (UniProt ID P46660), which is known to form an anti-parallel four-helix bundle~\cite{kreplak2006neurofilament,herrmann2004intermediate}. As shown in~\Cref{fig:esmpair-newcases}(g), both ESMPair and AF-Multimer correctly predict the three binding interfaces from NFM and
  $\alpha$-internexin. However, ESMPair predicted the two coiled coils to pack as a four-helix bundle, which is consistent with the experimental evidence, while AF-Multimer predicted the two coiled coils to be separated. This case demonstrates the potential to apply
  ESMPair to model unresolved protein complexes.

  \paragraph{ESMPair is more robust than AF-Multimer on different sequence lengths.}
  We split the targets from pConf70, pConf80 and DockQ49 datasets via the sequence length into two groups: one is the targets with $\geq 100$ residues and the other one owns the targets with $< 100$ residues. Note that we use the shorter protein between the two chains as
  the length of targets. We provide the average Top-1 Best DockQ comparison between the targets as shown in Table~\ref{tab:esmpair-seqlen}. The results demonstrate that ESMPair performs consistently better than AF-M in different lengths. Moreover, ESMPair is robust for complexes with
  variable lengths.

  \paragraph{ESMPair outperforms AF-Multimer with or without the full features.}
  We use the pConf70 set for comparing the performance between ESMPair and AF-Multimer on the full feature settings, i.e.\ adding the template information and AMBER force-field. For each target, we run each of the five models once, as shown in Table~\ref{tab:esmpair-fullfeatures}. We can
  conclude that (1) the full feature setting indeed significantly improves the performance of ESMPair; (2) ESMPair rivals AF-Multimer in all settings.


  \subsection{Ensemble improves the prediction accuracy}
\begin{figure}[ht!]
    \centering
    \includegraphics[width=\textwidth]{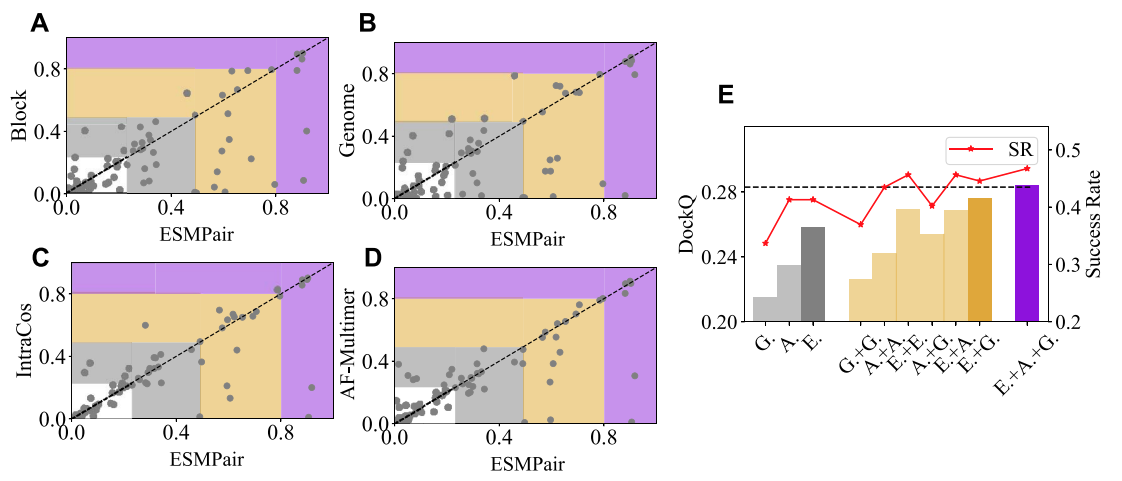}
    \caption{Comparison of ESMPair with four alternative MSA pairing approaches (a--d) and ensemble strategies (e) on pConf70 targets. (a--d) Each point shows the DockQ score of a target for ESMPair (x-axis) versus the compared method (y-axis).
  Points below the diagonal indicate ESMPair outperforms the alternative. Highlighted regions denote incorrect (white), acceptable (gray), medium (yellow), and high-quality (purple) predictions by DockQ score. (e) Gray bars show single-strategy
  performance, where G.\ = Genome, A.\ = AF-Multimer, and E.\ = ESMPair. ESMPair achieves the best single-strategy result (0.259 DockQ, 42.4\% success rate). Yellow bars show pairwise ensembles, with ESMPair + Genome performing best (0.277
  DockQ, 44.6\% success rate). The purple bar shows the three-strategy ensemble achieving the highest overall performance (0.285 DockQ, 46.8\% success rate).}
    \label{fig:esmpair-ensemble}
  \end{figure}
  From ~\Cref{fig:esmpair-ensemble}(a--d), we found that different MSA pairing methods have their own advantages; even block diagonalization performs slightly better than ESMPair on about 30\% of targets, which implies that they can complement each other. To verify that, we combine
   10 models predicted by any two of the MSA pairing methods, then we report the average of Top-5 Best DockQ score, as shown in~\Cref{fig:esmpair-ensemble}(e). The ensemble strategies, i.e.\ the
  yellow and purple bars, significantly outperform the corresponding single strategy, i.e.\ the gray bars. ESMPair in addition to any one of the single strategies always
   has a better performance than the one without ESMPair; for example, the SR of ESMPair + Genome is 44.6\% versus 40.4\% of AF-Multimer + Genome. Finally, the ensemble of all three strategies, i.e.\ the purple bar, reaches the best performance with 0.285 DockQ score and
  46.8\% Success Rate, which motivates us that instead of merely using a single strategy to build interologs, the ensemble MSA pairing strategy may be the silver bullet to identify more effective interologs.

  \subsection{Factors influencing prediction accuracy}
  \begin{figure}[ht!]
    \centering
    \includegraphics[width=\textwidth]{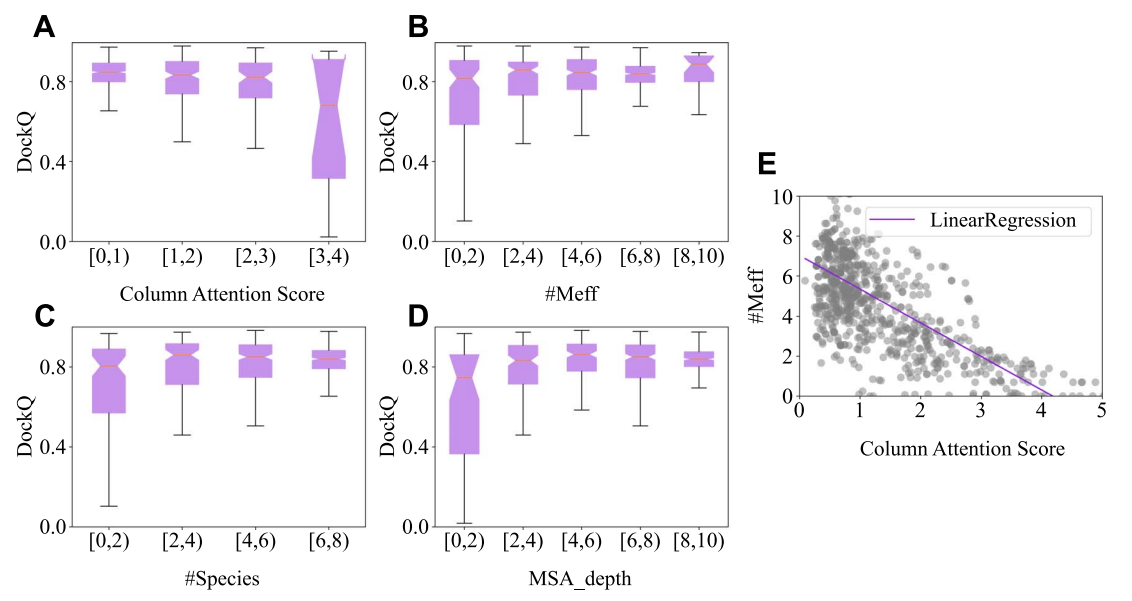}
    \caption{Factors affecting structure prediction performance. Correlation between average Top-5 Best DockQ score and (a) column attention score (log-scale) predicted by ESM-MSA-1b, (b) number of effective sequences (Meff), (c) number of
  species, and (d) depth of paired MSA (log-scale). (e) Distribution of column attention score versus the number of effective interologs. The red curve shows the fitted linear regression model (Pearson $r \approx -0.70$), indicating that higher
  column attention scores correspond to fewer effective interologs.}
    \label{fig:esmpair-factors}
  \end{figure}
  We investigate the connections between the performance of ESMPair and some key factors of the formed MSA of interologs, such as the column-wise attention score (i.e.\ ColAttn\_score), the number of effective sequences within MSA measured by Meff (i.e.\ \#Meff), the
  number of species (i.e.\ \#Species) and the depth of MSA (i.e.\ Msa:Depth). To be specific, we predict 1689 heterodimers sampled from PDB without filtering and divide them into different regions according to the value of each factor. Notably, for ColAttn\_score, we
  average the score of each single chain in interolog, then re-scale it in the logarithm form, and then average ColAttn\_score of all interologs from the paired MSA as the final score of the target. For \#Meff, \#Species and Msa:Depth, we directly calculate the
  corresponding statistics based on the interologs.

  The correlations between DockQ score and each of the above factors are illustrated in~\Cref{fig:esmpair-factors}. \#Meff, \#Species and Msa:Depth have a similar trend where the predicted structure accuracy improves with the increasing of these factors. It implies that MSA with
  more diversity represents more co-evolutionary information that benefits structure predictions of AF-Multimer, which also aligns with previous insights~\cite{rao2021msa}. Moreover, increasing ColAttn\_score results in decreasing structure prediction accuracy. Considering the
   self-attention mechanism in the PLM, given a sequence as the query, the self-attention mechanism aims at identifying the sequence with high homology affinity, i.e.\ the sequence with a high similarity score~\cite{rao2021msa}. Therefore, a large ColAttn\_score indicates the
  MSA with a low \#Meff, which potentially results in inaccurate structure prediction. To justify our speculation, we explicitly characterize the dependency between ColAttn\_score and \#Meff, as shown in~\Cref{fig:esmpair-factors}(e). ColAttn\_score has shown a negative
  correlation to the \#Meff, with the Pearson correlation coefficient of $-0.70$, which elucidates that a higher ColAttn\_score reflects MSA with lower sequence diversity.

\section{Conclusion}

This paper explores a series of simple yet effective MSA pairing
algorithms based on pre-trained PLMs for constructing effective interologs. To the best of our knowledge, this is the first
time that PLMs are used to construct joint MSAs. Experimental results have confirmed that the proposed ESMPair significantly outperforms the state-of-the-art phylogeny-based protocol
adopted by AlphaFold-Multimer. What is more, ESMPair performs
significantly better on targets from eukaryotes, which are hard
to be predicted accurately by AF-Multimer. We further confirm
that, instead of using the conventional single strategy to build
interologs, the ensemble MSA pairing strategy can largely improve
the structure prediction accuracy. Generally, ESMPair has a profound impact on biological applications depending on the high quality MSA. In the future, we will continue to explore more
potential ways to leverage the advantages of PLM in building and
choosing MSA. We also look forward to applying our proposed
methods to improve current MSA-based applications.

\bibliographystyle{siam}
\bibliography{references}
\chapter{Redesign Selective Protein Binders Using Contrastive Decoding}

\section{Introduction}

Designing protein binders that target specific proteins with high affinity and specificity is a fundamental challenge in protein engineering with broad applications in therapeutics and basic research~\cite{kuhlman2019advances,kortemme2024novo,chu2024sparks}. While recent advances in deep learning have improved design capabilities, generating binders that achieve both tight binding and target specificity without extensive experimental optimization remains difficult~\cite{zambaldi2024denovo, pacesa2025one}.

Current structure-based binder design, also known as one-sided interface design~\cite{kuhlman2019advances}, typically follows a three-stage workflow~\cite{zambaldi2024denovo, pacesa2025one, watson2023novo}: (1) backbone or all-atom structure generation conditioned on the target protein, (2) sequence design to optimize binding properties and generate diverse candidates, and (3) filtering and ranking of designs using machine learning and physics-based scoring functions. Recent years have seen substantial progress in each component. However, existing fixed-backbone design methods do not adequately address the specific requirements of binder design, including affinity optimization, reduction of off-target binding, and accurate representation of interface side-chain conformations that determine binding specificity.

Here we focus on the fixed-backbone design component in the context of binder design. Fixed-backbone binder design aims to generate sequences for a binder that can fold independently and bind a target protein to form a predetermined complex structure~\cite{kuhlman2019advances}. ProteinMPNN has emerged as a widely adopted model for this task due to its efficiency and effectiveness at designing sequences from backbone structures, particularly for monomeric proteins with idealized scaffolds. However, ProteinMPNN and similar fixed-backbone design models have three key limitations for binder design. First, these models operate solely on backbone atom positions and cannot capture side-chain conformations that are critical for defining binding interfaces and specificity. Second, successful binder design requires simultaneously optimizing two objectives: the binder must fold stably in its unbound state and form a stable complex with the target. Current autoregressive sequence design models generate sequences based on per-residue conditional probabilities and lack explicit mechanisms to jointly optimize the two objectives. Third, current fixed-backbone design models lack capabilities to improve binding specificity.

To address these limitations, we introduce RedNet \footnote{Code and data are available at \url{https://github.com/zw2x/rednet_public}.}, a framework for binder sequence design with two main innovations. First, we develop a multiscale graph neural network architecture that incorporates both backbone geometry and side-chain
  information from the target, enabling more accurate modeling of binding interfaces. Second, we introduce a contrastive decoding algorithm that leverages the trained model to improve binding while simultaneously reducing off-target
  interactions. To our best knowledge, this is the first principled algorithm that enables deep learning-based fixed-backbone binder design to explicitly improve both affinity and specificity.

  We evaluate RedNet on multiple benchmarks relevant to fixed-backbone binder design. On native sequence recovery, RedNet achieves 43\% on heterodimers, a 30\% relative improvement over ESM-IF (33\%) and 16\% over ProteinMPNN (37\%). On heterodimer self-consistency evaluated by AlphaFold3, RedNet
  with contrastive decoding (RedNet-CD) matches native sequence success rates (68\%) on high-quality targets, outperforming ProteinMPNN (59\%) and ESM-IF (61\%). Rosetta energetic analysis confirms that RedNet-CD and RedNet-Ens (an ensemble of
  RedNet and RedNet-CD) produce designs with native-like or superior binding energetics, hydrogen bonding, and surface hydrophobicity. Additionally, we curate a new benchmark from the PDB to assess binding selectivity against structurally
  similar off-targets. On this benchmark, RedNet-CD achieves 64.81\% energetic selectivity at the base threshold, nearly doubling baseline RedNet (33.33\%) and outperforming all other methods, demonstrating that contrastive decoding specifically
   enhances the ability to discriminate between on-target and off-target interactions.

\subsection{Related Works}

\paragraph{Fixed backbone design.}
  Physics-based fixed backbone design relies on two components: an energy function to model sequence-structure compatibility and a search algorithm to explore sequence space. Energy functions modeling van der Waals interactions, hydrogen bonding, electrostatics, and
  solvation are parameterized to reproduce features of natural proteins~\cite{alford2017rosetta}. They are optimized to improve accuracy using small molecule and macromolecular data~\cite{park2016simultaneous}. Search algorithms fall into two categories: stochastic approaches such as Rosetta use
  simulated annealing to identify low-energy sequences~\cite{alford2017rosetta}, while deterministic approaches such as OSPREY employ dead-end elimination to provably identify the minimum energy conformations~\cite{hallen2018osprey}. Although both approaches have demonstrated experimental
  successes, they become computationally demanding for large sequence spaces and typically require extensive experimental screening to identify functional designs.

  \paragraph{Deep learning for fixed backbone design.}
  Deep learning methods have substantially improved both computational efficiency and experimental success rates for fixed backbone design. These approaches can be categorized as autoregressive (AR) or non-autoregressive (NAR). AR models, including Structured
  Transformer~\cite{ingraham2019generative} and ProteinMPNN~\cite{dauparas2022robust}, generate sequences iteratively using graph-based encoders that capture local structural context. NAR models such as PiFold~\cite{gao2023pifold} generate entire sequences in a single forward pass, achieving significant
  speedups while maintaining competitive accuracy. ProteinMPNN has demonstrated substantially higher experimental success rates than Rosetta across diverse design challenges, in particular monomeric proteins~\cite{dauparas2022robust}. These methods are now widely adopted in de novo
  protein design workflows.

  \paragraph{One-sided interface design.}
  One-sided interface design, where a protein binder is designed against a fixed target, has broad applications from modulating signaling receptors to neutralizing pathogens. Physics-based approaches have achieved limited success, with functional designs typically
  restricted to targets with favorable features such as hydrophobic patches or concave surfaces, and to small, easily stabilized scaffolds~\cite{chevalier2017design}.

  Deep learning has enabled substantial progress in this area. Two-stage approaches such as RFdiffusion~\cite{watson2023novo} first generate binder backbones via structure diffusion, then design sequences using ProteinMPNN. End-to-end methods such as BindCraft~\cite{pacesa2025one}
  directly optimize sequences through backpropagation of AlphaFold2 confidence metrics. Both approaches have produced experimentally validated binders. However, success rates vary considerably across targets (0\% to $>$90\%) \cite{zambaldi2024denovo}, and many designs require further optimization
  to achieve high binding affinities.

  \paragraph{Multistate design for binding specificity.}
  Many applications require designing proteins with defined binding specificities (selectively binding one target while avoiding others). Physics-based multistate design algorithms, such as Rosetta MSD~\cite{humphris2005multistate}, address this by simultaneously optimizing sequences
  across multiple structural states to maximize energy gaps between desired and undesired interactions. Recent work has incorporated deep learning fixed backbone design into multistate workflows~\cite{hong2024integrative} or extended to modeling structural ensembles, demonstrating success in applications such as conformational
  switches~\cite{guo2025deep}. However, whether these approaches generalize to tuning binding specificities remains unclear. Alternative strategies employ experimental screening heuristics, such as differential yeast display, to identify selective binders~\cite{zhou2024general}, but such
   approaches do not leverage computational design capabilities to systematically optimize both affinity and specificity.

\section{Data and Methods}
\subsection{Protein Graph Representation}

We utilize multi-scale graph representations to model the backbone and side-chain geometry of protein complex structures.

\paragraph{Backbone representation for complex structures.} We represent the protein complex as a residue-level graph $G = (V, E)$, where each node $v_i \in V$ corresponds to a residue and edges $e_{ij} \in E$ connect spatially proximal residues within a distance cutoff. Local frames are derived from backbone atom coordinates (N, C$\alpha$, C, O) and edges are constructed using $k$-nearest neighbors based on C$\alpha$ distances as in GLINTER~\cite{xie2022deep}.

\paragraph{Sidechain representation for target chains.} For target chains where sequence information is available, we construct an all-atom graph $G_{\text{atom}} = (V_{\text{atom}}, E_{\text{atom}})$ to capture detailed sidechain interactions. Each node $v_a \in V_{\text{atom}}$ represents a heavy atom, with features encoding local chemical environment; shown in later section \Cref{sec:rednet-feat}. Edges connect atoms within a distance cutoff and encode pairwise distances and other features.

\subsection{Network Architectures}

 \subsubsection{Overview}
\begin{figure}[ht!]
    \centering
    \includegraphics[width=0.8\textwidth]{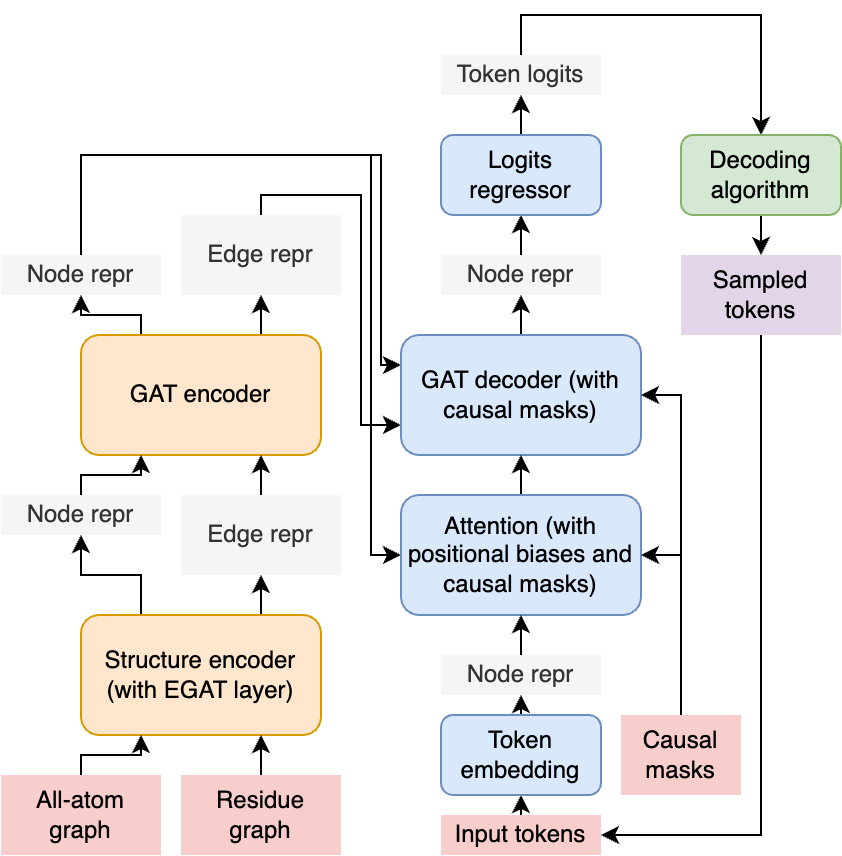}
    \caption{Overview of the RedNet architecture. Graph neural networks encode protein structure into node and edge representations, which are then decoded by a causal transformer to autoregressively predict amino acid sequences.}
    \label{fig:rednet_overview}
\end{figure} 
  The diagram of the overall architecture is shown in \Cref{fig:rednet_overview}. We employ graph neural networks to capture the protein graphs and a causal transformer to decode amino acids at each timestep. We propose several improvements over existing encoder-decoder graph neural network architectures for protein design.

  \paragraph{Graph attention network handles heterogeneous neighborhoods.} 
  \renewcommand{\KwIn}[2]{\textbf{def} #1(#2):}

  \SetKwProg{Fn}{def}{:}{}
  \SetKwComment{Comment}{\# }{}
  \SetKw{KwReturn}{return}

  \SetKwFunction{GATLayer}{GATLayer}
  \SetKwFunction{Linear}{Linear}
  \SetKwFunction{MLP}{MLP}
  \SetKwFunction{Softmax}{Softmax}
  \SetKwFunction{LeakyReLU}{LeakyReLU}
  \SetKwFunction{Sigmoid}{Sigmoid}
  \SetKwFunction{Gather}{Gather}

  \newcommand{\NodeRepr}{$s$}
  \newcommand{\EdgeRepr}{$p$}
  \providecommand{\EdgeIdx}{$\mathcal{E}$}
  \newcommand{\EdgeMask}{$M$}
  \newcommand{\NodeOut}{$s^{\text{out}}$}

  \begin{algorithm}[t!]
    \setlength{\algomargin}{0em}
    \DontPrintSemicolon
    \SetAlgoSkip{medskip}

    \caption{GAT  Layer}\label{algo:rednet_gat}

    \Comment{\NodeRepr: node features; \EdgeRepr: edge features; \EdgeIdx: neighbor indices; \EdgeMask: edge mask}

    \Fn{\GATLayer{\NodeRepr, \EdgeRepr, \EdgeIdx, \EdgeMask}}{

      \Comment{Build edge messages from nodes and edges}
      $m_{ij} \gets \Linear(p_{ij}) + \Gather(W^{\text{src}} s, \mathcal{E}) + W^{\text{tgt}} s_i$\;
      $m_{ij} \gets \MLP(m_{ij})$\;

      \BlankLine
      \Comment{Global pooling with gating}
      $o_i^{\text{global}} \gets \frac{1}{|\delta(i)|} \sum_{j \in \delta(i)} m_{ij}$\;
      $\Delta s \gets \Linear(\Sigmoid(W^g s_i) \odot o_i^{\text{global}})$\;

      \BlankLine
      \Comment{Graph attention with gating}
      $\alpha_{ij} \gets W^A \cdot \LeakyReLU(\Linear(m_{ij}))$\;
      $\alpha_{ij} \gets \Softmax_{j \in \delta(i)}(\alpha_{ij})$\;
      $v_{ij} \gets \Linear(m_{ij})$\;
      $o_i^{\text{gat}} \gets \sum_{j \in \delta(i)} \alpha_{ij} v_{ij}$\;
      $\Delta s \gets \Delta s + \Linear(\Sigmoid(W^{g'} s_i) \odot o_i^{\text{gat}})$\;

      \BlankLine
      \Comment{Residual updates}
      $s \gets s + \text{Dropout}(\Delta s)$\;
      $s \gets s + \text{Dropout}(\MLP(s))$\;

      \KwReturn $s$, $p$\;
    }
  \end{algorithm}
  
  The pseudocode for our variant of GAT \cite{velickovic2018graph,brody2022gatv2} is shown in \Cref{algo:rednet_gat}. Edge messages are first constructed by combining projected edge features with source node features (via gather) and target node features. These messages are refined through an MLP. For
  autoregressive decoding, causal masks are precomputed based on the sampled decoding order and edge indices; the edge mask $M$ ensures that each residue can only attend to neighbors that have already been seen. The layer then applies two complementary aggregation strategies: (1) a global pooling branch that computes the mean of neighboring messages, modulated by a learned sigmoid gate, and (2) a graph attention branch where attention weights are computed via LeakyReLU activation, allowing the model to selectively attend to informative neighbors while down-weighting noisy or less relevant nodes. Both branches use gating mechanisms initialized near zero. The final node representations are updated through residual connections with dropout regularization.

  \paragraph{Causal graph transformer with pairwise positional biases.}
    \renewcommand{\KwIn}[2]{\textbf{def} #1(#2):}

  \SetKwProg{Fn}{def}{:}{}
  \SetKwComment{Comment}{\# }{}
  \SetKw{KwReturn}{return}

  \SetKwFunction{GraphAttention}{GraphAttention}
  \SetKwFunction{GraphTransformer}{GraphTransformer}
  \SetKwFunction{Linear}{Linear}
  \SetKwFunction{MLP}{MLP}
  \SetKwFunction{Softmax}{Softmax}
  \SetKwFunction{Sigmoid}{Sigmoid}
  \SetKwFunction{LayerNorm}{LayerNorm}
  \SetKwFunction{Dropout}{Dropout}

  \newcommand{\Nodes}{$s$}
  \newcommand{\Edges}{$p$}
  \newcommand{\AttnMask}{$M$}

  \begin{algorithm}[t!]
    \setlength{\algomargin}{0em}
    \DontPrintSemicolon
    \SetAlgoSkip{medskip}

    \caption{Attention with Pairwise Biases}\label{algo:rednet_attn}

    \Comment{\Nodes: node features; \Edges: pairwise features; \AttnMask: attention mask}

    \Fn{\GraphAttention{\Nodes, \Edges, \AttnMask}}{

      \Comment{Project to queries, keys, values}
      $Q \gets \Linear(s)$\;
      $K, V \gets \text{split}(\Linear(s))$\;
      $B \gets \Linear(p)$ \Comment*[r]{pairwise bias}

      \BlankLine
      \Comment{Compute attention with pairwise bias}
      $A_{ij} \gets \frac{Q_i \cdot K_j}{\sqrt{d}} + B_{ij}$\;
      $A_{ij} \gets A_{ij}.\text{masked\_fill}(\neg M_{ij}, -\infty)$\;
      $\alpha_{ij} \gets \Softmax_j(A_{ij})$\;

      \BlankLine
      \Comment{Aggregate and project with gating}
      $o_i \gets \sum_j \alpha_{ij} V_j$\;
      $o_i \gets \Sigmoid(\Linear(s_i)) \odot o_i$\;

      \KwReturn $\Linear(o)$\;
    }

  \end{algorithm}

    MPNNs suffer from limited expressivity, over-smoothing, and over-squashing \cite{xu2018how,li2018oversmoothing,alon2021on}. Adding proper normalization \cite{zhou2020dgn,cai2021graphnorm}, positional biases and global attentions \cite{dwivedi2020generalization,ying2021graphormer,rampasek2022recipe} may address some of these issues; however, long-range interactions remain difficult to capture. \textit{De novo} binder design
  methods~\cite{pacesa2025one} tend to exploit shape complementarity and short-range hydrophobicity for generating successful designs, yet long-range features of natural proteins—such as allosteric pathways—which are
  essential for certain biological functions \cite{wodak2019allostery}, may not be adequately captured. We hypothesize that adding global-attention may improve sequence modeling of long range interactions. The pseudocode is shown in \Cref{algo:rednet_attn}.

\subsubsection{Multiscale-Graph Featurizers}\label{sec:rednet-feat}
\begin{table}[ht!]
      \caption{Summary of input features. Core atoms: N, C$\alpha$, C, O, pseudo-C$\beta$ ($C{=}5$).
      The residue graph is a $K$-NN graph ($K{=}48$) over C$\alpha$ distances.
      The atom graph connects each C$\alpha$ to nearby atoms via a radius graph ($r{=}15$\,\AA, max $k{=}96$).
      RBF: $\phi(d) = \exp(-(d - \mu_i)^2)$, $\mu_i$ linearly spaced in $[2, 22]$, $D{=}16$ bins.
      Atom type vocabulary $A{=}37$; residue type vocabulary $R{=}33$.
      $N$: residues; $M$: atoms; $E$: atom graph edges.}
      \label{tab:input_features}
      \centering
      \small
      \resizebox{\textwidth}{!}{%
      \begin{tabular}{l l l}
      \toprule
      Feature & Shape & Description \\
      \midrule
      \multicolumn{3}{l}{\textit{Residue graph: edge features}} \\
      Core atom pairwise RBF & $(N, K, C^2 D)$ & RBF over pairwise distances between core atoms \\
      Core atom inverse distance & $(N, K, C^2)$ & $(1+d_{ij})^{-1}$ for each core atom pair \\
      Relative residue index & $(N, K)$ & Sequence offset encoding \\
      Same-chain indicator & $(N, K)$ & 1 if same chain \\
      Frame-relative positions & $(N, K, 3C)$ & Core atom coordinates in local N--C$\alpha$--C frames \\
      C$\beta$--sidechain RBF & $(N, K, 32D)$ & RBF from pseudo-C$\beta$ to sidechain atoms; design chains masked \\
      \midrule
      \multicolumn{3}{l}{\textit{Atom graph: node features}} \\
      Atom type & $(M, A)$ & One-hot over atom types \\
      Residue type & $(M, R)$ & One-hot over parent residue type \\
      Atom exists & $(M, 1)$ & 1 if atom resolved in structure \\
      \midrule
      \multicolumn{3}{l}{\textit{Atom graph: edge features}} \\
      RBF distance & $(E, D)$ & RBF over C$\alpha$-to-atom distance \\
      Euclidean distance & $(E, 1)$ & C$\alpha$-to-atom distance \\
      Residue index offset & $(E, 65)$ & Clamped relative residue index one-hot ($\pm 32$) \\
      Same-chain indicator & $(E, 1)$ & 1 if same chain \\
      \midrule
      \multicolumn{3}{l}{\textit{Atom graph: 3D coordinates (equivariant attention)}} \\
      Centroid positions & $(N, 3)$ & C$\alpha$ coordinates \\
      Atom positions & $(M, 3)$ & All-atom coordinates \\
      \bottomrule
      \end{tabular}%
      }
  \end{table}
    Node and edge features are derived from backbone atom coordinates (N, C$\alpha$, C, O) and include local geometric descriptors; shown in \Cref{tab:input_features}.

    \paragraph{All-atom equivariant GAT.}
    \renewcommand{\KwIn}[2]{\textbf{def} #1(#2):}

  \SetKwProg{Fn}{def}{:}{}
  \SetKwComment{Comment}{\# }{}
  \SetKw{KwReturn}{return}

  \SetKwFunction{EGATLayer}{EGATLayer}
  \SetKwFunction{Linear}{Linear}
  \SetKwFunction{Softmax}{Softmax}
  \SetKwFunction{LeakyReLU}{LeakyReLU}

  \newcommand{\Query}{$q$}
  \newcommand{\Key}{$k$}
  \newcommand{\Xcoord}{$x$}
  \newcommand{\Ycoord}{$y$}
  \providecommand{\EdgeIdx}{$\mathcal{E}$}
  \newcommand{\EdgeAttr}{$e$}
  \newcommand{\Hout}{$h^{\text{out}}$}
  \newcommand{\Xout}{$x^{\text{out}}$}

\begin{algorithm}[t!]
  \setlength{\algomargin}{0em}
  \DontPrintSemicolon
  \SetAlgoSkip{medskip}  

  \caption{Equivariant Graph Attention Layer}\label{algo:rednet_egat}

  \Comment{\Query, \Key: node features; \Xcoord, \Ycoord: coordinates; \EdgeIdx: edges; \EdgeAttr: edge features}

  \Fn{\EGATLayer{\Query, \Key, \Xcoord, \Ycoord, \EdgeIdx, \EdgeAttr}}{

    \Comment{Compute relative positions and distances}
    $(i, j) \gets $\EdgeIdx\;
    $z_{ij} \gets y_j - x_i$\;
    $d_{ij} \gets \|z_{ij}\|^2$\;

    \Comment{Build edge messages}
    $m_{ij} \gets \Linear([q_i \| k_j \| d_{ij} \| e_{ij}])$\;

    \Comment{Compute multi-head attention}
    $\alpha_{ij} \gets W^Q q_i + W^K k_j + W^E m_{ij}$\;
    $\alpha_{ij} \gets W^A \cdot \LeakyReLU(\alpha_{ij})$\;
    $\alpha_{ij} \gets \Softmax_{j \in \delta(i)}(\alpha_{ij})$ \Comment*[r]{over neighbors $j$}

    \Comment{Aggregate values}
    $v_{ij} \gets \Linear([q_i \| k_j \| d_{ij} \| e_{ij}])$\;
    $o_i \gets \sum_{j \in \delta(i)} \alpha_{ij} v_{ij}$\;
    \Hout $\gets \Linear([q_i \| o_i])$\;


    \KwReturn \Hout
  }
  \end{algorithm}
    The pseudo-algorithm is shown in \Cref{algo:rednet_egat}. The layer takes as input query and key node features ($q$, $k$), their corresponding coordinates ($x$, $y$), edge indices $\mathcal{E}$, and edge attributes $e$. Edge messages are
  constructed by concatenating source and target node features with squared pairwise distances and edge attributes, ensuring that the representation is invariant to global rotations and translations. Multi-head attention scores are computed
  using a GAT-style mechanism with LeakyReLU activation~\cite{brody2022gatv2}, allowing the model to learn which neighboring atoms are most relevant for each query node. Output node features are obtained by aggregating attention-weighted value vectors
  and projecting with a residual connection.

\subsubsection{Losses}
\paragraph{Causal language model objectives.} We train the model using a cross-entropy loss over amino acid tokens \cite{ingraham2019generative}. Given predicted logits $\hat{y}_i \in \mathbb{R}^{|\mathcal{V}|}$ and ground truth token $y_i$ at position $i$, the nodewise loss is defined as:
  $$
  \mathcal{L}_{\text{node}} = \sum_{m \in \mathcal{M}} w_m \cdot \frac{1}{|M_m|} \sum_{i \in M_m} \text{CE}(\hat{y}_i, y_i)
  $$
  where $\mathcal{M}$ denotes a set of binary masks (e.g., design site mask, prediction mask) with corresponding weights $w_m$, and $\text{CE}(\cdot, \cdot)$ is the cross-entropy loss.

\paragraph{Edgewise regularization.} To encourage the model to learn informative pairwise representations, we add an auxiliary edgewise cross-entropy loss. During training, we find this regularization to prevent overfitting. For each node $i$ and its $k$-nearest neighbors, the model predicts a joint token distribution over
  residue pairs. Given edgewise logits $\hat{z}_{ij} \in \mathbb{R}^{|\mathcal{V}|^2}$ and ground truth pair token $z_{ij} = y_i \cdot |\mathcal{V}| + y_j$, the edgewise loss is:
  $
  \mathcal{L}_{\text{edge}} = \frac{1}{|E|} \sum_{(i,j) \in E} \text{CE}(\hat{z}_{ij}, z_{ij})
  $
  where $E$ is the set of valid edges determined by the prediction mask. The total loss is $\mathcal{L} = \mathcal{L}_{\text{node}} + \lambda_{\text{edge}} \mathcal{L}_{\text{edge}}$, where $\lambda_{\text{edge}}$ controls the regularization strength. In our experiments $\lambda_{\text{edge}}$ is set to 1.

\subsection{Contrastive Decoding and Scoring}

 Contrastive decoding \cite{li2023contrastive,obrien2023contrastive} modifies the predicted logits at each timestep to amplify features that distinguish the on-target bound structure from the off-target bound structure. Here, the bound structure consists of the backbone coordinates of the
  complex and the side-chain coordinates of the target chain. Specifically, at timestep $t$, the modified logits are computed as:
  \begin{equation*}
      \ell(s_t) = (1 + \alpha) \log p(s_t | s_{<t}, r_{\texttt{on}}, x_{\texttt{on}}) - \alpha \log p(s_t | s_{<t}, r_{\texttt{off}}, x_{\texttt{off}})
  \end{equation*}
  where $s_t$ is the binder residue at position $t$, $s_{<t}$ denotes the previously sampled binder residues, $r_{\texttt{on}}$ and $r_{\texttt{off}}$ are the target sequences for the on-target and off-target complexes, $x_{\texttt{on}}$ and
  $x_{\texttt{off}}$ are the corresponding bound structures, and $\alpha \ge 0$ controls the strength of the contrastive penalty. When $\alpha = 0$, this reduces to standard decoding from the on-target distribution. As $\alpha$ increases, the
  model increasingly penalizes residue choices that are also favorable under the off-target context, thereby promoting specificity.

  To prevent the contrastive term from selecting low-probability tokens, we constrain the candidate set at timestep $t$ to tokens with sufficiently high probability under the on-target distribution:
  \begin{equation*}
  \mathcal{S}_t = \{s_t: p(s_t | s{<t}, r_{\texttt{on}}, x_{\texttt{on}}) \ge \beta \max_{s \in \mathcal{A}} p(s | s_{<t}, r_{\texttt{on}}, x_{\texttt{on}})\}
  \end{equation*}
  where $\mathcal{A}$ is the amino acid alphabet and $\beta \in [0, 1]$ is a truncation threshold. This adaptive truncation ensures that only plausible residues are considered, while allowing the contrastive term to discriminate among them. We
  then sample from the modified categorical distribution restricted to the candidate set:
  \begin{equation*}
  p_t(s) = \text{softmax}_{\mathcal{S}_t}(\ell(s))
  \end{equation*}
  The pseudocode is shown in \Cref{alg:contrastive_decoding}.
  \begin{algorithm}[t]
  \caption{Contrastive Decoding for Binder Sequence Design}
  \label{alg:contrastive_decoding}
  \Fn{\textnormal{ContrastiveDecode}($r_{\texttt{on}}, x_{\texttt{on}}, r_{\texttt{off}}, x_{\texttt{off}}, \alpha, \beta, \tau, L, \mathcal{A}$)}{
      $s \gets [\texttt{mask}]^L$ \Comment{Initialize binder sequence with all mask tokens}
      
      \For{$t = 1, \ldots, L$}{
          $\ell(a) \gets (1 + \alpha) \log p(a \mid s_{<t}, r_{\texttt{on}}, x_{\texttt{on}}) - \alpha \log p(a \mid s_{<t}, r_{\texttt{off}}, x_{\texttt{off}})$\;
          
          $p_{\max} \gets \max_{a \in \mathcal{A}}\, p(a \mid s_{<t}, r_{\texttt{on}}, x_{\texttt{on}})$\;
          
          $\mathcal{S}_t \gets \{a \in \mathcal{A} : p(a \mid s_{<t}, r_{\texttt{on}}, x_{\texttt{on}}) \ge \beta \cdot p_{\max}\}$\;
          
          $s_t \sim \mathrm{softmax}_{\mathcal{S}_t}(\ell(a) / \tau)$\; \Comment{Sample with temperature $\tau$}
          
          Set $s[t]$ to $s_t$\;
      }
      \KwReturn $s$\;
  }
  \end{algorithm}

  This decoding framework also enables direct sampling of sequences that optimize the binding free energy $\Delta G$, which relates to the folding free energies as follows \cite{gilson1997statistical}:
  \begin{equation*}
    \Delta G = \Delta G_{\texttt{complex}} - \Delta G_{\texttt{binder}} - \Delta G_{\texttt{target}}
  \end{equation*}
   where $\Delta G$ is the binding free energy, $\Delta G_{\texttt{complex}}$ is the folding free energy of the bound complex, and $\Delta G_{\texttt{binder}}$ and $\Delta G_{\texttt{target}}$ are the folding free energies of the unbound binder
  and target chains, respectively. Multiple studies \cite{dieckhaus2024transfer} have  demonstrated that folding free energy correlates with the log-likelihood from fixed-backbone sequence design models. Therefore, the right-hand side can be approximated as:
  \begin{equation*}
    \Delta G \approx \log p(s, r | x_{\texttt{bound}}) - \log p(s | x_{\texttt{binder}}) - \log p(r | x_{\texttt{target}})
  \end{equation*}
  where $s$ and $r$ are the binder and target chain sequences, $x_{\texttt{bound}}$ is the structure of the bound complex, and $x_{\texttt{binder}}$ and $x_{\texttt{target}}$ are the structures of the unbound binder and target chains,
  respectively. Since the target sequence $r$ is fixed during binder design, we can omit terms that do not depend on $s$, yielding the following contrastive decoding formulation:
  \begin{equation*}
      \ell(s_t) = (1 + \alpha) \log p(s_t | s_{<t}, r, x_{\texttt{bound}}) - \alpha \log p(s_t | s_{<t}, x_{\texttt{binder}})
  \end{equation*}
  This formulation encourages the model to select residues that are favorable in the bound state but unfavorable in the unbound state, thereby approximating the thermodynamic preference for complex formation. The same candidate set truncation
  and sampling procedure described above are applied to this formulation.

  The two contrastive decoding formulations share the same mathematical form: in both cases, the model amplifies residue choices that are favorable under the on-target context and unfavorable under an alternative context. In the specificity
  formulation, the alternative context is the off-target bound structure; in the affinity formulation, the alternative context is the unbound binder structure.

\subsubsection{Scoring}
We define a set of scoring metrics to evaluate designed sequences for binding.
  Straightforward evaluations of the autoregressive model, $\texttt{ll}$ and $\texttt{ll\_global}$, measure respectively the average log-likelihood of the designed binder sequence and the full complex sequence given the bound complex structure,
  assessing sequence structure compatibility. $\texttt{ll\_mt}$ restricts the average to mutated positions only. $\texttt{ll\_ref}$ further refines this by computing the log-likelihood difference between the designed and wild-type residues at
  mutated positions, directly measuring whether the mutations are predicted to be more favorable than the original sequence.

  The contrastive scores, $\texttt{ll\_cd}$ and $\texttt{ll\_cd\_ref}$, approximate binding free energy by subtracting the unbound binder log-likelihood from the bound complex log-likelihood, mirroring the thermodynamic cycle formulation.
  $\texttt{ll\_cd}$ captures the overall binding preference across the binder chain, while $\texttt{ll\_cd\_ref}$ focuses on mutated positions and uses the wild-type as a reference, providing a mutation-specific estimate of the change in binding
   affinity relative to the original sequence. Concurrent works \cite{dutton2024improving,frellsen2025zero,deng2025predicting} have also explored similar scoring metrics and demonstrated improved correlation with experimental measurements of stability.
   
\begin{align*}
      \texttt{ll} &= \frac{1}{N_{\texttt{binder}}} \sum_{i=1}^{N_{\texttt{binder}}} \ell_i (a_i)\\
      \texttt{ll\_global} &= \frac{1}{N_{\texttt{complex}}} \sum_{i=1}^{N_{\texttt{complex}}} \ell_i(a_i) \\
      \texttt{ll\_mt} &= \frac{1}{|\mathcal{M}|} \sum_{i \in \mathcal{M}} \ell_i(a_i) \\
      \texttt{ll\_ref} &= \frac{1}{|\mathcal{M}|} \sum_{i \in \mathcal{M}} \left( \ell_i(a_i) - \ell_i(a_i^{\text{wt}}) \right) \\
      \texttt{ll\_cd} & = \frac{1}{N_{\texttt{binder}}} \sum_{i=1}^{N_{\texttt{binder}}} \ell_i (a_i) - \frac{1}{N_{\texttt{binder}}} \sum_{i=1}^{N_{\texttt{binder}}} \ell_i^u (a_i) \\
      \texttt{ll\_cd\_ref} & = \frac{1}{|\mathcal{M}|} \sum_{i \in \mathcal{M}} \left( \ell_i(a_i) - \ell_i(a_i^{\text{wt}}) \right) - \frac{1}{|\mathcal{M}|} \sum_{i \in \mathcal{M}} \left( \ell_i^u(a_i) - \ell_i^u(a_i^{\text{wt}}) \right)
  \end{align*}
  where $\ell_i = \log p_i$ is the predicted log-probability at position $i$ given the bound complex structure and the target sequence, $\ell_i^u$ is the predicted log-probability at position $i$ given the unbound binder structure, $a_i$ and
  $a_i^{\text{wt}}$ are the designed and wild-type residue types at position $i$, respectively, $N_{\texttt{binder}}$ is the sequence length of the binder chain, $N_{\texttt{complex}}$ is the total sequence length of the complex, and
  $\mathcal{M} = \{i : a_i \neq a_i^{\text{wt}}\}$ is the set of mutated positions in the binder chain.
\subsection{Datasets}

\subsubsection{Training Set}
  We use structures released before 2023-01-01 from Protein Data Bank (PDB) \cite{berman2000protein} as the training and validation set. We filter out structures with more than 20 polymer chains, resolutions worse than 5 \AA, and experimental methods other than X-ray diffraction and electron microscopy. We retain only polypeptide(L) chains with 10–5000 residues, fewer than 10\%
  unknown residues

  \paragraph{Validation Set.}
  We use the subset of structures that are released between 2022-05-01 and 2022-12-31 as the validation set. We remove similar chains using the same procedure as the PDB test sets described below.

  \subsubsection{Test Datasets}

  \paragraph{Low-Homology PDB Test Set.}
  We select structures released between 2023-01-01 and 2023-12-31. To prevent potential sequence leakage, we search test sequences against training sequences using MMseqs2 \cite{stebinegger2017mmseqs2} and retain only (design) chains with e-value $> 1$, which is typically stricter than commonly used 30\% sequence identity threshold. To reduce overrepresentation
  of certain protein families, we cluster the remaining test sequences at $40\%$ sequence identity using MMseqs2. We categorize each structure by interface type based on the similarity of interacting chains: monomer (no interface), homodimer
  (two similar chains), or heterodimer (two dissimilar chains). For each interface type, we select one representative from each cluster and randomly sample $300$ monomers, $150$ homodimers, and $150$ heterodimers to create a balanced test set of
   $600$ structures.

  To test heterodimer self-consistency, we select heterodimers whose total number of residues is at most $500$, due to computational constraints. This filtering retains $107$ samples.

  \paragraph{Selective Binder Test Set.}\label{subsec:rednet_sel}
  
Heterodimeric protein complexes are curated from all protein chains in the PDB (released prior to 2025-04-14). Two chains form an interacting heterodimer if their minimum C$\alpha$--C$\alpha$ distance is $\le 10$\,\AA\,within a bioassembly and they share less than $90\%$
  sequence identity as determined by MMseqs2 clustering.

  We filter chains to retain those between 20 and 500 residues in length, with fewer than $10\%$ unknown amino acids and no single amino acid type exceeding $50\%$ of the sequence. For each chain cluster, we identify all its interacting
  heterodimers, retaining only clusters with interacting partners from at least two PDB entries. To prevent promiscuous clusters that interact non-specifically with many targets from dominating the test set, we exclude clusters with interacting
  partners from more than 30 PDB entries. This step retains 991 unique clusters and 3,246 interacting heterodimers.

  Within each chain cluster (designated as the binder cluster, i.e., the chain to be redesigned), we randomly select one interacting heterodimer as the on-target pair. Each remaining heterodimer in the cluster serves as a candidate off-target
  pair. We structurally align the binder chains of the on-target and off-target pairs using TM-align (these belong to the same cluster), and separately align their respective target chains. We discard an off-target candidate if any of the
  following conditions hold: (1) the target chains share $100\%$ sequence identity (i.e., the on-target and off-target receptors are identical), (2) the binder chain alignment has coverage below $90\%$, (3) the binder chain alignment has
  sequence identity below $90\%$, or (4) the binder chain alignment RMSD exceeds $2.5$\,\AA. After filtering, 691 unique binder clusters remain.

  To assess off-target interaction difficulty, we define a difficulty score based on the Jaccard similarity of residue contact pairs (C$\alpha$--C$\alpha$ $\le 10$\,\AA) between the on-target and off-target complexes:
  \begin{equation*}
      \text{Difficulty} = \text{JaccardSimilarity}(\mathcal{C}_{\text{on-target}}, \mathcal{C}_{\text{off-target}})
  \end{equation*}
  where $\mathcal{C}_{\text{on-target}}$ and $\mathcal{C}_{\text{off-target}}$ denote the sets of inter-chain residue contact pairs in the on-target and off-target complexes, respectively. Higher scores indicate more shared interface contacts,
  making selective design more challenging.

  We retain off-target interactions with Difficulty $< 0.9$, yielding 656 unique binder clusters. Within each cluster, we select the off-target with the lowest Jaccard similarity (i.e., the most dissimilar interface), producing 656 non-redundant
   on-/off-target pairs. Evaluating each pair requires two AlphaFold3 cofolding passes (one for on-target, one for off-target), each with 10 recycles and 5 diffusion samples, followed by Rosetta relaxation (3 repeats) for both predicted
  complexes. Given the computational cost, we uniformly sample 180 on-/off-target pairs for our benchmark.

  \subsubsection{Benchmarking Methods}

  We compare our method against widely adopted fixed-backbone sequence design methods: ESM-IF~\cite{hsu2022learingif}, ProteinMPNN~\cite{dauparas2022robust}, and PiFold~\cite{gao2023pifold}.

\section{Results}

\subsection{All-atom graph transformer improves sequence recovery of monomeric and dimeric structures}

We benchmark RedNet against widely adopted fixed backbone design models, including ESM-IF, ProteinMPNN, and PiFold. We evaluate native sequence recovery (NSR), wild-type log-likelihood (LL), and perplexity (PPL) across monomers, homodimers, and heterodimers (\cref{tab:seq_recovery}). ESM-IF and PiFold are trained without noise augmentation, while ProteinMPNN and RedNet are compared across noise levels $\sigma \in \{0, 0.02, 0.1, 0.2, 0.3\}$ to assess robustness to backbone coordinate perturbations.

\begin{table}[t]
\caption{Performance comparison on monomers, homodimers, and heterodimers. $\sigma$: backbone coordinate noise level. NSR: Native Sequence Recovery. LL: Log-Likelihood. PPL: Perplexity. For RedNet and ProteinMPNN, we test performance at different noise levels $\sigma \in \{0.02, 0.1, 0.2, 0.3\}$. ESM-IF and PiFold are tested at $\sigma=0$.}
\label{tab:seq_recovery}
\centering
\small
\resizebox{\textwidth}{!}{
\begin{tabular}{l c c c c c c c c c c}
\toprule
\multirow{2}{*}{Model} & \multirow{2}{*}{$\sigma$} & \multicolumn{3}{c}{Monomer} & \multicolumn{3}{c}{Homodimer} & \multicolumn{3}{c}{Heterodimer} \\
& & NSR & LL & PPL & NSR & LL & PPL & NSR & LL & PPL \\
\midrule
ESM-IF & 0 & 0.38 & -1.85 & \textbf{7.33} & 0.43 & -1.35 & 4.29 & 0.33 & -2.21 & 13.39 \\
PiFold & 0 & 0.40 & -1.92 & 9.65 & 0.45 & -1.73 & 6.10 & 0.35 & -2.08 & 9.16 \\
RedNet (ours) & 0 & \textbf{0.43} & \textbf{-1.74} & 9.15 & \textbf{0.49} & \textbf{-1.28} & \textbf{3.86} & \textbf{0.43} & \textbf{-1.76} & \textbf{6.58} \\
\midrule
ProteinMPNN & 0.02 & 0.36 & \textbf{-1.91} & \textbf{7.60} & 0.42 & -1.55 & 5.12 & 0.37 & -2.00 & 8.32 \\
RedNet (ours) & 0.02 & \textbf{0.37} & -1.92 & 8.30 & \textbf{0.43} & \textbf{-1.44} & \textbf{4.64} & \textbf{0.39} & \textbf{-1.91} & \textbf{7.51} \\
\midrule
ProteinMPNN & 0.1 & \textbf{0.33} & -2.02 & \textbf{8.36} & \textbf{0.40} & -1.64 & 5.58 & 0.34 & -2.11 & 9.25 \\
RedNet (ours) & 0.1 & \textbf{0.33} & \textbf{-2.01} & 8.68 & \textbf{0.40} & \textbf{-1.53} & \textbf{5.07} & \textbf{0.35} & \textbf{-2.04} & \textbf{8.67} \\
\midrule
ProteinMPNN & 0.2 & 0.31 & -2.08 & \textbf{8.84} & 0.37 & -1.70 & 5.90 & 0.33 & -2.14 & 9.46 \\
RedNet (ours) & 0.2 & \textbf{0.32} & \textbf{-2.06} & 8.91 & \textbf{0.38} & \textbf{-1.60} & \textbf{5.38} & \textbf{0.34} & \textbf{-2.08} & \textbf{8.84} \\
\midrule
ProteinMPNN & 0.3 & \textbf{0.30} & -2.12 & \textbf{9.18} & \textbf{0.36} & -1.76 & 6.28 & 0.31 & -2.21 & 10.12 \\
RedNet (ours) & 0.3 & \textbf{0.30} & \textbf{-2.11} & 9.23 & \textbf{0.36} & \textbf{-1.63} & \textbf{5.54} & \textbf{0.32} & \textbf{-2.13} & \textbf{9.27} \\
\bottomrule
\end{tabular}
}
\end{table}

\paragraph{Monomers.} On monomeric structures without noise augmentation, RedNet achieves the highest sequence recovery (NSR $=$ 0.43) compared to ESM-IF (0.38), PiFold (0.40), and ProteinMPNN at $\sigma = 0.02$ (0.36). RedNet achieves the most favorable wild-type log-likelihood (LL $=$ $-$1.74 vs.\ $-$1.85 for ESM-IF, $-$1.92 for PiFold), indicating better calibration of sequence probabilities on native sequences. At matched noise levels, RedNet consistently outperforms ProteinMPNN: at $\sigma = 0.02$, RedNet achieves NSR $=$ 0.37 versus ProteinMPNN's 0.36; at $\sigma = 0.1$, both models converge to NSR $=$ 0.33, though RedNet maintains marginally better log-likelihood ($-$2.01 vs.\ $-$2.02). Performance degradation with increasing noise is comparable between models, with NSR decreasing from 0.43 to 0.30 for RedNet across the noise range tested.

\paragraph{Homodimers.} RedNet demonstrates strong performance on homodimers, achieving NSR $=$ 0.49 at $\sigma = 0$, compared to ESM-IF (0.43), PiFold (0.45), and ProteinMPNN at $\sigma = 0.02$ (0.42). RedNet also yields the lowest perplexity (PPL $=$ 3.86 vs.\ 4.29 for ESM-IF, 6.10 for PiFold), suggesting higher confidence in native sequence predictions at symmetric protein--protein interfaces. The log-likelihood gap is substantial: RedNet achieves LL $=$ $-$1.28 compared to $-$1.35 for ESM-IF and $-$1.73 for PiFold. Across noise levels, RedNet maintains advantages over ProteinMPNN: at $\sigma = 0.1$, RedNet achieves NSR $=$ 0.40 and PPL $=$ 5.07 versus ProteinMPNN's NSR $=$ 0.40 and PPL $=$ 5.58; at $\sigma = 0.3$, RedNet achieves LL $=$ $-$1.63 versus $-$1.76 for ProteinMPNN.

\paragraph{Heterodimers.} The performance gap is most significant on heterodimeric interfaces, which are most relevant to one-sided interface design. At $\sigma = 0$, RedNet achieves NSR $=$ 0.43, outperforming ESM-IF (0.33), PiFold (0.35), and ProteinMPNN at $\sigma = 0.02$ (0.37). This represents a 10 percentage point improvement over ESM-IF (30\% relative) and a 6 percentage point improvement over ProteinMPNN at $\sigma = 0.02$ (16\% relative). The perplexity gap is large: RedNet achieves PPL $=$ 6.58 compared to ESM-IF (PPL $=$ 13.39), PiFold (PPL $=$ 9.16), and ProteinMPNN (PPL $=$ 8.32). Log-likelihood differences follow similar patterns.

\paragraph{Robustness to coordinate noise.} We compare RedNet and ProteinMPNN across increasing noise levels to assess robustness to coordinate perturbations. Both models show expected degradation with increasing noise: RedNet's heterodimer NSR decreases from 0.43 ($\sigma = 0$) to 0.32 ($\sigma = 0.3$), while ProteinMPNN's decreases from 0.37 ($\sigma = 0.02$) to 0.31 ($\sigma = 0.3$). RedNet maintains consistent advantages on dimer interfaces across all noise levels. At $\sigma = 0.2$, RedNet achieves heterodimer PPL $=$ 8.84 versus ProteinMPNN's PPL $=$ 9.46. At $\sigma = 0.3$, RedNet achieves PPL $=$ 9.27 compared to ProteinMPNN's PPL $=$ 10.12. However, the sequence recovery rate gap between RedNet and ProteinMPNN decreases as the noise level increases. It suggests that RedNet, which captures side-chain information, is more sensitive to coordinate perturbation compared to backbone-only models like ProteinMPNN.

   \subsection{Contrastive scoring improves zero-shot binding affinity prediction}

  \begin{table}[ht!]
\caption{Zero-shot binding affinity prediction on SKEMPI v2.0. Spearman correlations ($\rho$), Kendall's $\tau$, and NDCG scores averaged across assays in SKEMPI v2.0. Contrastive scoring methods include standard binder log-likelihood (\texttt{ll}), global complex log-likelihood (\texttt{global}), mutated-residue log-likelihood (\texttt{mt}), reference-normalized (\texttt{ref}), contrastive log-likelihood (\texttt{cd\_ll}), and contrastive reference-normalized (\texttt{cd\_ll\_ref}). $\sigma$: backbone coordinate noise level.}
\label{tab:skempi_affinity}
\centering
\small
\resizebox{\textwidth}{!}{
\begin{tabular}{l c c c c c c c c c c c c c c c c c c c}
\toprule
\multirow{2}{*}{Model} & \multirow{2}{*}{$\sigma$} & \multicolumn{6}{c}{Spearman $\rho$} & \multicolumn{6}{c}{Kendall $\tau$} & \multicolumn{6}{c}{NDCG} \\
& & ll & global & mt & ref & cd\_ll & cd\_ll\_ref & ll & global & mt & ref & cd\_ll & cd\_ll\_ref & ll & global & mt & ref & cd\_ll & cd\_ll\_ref \\
\midrule
ESM-IF & 0 & 0.16 & 0.24 & 0.20 & 0.20 & 0.15 & 0.15 & 0.12 & 0.18 & 0.14 & 0.14 & 0.15 & 0.10 & 0.77 & 0.79 & 0.79 & 0.78 & 0.77 & 0.77 \\
PiFold & 0 & 0.17 & -0.17 & -0.03 & -0.12 & 0.15 & 0.00 & 0.13 & -0.13 & -0.02 & -0.09 & 0.11 & 0.00 & 0.78 & 0.67 & 0.72 & 0.69 & 0.75 & 0.71 \\
ProteinMPNN & 0.02 & 0.17 & 0.17 & \textbf{0.23} & \textbf{0.26} & 0.10 & 0.24 & 0.12 & 0.12 & \textbf{0.17} & \textbf{0.18} & 0.07 & 0.18 & 0.78 & 0.78 & 0.80 & 0.80 & 0.76 & \textbf{0.78} \\
RedNet & 0 & 0.18 & 0.23 & 0.21 & 0.22 & 0.23 & 0.26 & 0.13 & 0.17 & 0.15 & 0.16 & 0.17 & 0.18 & \textbf{0.79} & \textbf{0.80} & 0.80 & 0.80 & 0.78 & \textbf{0.78} \\
RedNet & 0.02 & \textbf{0.21} & \textbf{0.26} & 0.22 & 0.24 & \textbf{0.26} & \textbf{0.28} & \textbf{0.15} & \textbf{0.19} & 0.16 & \textbf{0.18} & \textbf{0.19} & \textbf{0.20} & \textbf{0.79} & \textbf{0.80} & \textbf{0.81} & \textbf{0.81} & \textbf{0.79} & \textbf{0.78} \\
\bottomrule
\end{tabular}
}
\end{table}

  Previous works have demonstrated the effectiveness of contrastive scoring for zero-shot monomer stability prediction using ProteinMPNN and ESM-IF, but have not empirically validated improvements in zero-shot binding affinity prediction
   in the context of binder design. We evaluate fixed-backbone design models on predicting binding affinity changes upon mutation in a zero-shot manner using the SKEMPI v2.0 dataset, benchmarking RedNet against ProteinMPNN, ESM-IF, and PiFold
  (\cref{tab:skempi_affinity}). Since PiFold and ESM-IF do not have released models trained with high noise levels (over 0.1\,\AA), we benchmark all models at lower noise levels (0 and 0.02\,\AA) for fairness.

  \paragraph{Overall performance.} RedNet ($\sigma{=}0.02$) with contrastive scoring methods \texttt{cd\_ll} and \texttt{cd\_ll\_ref} consistently outperforms all other model, scoring combinations across Spearman's $\rho$, Kendall's $\tau$, and
  NDCG, achieving the highest Spearman correlation of 0.28 (\texttt{cd\_ll\_ref}), Kendall's $\tau$ of 0.20 (\texttt{cd\_ll\_ref}), and NDCG of 0.81 (\texttt{mt}, \texttt{ref}). This indicates that RedNet's likelihood estimates using different scoring methods are better
  aligned with binding affinity than competing methods.

  \paragraph{RedNet consistently outperforms other models.} RedNet ($\sigma{=}0.02$) provides the highest Spearman correlations in \texttt{ll} (0.21 vs.\ 0.17 for ProteinMPNN and ESM-IF) and \texttt{global} (0.26 vs.\ 0.24 for ESM-IF and 0.17
  for ProteinMPNN), with more pronounced gains in Kendall's $\tau$, where it leads in five of six scoring methods. The exceptions are \texttt{mt} and \texttt{ref}, where ProteinMPNN achieves higher Spearman correlations (0.23 and 0.26,
  respectively, versus 0.22 and 0.24 for RedNet), though RedNet matches or exceeds ProteinMPNN on these metrics in Kendall's $\tau$ and NDCG. These results suggest that RedNet's all-atom featurization provides a consistent advantage over
  backbone-only models for zero-shot binding affinity prediction.

  \paragraph{Contrastive scoring improves RedNet but not all models.} The contrastive methods \texttt{cd\_ll} and \texttt{cd\_ll\_ref} consistently improve RedNet at both noise levels across all three metrics. In contrast, contrastive scoring
  degrades ProteinMPNN (Spearman drops from 0.26 with \texttt{ref} to 0.24 with \texttt{cd\_ll\_ref}, and from 0.23 with \texttt{mt} to 0.10 with \texttt{cd\_ll}) and ESM-IF (Spearman drops from 0.24 with \texttt{global} to 0.15 with
  \texttt{cd\_ll} and \texttt{cd\_ll\_ref}). While contrastive scoring does improve PiFold, its baseline correlations are near zero (Spearman \texttt{mt}~$= -0.03$, \texttt{ref}~$= -0.12$), suggesting that PiFold fails at zero-shot binding
  affinity prediction altogether and is not a meaningful basis for comparing scoring methods. We hypothesize that RedNet benefits from contrastive scoring because it is trained on both bound complexes and monomers and leverages detailed all-atom
   structure at interfaces, making it more sensitive to differences between bound and unbound states.

  \paragraph{Effect of noise augmentation.} Comparing RedNet at $\sigma{=}0.02$ versus $\sigma{=}0$, we observe consistent improvements. For \texttt{ll}, Spearman's $\rho$ increases from 0.18 to 0.21; for \texttt{global}, from 0.23 to 0.26; for
  \texttt{cd\_ll}, from 0.23 to 0.26; and for \texttt{cd\_ll\_ref}, from 0.26 to 0.28. Similar gains appear in Kendall's $\tau$ and NDCG. This suggests that training with backbone coordinate noise improves the model's sensitivity to binding
  affinities, despite yielding lower native sequence recovery.

  Overall, contrastive scoring with RedNet trained on all-atom structures with noise augmentation yields the best zero-shot binding affinity predictions, motivating the development of
  decoding methods that can improve contrastive scores and thus the affinity of designed binders.
  
\subsection{Contrastive decoding improves structural self-consistency of binders}
  \begin{table}[ht!]
    \caption{Heterodimer self-consistency results on all 107 targets. $\sigma$: backbone coordinate noise level (\AA). SR: Success Rate (0--100\%), defined as pTM $>$ 0.55, ipTM $>$ 0.5, and Dsn pLDDT $>$ 80, following BindCraft. Dsn pLDDT:
  AlphaFold3 predicted LDDT for the design chain (0--100). ipTM: interface predicted Template Modeling score (0--1). pTM: AlphaFold3 predicted TM-score of the complex (0--1). Tgt pLDDT: AlphaFold3 predicted LDDT for the target chain (0--100).
  RedNet-CD uses contrastive decoding with $\alpha=1$, $\beta=0.9$. All models are sampled at temperature $=0.001$. Higher is better for all metrics. \textbf{Bold}: best; \underline{underline}: second best.}
    \label{tab:self_consistency_all}
    \centering
    \small
    \resizebox{\textwidth}{!}{
    \begin{tabular}{l c c c c c c}
    \toprule
    Model & $\sigma$ & SR $\uparrow$ & Dsn pLDDT $\uparrow$ & ipTM $\uparrow$ & pTM $\uparrow$ & Tgt pLDDT $\uparrow$ \\
    \midrule
    Native & 0 & \underline{29\%} & 51.31 & \textbf{0.39} & \underline{0.53} & \textbf{62.61} \\
    ProteinMPNN & 0.02 & \textbf{30\%} & \textbf{55.12} & 0.36 & \underline{0.53} & 62.21 \\
    ESM-IF & 0 & 27\% & \underline{53.77} & \underline{0.38} & \underline{0.53} & \underline{62.50} \\
    PiFold & 0 & 28\% & 50.29 & 0.36 & 0.52 & 61.71 \\
    RedNet & 0.02 & 25\% & 53.46 & 0.37 & \textbf{0.54} & 62.45 \\
    RedNet-CD & 0.02 & \textbf{30\%} & 53.12 & 0.37 & \underline{0.53} & 61.86 \\
    \bottomrule
    \end{tabular}
    }
    \end{table}
  \begin{table}[ht!]
    \caption{Self-consistency performance on high-quality subset (44 targets). $\sigma$: backbone coordinate noise level (\AA). SR: Success Rate (0--100\%), defined as pTM $>$ 0.55, ipTM $>$ 0.5, and Dsn pLDDT $>$ 80, following BindCraft. Dsn
  pLDDT: AlphaFold3 predicted LDDT for the design chain (0--100). ipTM: interface predicted Template Modeling score (0--1). pTM: AlphaFold3 predicted TM-score of the complex (0--1). Tgt pLDDT: AlphaFold3 predicted LDDT for the target chain
  (0--100). RedNet-CD uses contrastive decoding with $\alpha=1$, $\beta=0.9$. All models are sampled at temperature $=0.001$. Higher is better for all metrics. \textbf{Bold}: best; \underline{underline}: second best.}
    \label{tab:self_consistency_hq}
    \centering
    \small
    \resizebox{\textwidth}{!}{
    \begin{tabular}{l c c c c c c}
    \toprule
    Model & $\sigma$ & SR $\uparrow$ & Dsn pLDDT $\uparrow$ & ipTM $\uparrow$ & pTM $\uparrow$ & Tgt pLDDT $\uparrow$ \\
    \midrule
    Native & 0 & \textbf{68\%} & \textbf{68.24} & \textbf{0.61} & \textbf{0.78} & \textbf{86.76} \\
    ProteinMPNN & 0.02 & 59\% & 67.48 & 0.55 & 0.75 & 84.98 \\
    ESM-IF & 0 & 61\% & 67.35 & \underline{0.59} & \underline{0.77} & \underline{85.75} \\
    PiFold & 0 & \underline{64\%} & 62.91 & 0.56 & 0.76 & 85.02 \\
    RedNet & 0.02 & 57\% & 66.05 & 0.56 & 0.76 & 84.93 \\
    RedNet-CD & 0.02 & \textbf{68\%} & \underline{67.71} & 0.57 & 0.75 & 84.88 \\
    \bottomrule
    \end{tabular}
    }
    \end{table}

  In protein binder design applications \cite{pacesa2025one}, confidence metrics including pLDDT, ipTM, and pTM predicted by AlphaFold3 are used to filter designs and are shown to effectively guide wet-lab experiments. We withhold MSAs from AlphaFold3 to prevent artificial confidence inflation from evolutionary information, relying instead on target templates. To
  simulate realistic de novo design scenarios, we adopt the default success criteria from BindCraft.

  We benchmark RedNet against ESM-IF, ProteinMPNN, and PiFold on heterodimer design tasks across two datasets: all 107 heterodimeric targets (\Cref{tab:self_consistency_all}) and a high-confidence subset of 44 targets where the
  AlphaFold3-predicted pLDDT of the native target chain exceeds 70 (\Cref{tab:self_consistency_hq}).

  \paragraph{All targets.} On the full 107 heterodimers, all methods face significant challenges: native sequences achieve only 29\% success rate with mean ipTM of 0.39. ProteinMPNN achieves the highest design chain confidence (Dsn pLDDT $=$
  55.12), followed by ESM-IF (53.77), RedNet (53.46), and RedNet-CD (53.12), all above native sequences (51.31); only PiFold (50.29) underperforms. However, improved monomer confidence does not translate to superior heterodimer self-consistency:
   ProteinMPNN and RedNet-CD both achieve 30\% success rate.

  RedNet-CD (SR $=$ 30\%) achieves substantial gains over RedNet with standard ancestral sampling (SR $=$ 25\%), both at temperature 0.001. Success rate is a composite measure of design chain stability (Dsn pLDDT), interface quality (ipTM), and
  overall complex confidence (pTM), reflecting the dual objective of binder redesign: jointly optimizing interface affinity and design chain stability. Contrastive decoding achieves this dual goal more effectively than standard sampling, despite
   slightly lower individual confidence scores.

  \paragraph{High-quality targets.} We note that on the full 107-target set, AlphaFold3 produces low-confidence predictions for many targets (average 62.61 pLDDT of the target chains), indicating that the evaluation may be bottlenecked by structure prediction quality rather
  than design quality alone. To disentangle design capability from structure prediction error, we analyze a subset of 44 complexes where native sequences yield high-confidence predictions (pLDDT $>$ 70). RedNet-CD achieves 68\% success
  rate, matching native sequences and outperforming all other methods: ProteinMPNN (59\%), ESM-IF (61\%), and PiFold (64\%) with relative improvements of 15\%, 11\%, and 6\%, respectively.

  Contrastive decoding significantly improves RedNet's heterodimer self-consistency, increasing success rate from 57\% (standard sampling) to 68\%, an absolute improvement of 11 percentage points (19\% relative).

 \paragraph{Comparison with ProteinMPNN on high-quality targets.} ProteinMPNN produces competitive monomer confidence (Dsn pLDDT $=$ 67.48 vs.\ 67.71 for RedNet-CD) for designed binder chains, yet RedNet-CD achieves a substantially higher
  success rate (68\% vs.\ 59\%, a 15\% relative gain) for the complete complexes. This suggests that RedNet's designs are more consistently above the multi-objective success thresholds (pLDDT $>$ 80, pTM $>$ 0.55, ipTM $>$ 0.5), even when mean
  scores are comparable. The side-chain context and contrastive decoding in RedNet appear to provide critical information for optimizing interface interactions while maintaining binder chain stability, achieving native-level heterodimer
  self-consistency where backbone-only models and standard ancestral sampling fall short.

\begin{table}[ht!]
    \caption{Energetics and geometric properties of designed binders. Binding Score (REU): Rosetta binding score. Int SC (0--1): interface shape complementarity. Int Packstat (0--1): interface packing statistic. Int dG (REU): interface free
  energy change. Int dSASA (\AA$^2$): interface buried solvent-accessible surface area. REU: Rosetta Energy Units. Due to Rosetta relaxation failures, we analyze 91 of 107 heterodimers that are successfully relaxed for all methods. RedNet-Ens
  combines RedNet and RedNet-CD by selecting the design with the best binding score. \textbf{Bold}: best; \underline{underline}: second best.}
    \label{tab:energetics}
    \centering
    \small
    \resizebox{\textwidth}{!}{
    \begin{tabular}{l c c c c c c}
    \toprule
    Model & $\sigma$ & Binding Score $\downarrow$ & Int SC $\uparrow$ & Int Packstat $\uparrow$ & Int dG $\downarrow$ & Int dSASA $\uparrow$ \\
    \midrule
    Native & 0 & $-$172.42 & 0.65 & 0.53 & $-$53.35 & 1918.33 \\
    ProteinMPNN & 0.02 & \underline{$-$184.89} & \underline{0.66} & \textbf{0.55} & $-$46.98 & 1682.12 \\
    ESM-IF & 0 & $-$181.27 & 0.64 & 0.53 & $-$54.97 & \underline{1947.75} \\
    PiFold & 0 & $-$179.55 & \textbf{0.67} & 0.53 & \underline{$-$55.19} & 1870.66 \\
    RedNet & 0.02 & $-$179.88 & \underline{0.66} & \underline{0.54} & $-$52.49 & 1868.29 \\
    RedNet-CD & 0.02 & $-$182.26 & \underline{0.66} & \underline{0.54} & $-$54.47 & 1894.69 \\
    RedNet-Ens & 0.02 & \textbf{$-$188.04} & \textbf{0.67} & \textbf{0.55} & \textbf{$-$56.66} & \textbf{1965.96} \\
    \bottomrule
    \end{tabular}
    }
    \end{table}
  \begin{table}[ht!]
    \caption{Hydrophobicity and hydrogen-bond properties of designed interfaces. Surf Hydro (0--1): surface hydrophobicity. Int Nres: number of interface residues. Int HBonds: number of interface hydrogen bonds. Int HBond \%: percentage of
  interface residues involved in hydrogen bonds. Int dUnsat HB: number of unsatisfied interface hydrogen bonds. Int dUnsat HB \%: percentage of unsatisfied interface hydrogen bonds. Due to Rosetta relaxation failures, we analyze 91 of 107
  heterodimers that are successfully relaxed for all methods. RedNet-Ens combines RedNet and RedNet-CD by selecting the design with the best binding score. \textbf{Bold}: best; \underline{underline}: second best.}
    \label{tab:hydrophobicity}
    \centering
    \small
    \resizebox{\textwidth}{!}{
    \begin{tabular}{l c c c c c c c}
    \toprule
    Model & $\sigma$ & Surf Hydro $\downarrow$ & Int Nres $\uparrow$ & Int HBonds $\uparrow$ & Int HBond \% $\uparrow$ & Int dUnsat HB $\downarrow$ & Int dUnsat HB \% $\downarrow$ \\
    \midrule
    Native & 0 & \textbf{0.43} & \underline{20.10} & 6.97 & 46.15 & 2.98 & 17.90 \\
    ProteinMPNN & 0.02 & \textbf{0.43} & 17.34 & 5.44 & 40.30 & \underline{2.52} & 17.62 \\
    ESM-IF & 0 & 0.46 & 19.92 & 6.51 & 43.19 & 2.78 & 17.26 \\
    PiFold & 0 & 0.47 & 18.98 & 6.07 & 43.49 & 2.91 & 19.32 \\
    RedNet & 0.02 & 0.45 & 19.42 & 6.21 & 44.80 & \textbf{2.44} & 15.40 \\
    RedNet-CD & 0.02 & \underline{0.44} & 19.70 & \underline{7.01} & \underline{48.23} & 2.63 & \underline{14.77} \\
    RedNet-Ens & 0.02 & \underline{0.44} & \textbf{20.19} & \textbf{7.31} & \textbf{48.99} & 2.59 & \textbf{14.27} \\
    \bottomrule
    \end{tabular}
    }
    \end{table}

  \paragraph{Energetics and biochemical properties.} AlphaFold3's confidence metrics do not correlate well with stability or affinity and are not sensitive to mutational changes; higher self-consistency does not necessarily imply better binding. Other biochemical properties, such
  as surface hydrophobicity and aggregation propensity, are also important for practical binder design. We therefore compute energetics and biochemical properties using Rosetta following the BindCraft protocol~\cite{pacesa2025one}. Due to
  stochasticity in Rosetta relaxation, we run three repeats per design and select the structure with the lowest energy. We additionally construct RedNet-Ens, which combines RedNet and RedNet-CD by selecting the design with the best binding score
   for each target (\Cref{tab:energetics,tab:hydrophobicity}).

  \paragraph{Binding energetics.}
  RedNet-Ens achieves the most favorable binding score ($-$188.04 REU), outperforming all individual models including ProteinMPNN ($-$184.89), RedNet-CD ($-$182.26), ESM-IF ($-$181.27), and native sequences ($-$172.42). Contrastive decoding
  improves RedNet's binding score from $-$179.88 to $-$182.26, indicating that the contrastive objective steers sequence selection toward more energetically favorable complexes.

  Interface shape complementarity is comparable across methods (0.64--0.67), with PiFold and RedNet-Ens achieving the highest values (0.67). ProteinMPNN and RedNet-Ens have the best packing (Int Packstat $=$ 0.55). For interface energy,
  RedNet-Ens ($-$56.66 REU) and PiFold ($-$55.19) produce the most favorable values, both surpassing native interfaces ($-$53.35). Notably, ProteinMPNN yields the least favorable Int dG ($-$46.98) despite achieving the best packing, suggesting
  its designs may over-optimize local packing at the expense of global interface energetics.

  RedNet-Ens produces the largest buried surface area (Int dSASA $=$ 1965.96~\AA$^2$), exceeding native interfaces (1918.33~\AA$^2$), while ProteinMPNN yields the smallest (1682.12~\AA$^2$), indicating reduced interface coverage.

  Contrastive decoding consistently improves RedNet across interface metrics: binding score improves from $-$179.88 to $-$182.26, interface free energy from $-$52.49 to $-$54.47, and buried surface area from 1868.29 to 1894.69~\AA$^2$, while
  maintaining comparable shape complementarity (0.66) and packing (0.54).

  \paragraph{Matching native surface hydrophobicity.}
  High surface hydrophobicity may cause aggregation, reduced solubility, and unwanted off-target binding; thus excessive hydrophobicity of binders is undesirable in practice.
  Native interfaces exhibit the lowest surface hydrophobicity (0.43), suggesting a balanced mix of polar and nonpolar residues. ProteinMPNN (0.43), RedNet-CD, and RedNet-Ens (0.44) closely match native surface hydrophobicity, while PiFold is the most hydrophobic (0.47).

  \paragraph{RedNet-CD and RedNet-Ens improve hydrogen bond formation.}
  Hydrogen bonds at protein--protein interfaces contribute to binding affinity: buried polar atoms that lack hydrogen bond partners incur desolvation penalties that destabilize the complex
  \cite{pace2014contribution}. BindCraft accordingly filters for sufficient interface hydrogen bonds ($>$2) and penalizes unsatisfied ones ($<$3) \cite{pacesa2025one}.
  
  RedNet-Ens forms the most interface hydrogen bonds (7.31, 48.99\% of interface residues), exceeding native interfaces (6.97, 46.15\%). RedNet-CD also surpasses native (7.01), while RedNet (6.21) and ProteinMPNN (5.44) produce fewer, suggesting
   that contrastive decoding promotes hydrogen bond formation as a strategy to improve binding, compared to other approaches that rely more on hydrophobic packing.

  RedNet-Ens also achieves the lowest percentage of unsatisfied hydrogen bonds (14.27\%), outperforming all methods and native structures (17.90\%). RedNet achieves the fewest absolute unsatisfied hydrogen bonds (2.44), followed by ProteinMPNN
  (2.52).

  Overall, RedNet-CD and RedNet-Ens produce designs with native-like or superior energetic and biochemical properties, validating that contrastive decoding and ensembling different sampling strategies can yield high-quality interfaces without
  increasing surface hydrophobicity.

\subsection{Contrastive decoding improves binding specificities of binders}

  To our knowledge, no existing benchmarks or standardized metrics have been developed for computationally evaluating the binding specificity of designed protein binders.
  To address this gap, we assemble a selective binder test set from heterodimers in the PDB (detailed in \Cref{subsec:rednet_sel}).
  Each test case consists of an on-target and an off-target receptor that share structural similarity, requiring the designed binder to discriminate between them based on subtle differences in the interface.

  We evaluate specificity using two complementary approaches.
  First, we employ the Rosetta energy function to compute binding scores and assess energetic selectivity, as AlphaFold3's confidence metrics lack the sensitivity to detect mutational changes of binding affinities.
  Second, we use AlphaFold3-based co-folding tests to measure structural selectivity via ipTM thresholds.

  \begin{table}[ht!]
    \caption{Selectivity measured by Rosetta binding score difference (on-target $-$ off-target). A negative value indicates the binder prefers the on-target. SR (Diff $<$ X): percentage of cases where the on-target preference exceeds the energy
   gap threshold X. $\sigma$: backbone coordinate noise level (\AA). Higher is better for all metrics. \textbf{Bold}: best; \underline{underline}: second best.}
    \label{tab:binder_score_diff}
    \centering
    \small
    \begin{tabular}{l c c c c}
    \toprule
    Model & $\sigma$ & SR (Diff $<$ $-$10) $\uparrow$ & SR (Diff $<$ $-$5) $\uparrow$ & SR (Diff $<$ 0) $\uparrow$ \\
    \midrule
    Native & 0 & 25.93\% & 33.33\% & 50.00\% \\
    ProteinMPNN & 0.02 & 31.48\% & 44.44\% & 53.70\% \\
    ESM-IF & 0 & \textbf{35.19\%} & 42.59\% & 53.70\% \\
    PiFold & 0 & 27.78\% & 40.74\% & \underline{55.56\%} \\
    RedNet & 0.02 & 22.22\% & 27.78\% & 33.33\% \\
    RedNet-CD & 0.02 & \underline{33.33\%} & \textbf{51.85\%} & \textbf{64.81\%} \\
    \bottomrule
    \end{tabular}
    \end{table}

  \paragraph{Energetic analysis.}
  Following BindCraft~\cite{pacesa2025one}, we define a composite Binder Score as the sum of the binding free energy ($\Delta G_{\text{binding}}$) and the folding free energy of the designed binder chain ($\Delta G_{\text{binder}}$).
  Equivalently, this is the energy of the complex minus the energy of the receptor ($\Delta G_{\text{complex}} - \Delta G_{\text{receptor}}$).
  This metric captures the total energetic contribution of adding the binder to the system, encompassing both the strength of the interface interactions and the intrinsic stability of the binder in its bound conformation.
  For a binder to be selective, it must exhibit a lower (more favorable) Binder Score when interacting with the on-target receptor compared to the off-target receptor.
  The score difference (Score$_{\text{on}}$ $-$ Score$_{\text{off}}$) thus provides a direct measure of energetic selectivity: a negative value indicates a preference for the on-target (\Cref{tab:binder_score_diff}).

  We evaluate selectivity at three stringency thresholds on 54 on-/off-target pairs filtered to a Jaccard interface similarity of 0.5.
  In the native control group, 50\% of native binders exhibit a lower on-target Binder Score, which is close to the random baseline and validates that our benchmark is well-calibrated: native sequences, not having been optimized for selectivity
  between structurally similar receptors, serve as a sensible control for this energetic discrimination task.

  At the base threshold (Diff $< 0$), RedNet-CD achieves the highest success rate at 64.81\%, a 94\% relative improvement over RedNet without contrastive decoding (33.33\%), and outperforming PiFold (55.56\%, +17\% relative), ProteinMPNN
  (53.70\%, +21\% relative), ESM-IF (53.70\%, +21\% relative), and native sequences (50.00\%, +30\% relative).

  At the moderate threshold (Diff $< -$5), RedNet-CD maintains its lead at 51.85\%, an 87\% relative improvement over RedNet (27.78\%). It also outperforms ProteinMPNN (44.44\%, +17\% relative), ESM-IF (42.59\%, +22\% relative), PiFold (40.74\%,
   +27\% relative), and native sequences (33.33\%, +56\% relative).

  At the strictest threshold (Diff $< -$10), ESM-IF achieves the highest success rate at 35.19\%, followed by RedNet-CD at 33.33\%, which is still a 50\% relative improvement over RedNet (22.22\%) and comparable to ProteinMPNN (31.48\%).

 Overall, RedNet-CD leads at the base and moderate thresholds (64.81\% at Diff $< 0$ (vs.\ 55.56\% for PiFold) and 51.85\% at Diff $< -$5 (vs.\ 44.44\% for ProteinMPNN)), while nearly doubling baseline RedNet (33.33\% and 27.78\%). Only at
   the strictest threshold (Diff $< -$10) does ESM-IF (35.19\%) narrowly lead RedNet-CD (33.33\%). This demonstrates that contrastive decoding specifically enhances the model's ability to discriminate between structurally similar on-target and off-target interactions.

\begin{table}[ht!]
    \caption{Selectivity success measured by AlphaFold3 cofolding. $\sigma$: backbone coordinate noise level (\AA). Selectivity: proportion where on-target ipTM $>$ 0.55 and off-target ipTM $<$ 0.55. On-Target: proportion where on-target ipTM
  $>$ 0.55. Off-Target: proportion where off-target ipTM $>$ 0.55. \textbf{Bold}: best; \underline{underline}: second best.}
    \label{tab:selectivity_success}
    \centering
    \small
    \begin{tabular}{l c c c c}
    \toprule
    Model & $\sigma$ & Selectivity $\uparrow$ & On-Target $\uparrow$ & Off-Target $\downarrow$ \\
    \midrule
    Native & 0 & 3.70\% & \underline{74.07\%} & 74.07\% \\
    ProteinMPNN & 0.02 & \underline{7.41\%} & \textbf{77.78\%} & 75.93\% \\
    ESM-IF & 0 & \underline{7.41\%} & \textbf{77.78\%} & \underline{72.22\%} \\
    PiFold & 0 & 3.70\% & 72.22\% & 75.93\% \\
    RedNet & 0.02 & 5.56\% & \underline{74.07\%} & \underline{72.22\%} \\
    RedNet-CD & 0.02 & \textbf{9.26\%} & 72.22\% & \textbf{70.37\%} \\
    \bottomrule
    \end{tabular}
    \end{table}

\paragraph{Co-folding analysis.}
We also evaluate selectivity via the cofolding test (\cref{tab:selectivity_success}). It is worth noting that AlphaFold3's predictions and confidences are not sensitive to mutational changes and do not discriminate folding or binding free
  energies accurately. RedNet-CD achieves the highest selectivity rate at 9.26\%. However, both RedNet and RedNet-CD ($\sigma = 0.02$) have lower on-target success rates than ProteinMPNN and ESM-IF, only exceeding PiFold and matching the
   native control group. This is expected for RedNet-CD, since it optimizes for selectivity instead of on-target binding.

   We note several limitations of this benchmark. First, the 180 evaluated pairs are uniformly sampled from 656 candidates, which may not fully represent the diversity of selective design challenges in the PDB. Second, the Rosetta energy function used for energetic selectivity evaluation has known biases, which may favor certain types of interfaces over others. Third, AlphaFold3's confidence metrics are not
  sensitive to mutational changes in binding affinity, limiting the informativeness of the co-folding evaluation. Despite these limitations, the benchmark provides a standardized framework for comparing selectivity across design methods, and we expect it to be refined as more accurate scoring functions become available.

\subsection{Structural analysis of redesigned selective binder}

\begin{figure}[t]
  \centering
  \includegraphics[width=\linewidth]{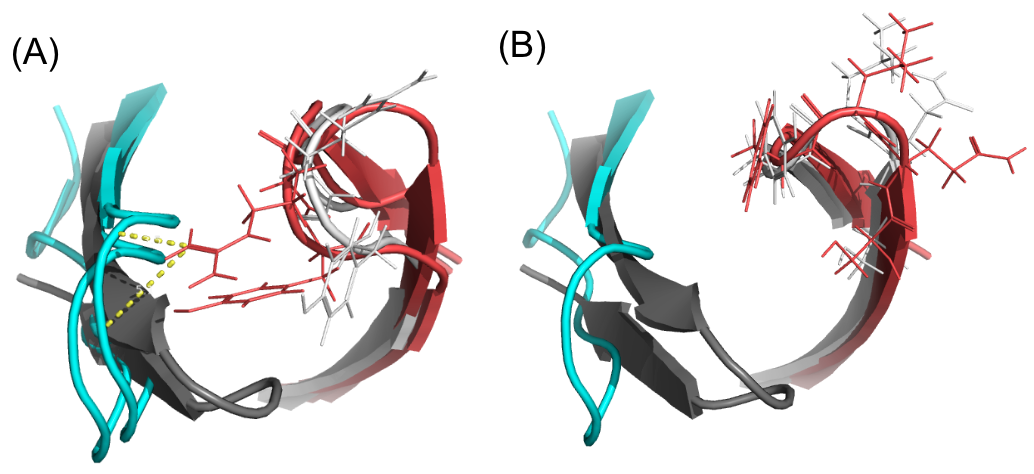}
  \caption{Structural analysis of the 6FOE--5WHJ selective binder pair. (A) Interactions of redesigned binders (red for the design chain of the on-target complex and white for the design chain of the off-target complex) to their respective on-target (cyan) and off-target (grey) partners. (B) Interactions of native binders to their respective on-target and off-target partners.}
  \label{fig:sel_6foe}
\end{figure}

\begin{figure}[t]
  \centering
  \includegraphics[width=\linewidth]{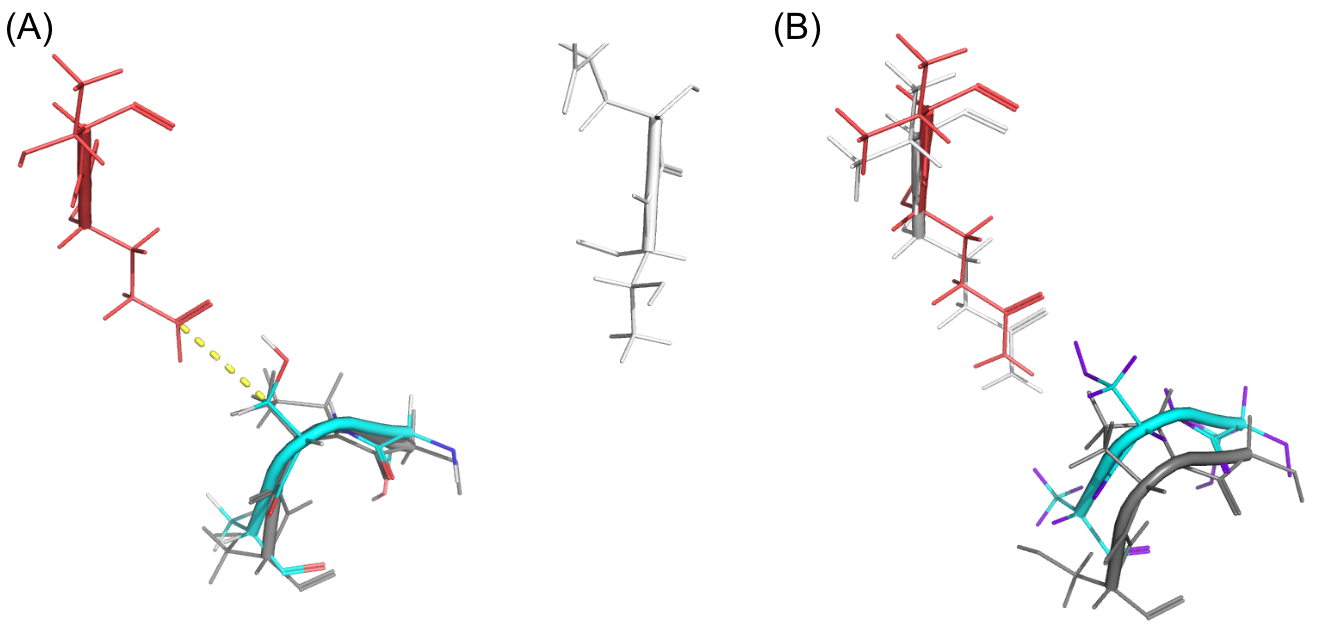}
  \caption{Structural analysis of the 5FFN--1LW6 selective binder pair. (A) Interactions of redesigned binders (red for the design chain of the on-target complex and white for the design chain of the off-target complex) to their respective on-target (cyan) and off-target (grey) partners. (B) Interactions of native binders to their respective on-target and off-target partners.}
  \label{fig:sel_5ffn}
\end{figure}

We investigate how contrastive decoding enables redesigning specific binders through two case studies (\cref{fig:sel_6foe,fig:sel_5ffn}).

\paragraph{Case 1: 6FOE--5WHJ (Fab).}
The first pair is 6FOE (on-target) and 5WHJ (off-target), both of which are Fab complexes.
Compared to the native binder, the redesigned binder mutates 4 contiguous residues at the interface from SQLY to GYRN.
In \cref{fig:sel_6foe}(A), we observe that the redesigned binder tends to form more favorable interactions by mutating L to R to exploit amino acids like F and W on the target chain of 6FOE; while the multipoint mutations are not favorable for the off-target partner.
Despite the backbone structures of the target chains being similar (RMSD $=$ 1.89~\AA), RedNet-CD is capable of exploiting side-chain differences to enhance on-target interactions.

\paragraph{Case 2: 5FFN--1LW6 (Subtilisin).}
The second pair is 5FFN (on-target) and 1LW6 (off-target).
The target chains of 5FFN and 1LW6 are both Subtilisin, and the native binders in 5FFN and 1LW6 are chymotrypsin inhibitors CI2A and CI2, respectively.
The interfaces of the two complexes share significant similarities (Jaccard similarity $=$ 0.38), and the two target chains are structurally similar (RMSD $=$ 2.08~\AA).
A segment on the interfaces of the redesigned binder is mutated from QV to ET.
In \cref{fig:sel_5ffn}(A), we observe that the redesigned binder is able to retain strong interactions with the polar interfaces of the on-target partner (SSA), while it has poor interactions with the off-target partner (ET).
This demonstrates that RedNet-CD can improve specificity by destabilizing the off-target interfaces.

\section{Conclusion}
We have presented RedNet, a framework for fixed-backbone binder sequence design that incorporates an all-atom graph transformer architecture and a contrastive decoding algorithm. The all-atom graph transformer captures side-chain information from the target, enabling improved sequence recovery over existing methods, particularly on heterodimeric interfaces that are most relevant to one-sided binder design. Contrastive scoring approaches improve zero-shot binding affinity prediction, and contrastive decoding further enhances heterodimer self-consistency and energetics, producing designs with native-like or superior physicochemical properties. On a newly curated selective binder benchmark, we demonstrate that contrastive decoding can discriminate between structurally similar on-target and off-target receptors by exploiting subtle side-chain differences at the interface. The flexibility of the contrastive decoding framework could enable a broader range of multistate design tasks without requiring model retraining.

\bibliographystyle{siam}
\bibliography{references}

\chapter{Conclusions and Future Directions}

This thesis investigates two fundamental aspects of modeling the sequence-structure relationship of protein complexes using deep learning: domain-specific architectures and search algorithms. These two aspects are closely related to the
longstanding problems of parameterization and sampling in modeling protein dynamics and protein design.

Protein structures are hierarchical, spanning atoms and residues, single domains and chains, to multi-chain assemblies. This hierarchy enables the design of domain-specific deep learning architectures that extract the most from biomolecular
  data such as protein structures in the PDB for protein complex prediction and design. We explore how graph neural networks and transformers with domain-specific inductive biases can accurately and efficiently model different aspects of protein
   structure. 
   
   Proteins are not only physical but also evolutionary and contextual; different aspects of proteins can be measured by different experimental techniques, and it is important that deep learning can integrate different modalities of
  experimental data to make biologically relevant inferences of protein conformational changes and functional effects, accurately and efficiently across different contexts. We develop deep learning-based algorithms
  that integrate protein structures, evolutionary information such as multiple sequence alignments of monomers and inferred paralogs, and assay data, demonstrating the effectiveness of deep learning in extracting complementary information from
  diverse data sources. 
  
  Protein complex structure prediction and structure-based binder design are closely related problems: the former aims to model essential aspects of the distribution of protein complex conformations given sequences, and the
   latter solves the corresponding inverse problem. We successfully apply the multiscale architectures developed for modeling protein complexes to the inverse design problem, further demonstrating the generality of these deep learning approaches
   for protein complex modeling and design.

  Another core aspect of modeling protein complexes is the search problem. Both the sequence and structure spaces of proteins are vast, and efficiently sampling these spaces using deep learning models is an active area of research. In GLINTER (chapter 2),
   we use predicted interfacial contacts to constrain the search of docking poses; recent works have also explored using predicted protein contacts to guide the search of protein complex conformations and protein-protein interactions with
  pretrained models. In ESMPair (chapter 3), we demonstrate how to utilize deep learning models as scoring functions and transferable heuristics to guide the search for plausible interacting paralog pairs, boosting heterodimer prediction accuracy. In
  RedNet (chapter 4), we develop sampling algorithms based on thermodynamic principles and autoregressive architectures to improve binding affinities and specificities — two core engineering parameters in protein binder design applications.

\section{Future improvements}

\paragraph{Efficient deep learning architectures for all-atom biomolecular structures.} AlphaFold2 established domain-specific end-to-end architectures for protein structure modeling, and AlphaFold3 effectively extended AlphaFold2 to
  additional chemical modalities, including nucleic acids, small molecules, glycans, and other common post-translational modifications such as phosphorylation. However, many challenges remain in current AlphaFold architectures and their
  variants. The memory cost of storing pairwise activations, the inference cost of pairwise attention and multiplicative updates, and even naive full attention for large complexes and all-atom conformations are all expensive. Graph neural networks, as explored in our
  work, and other sparse architectures such as linear attention and local attention, are attractive — and even necessary — alternatives, particularly for modeling and simulating all-atom structures and large complexes.

  Developing effective graph neural networks and sparse architectures, distilling existing models \cite{polino2018model}, and optimizing kernels \cite{dao2022flashattention} to greatly improve efficiency without
  sacrificing accuracy could enable protein complex modeling in many applications, such as more thorough exploration of conformational landscapes, higher-throughput screening of designs and mutation scanning, and modeling pathways involving many
   complexes.

  \paragraph{Improved integration of multiple data and chemical modalities.}
  Improving architectures and training models that can effectively integrate multiple sequence / structure alignments, structured and unstructured functional annotations, molecular dynamics trajectories \cite{mirarchi2024mdcath,lewis2025scalable}, and assay data \cite{tsuboyama2023mega} — and that can generalize to related
  problems — is another ongoing challenge.

  Many pretrained models have incorporated different data modalities \cite{su2023saprot,nijkamp2023progen2,hayes2025simulating,lewis2025scalable,chen2025xtrimopglm}; however, few have convincingly demonstrated that a pretrained model can zero- or few-shot generalize to related tasks reliably \cite{notin2023proteingym,chungyoun2024flab,bhatnagar2025scaling,akiyama2025scaling}.
  RedNet attempts to generalize to binding affinity prediction in a zero-shot setting; despite outperforming existing methods, there remains substantial room for improvement. Given that the contrastive decoding algorithm for binder design
  follows thermodynamic principles, it is natural to fine-tune current models on folding free energy using supervised \cite{dieckhaus2024transfer} or reinforcement learning \cite{widatalla2024aligning} approaches, such as with the Megascale dataset \cite{tsuboyama2023mega}, and generalize to binding affinity prediction.

  AlphaFold3-like architectures are capable of handling different chemical modalities, which greatly extends their generality. This opens many new possibilities in modeling cross-modality molecules such as metalloenzymes and
  protein--small-molecule conjugates, where conventional molecular dynamics approaches are lacking. Currently, GLINTER and RedNet are trained only on protein structures but can be trivially extended to model all-atom structures due to their
  heavy-atom representations; a natural next step would be to extend them to other modalities in the PDB.

\paragraph{Improved deep learning algorithms for modeling protein physics.}

  Existing complex prediction and design deep learning models are typically trained to recover the geometry of native structures, such as distograms and heavy-atom coordinates and tend to make blurry predictions. Several types of interatomic interactions are not well captured, including hydrogen bonding, solvent effects,
  and electrostatics \cite{musil2021physics}. Improved treatment of these interactions could yield more accurate atomic details, better decoy ranking, and more reliable modeling of conformational changes — all of which remain challenging for current protein complex
  prediction and design models. Such improvements would also enable new applications, such as designing pH-sensitive binders \cite{strauch2014computational,boyken2019novo} or highly functional enzymes \cite{rothlisberger2008kemp}.

 Rotation and translation equivariance are inherent properties of protein dynamics. In practice, however, state-of-the-art structure prediction models — AlphaFold2 (which, despite using invariant point attention, showed in ablation studies that
  equivariance contributes little to final performance) and AlphaFold3 (which removes equivariance altogether) — do not benefit greatly from equivariant architectures. This is not to say equivariance is unimportant for modeling protein
  conformations and dynamics. For tasks requiring structural validity \cite{fu2022forces}, equivariance has proven beneficial and enables better extrapolation at test time. Equivariant architectures also offer an elegant way to capture many-body effects, which may
  be essential for certain protein dynamics applications beyond recapitulating experimental structures. GLINTER is among the first methods to use equivariant neural networks that exploit the
  regularity of amino acid backbone geometry to predict inter-protein contacts. RedNet combines the same backbone geometry with equivariant graph attention networks for all-atom structures in protein design, demonstrating promising improvements
  over pairwise distance-based features. It remains an
  open question whether and how to incorporate equivariance — in terms of training dynamics and hardware-aligned architecture design — for modeling protein dynamics, and for which application-specific metrics equivariances are essential.

  Diffusion models, closely related to energy-based models, are promising alternatives for modeling protein biophysics. Diffusion modeling and molecular dynamics have deep theoretical connections and have seen fruitful cross-pollination in
  techniques for accelerated sampling and fine-tuning. Recent works from other groups as well as our ongoing work based on RedNet architectures have shown promising results in modeling protein physics with diffusion models.

\paragraph{End-to-end protein design architectures.} Current de novo protein design pipelines typically rely on a three-stage setup \cite{chu2024sparks}: backbone \cite{watson2023novo,ingraham2023illuminating} or all-atom structure generation \cite{chu2024allatom,qu2024pallatom} structure-based sequence redesign \cite{dauparas2022robust}, and ranking using deep learning
  structure prediction models \cite{roney2022state} or physics-based force fields \cite{bennett2023improving,pacesa2025one}. These stages are deeply connected from a Bayesian perspective, and a capable end-to-end model could improve the accuracy of all three while increasing efficiency. RedNet can readily
  scan mutations and is straightforward to extend to simultaneous side-chain conformation prediction. Alternative architectures and training algorithms, such as energy-based diffusion models \cite{roney2025protein}, are also promising and may prove more general — owing
   to their connection with protein dynamics — and more flexible for end-to-end de novo design.

  \paragraph{Accelerated search of conformation and sequence spaces.} Current deep learning complex prediction and design models use off-the-shelf search algorithms at inference time. Despite this, significant accuracy gains have been
  achieved by increasing the number of seeds and recycling iterations \cite{johansson2022improving, gao2022af2complex} and tuning hyperparameters such as penalty weights and temperatures \cite{frank2024scalable, pacesa2025one}. ESMPair is a proof-of-concept deep learning algorithm for searching interacting paralogs to improve structure predictions. It can
  be extended, like AFCluster\cite{wayment2024predicting}, to predict multiple conformations of protein complexes, and coupled with improved confidence prediction and decoy ranking, can further improve complex prediction accuracy.

\paragraph{Improved treatment of modality- and application-specific constraints for protein design.} Compared to structure prediction, protein design is more open-ended: different applications impose different engineering requirements, and
  effective design typically demands domain knowledge about both the modality (e.g., whether to use a nanobody or monoclonal antibody, and which regions to maintain versus mutate to preserve functional residues) and the application (e.g.,
  whether the engineering parameters are geometrical, mechanical, or thermodynamic). For practical applications, it is important to make deep learning models more controllable to accommodate these diverse constraints and to optimize different
  objectives \cite{zhou2024general} . RedNet demonstrates that by considering the requirements of designing specific binders, one can develop deep learning algorithms that are more effective at designing and distinguishing binders with improved specificity. RedNet can be easily finetuned to model important therapeutic modalities — including monoclonal antibodies \cite{eguchi2022igvae,dreyer2023abmpnn,hoie2025antifold,bennett2026atomically} and, more broadly, the immunoglobulin super family — to improve design performance.

\bibliographystyle{siam}
\bibliography{references,conclu_references}

\clearpage

\hypersetup{linkcolor=cyan}
\setcitestyle{numbers}

\end{document}